\documentclass[compsoc]{IEEEtran}

\usepackage{cite}
\usepackage[pagebackref=true,breaklinks=true,colorlinks,bookmarks=false]{hyperref}       % hyperlinks
\usepackage{url}            % simple URL typesetting
\usepackage{float}
\usepackage{graphicx}
\usepackage[export]{adjustbox}% http://ctan.org/pkg/adjustbox
\usepackage{subfig}
\usepackage[table]{xcolor}
\usepackage{amsmath,mathrsfs}
\usepackage{amsfonts}
\usepackage{array}
\usepackage{multirow}
\usepackage{CJK}
\usepackage{comment}
\usepackage{cancel}
\usepackage{booktabs}
\usepackage{mmstyles}
\usepackage{tikz}

\usepackage{threeparttable}

\usepackage{framed}

\allowdisplaybreaks[4]

\usepackage{xcolor}

\usepackage{makecell}

\usepackage{enumitem}

\usepackage{ifthen} % 引入条件判断包

\usepackage{supply_material/cross_cite}
\usepackage{lettrine}
\usepackage{standalone}

\def\Var{{\textrm{Var}}\,}

\usepackage{siunitx}

\newboolean{editedmode}
\setboolean{editedmode}{true} % 将 edited 模式设置为 true 或 false

\newif\ifeditedmode
\editedmodetrue % 打开 edited 模式
\editedmodefalse % close edited mode

\newcommand{\edited}[1]{%
    \ifeditedmode
        \begingroup
        \color{blue}#1%
        \endgroup
    \else
        #1%
    \fi
}

\begin{document}
\title{
%{$\rho$-Vision}: Camera RAW Snapshots Are Enough for Real-Time Resource-Efficient Visual Analytics
%
{
% {$\rho$-Vision}: 
Efficient Visual Computing with Camera RAW Snapshots} \\
% \zhihao{{$\rho$-Vision}: Enabling RAW-CV through the Simulation and the Compression} \\
%\Salnote{{$\rho$-Vision}:  Efficient Computer Vision with RAW Images ???} \\
%\xzhang{\sys: Unpaired RAW Image Enhancement (Generation) for Efficient Computer Vision ???}
}

\author{Zhihao Li,
        Ming Lu,
        Xu Zhang,
        Xin Feng,
        M. Salman Asif, % ~\IEEEmembership{Senior Member,~IEEE}
        and
        Zhan Ma%,~\IEEEmembership{Senior Member,~IEEE}

%\thanks{This paper is supported in part by National Natural Science Foundation of China (62022038, U20A20184).  (Corresponding Authors: Z. Ma)}
\thanks{This paper is supported in part by the National Natural Science Foundation of China (62022038, U20A20184, 62171174), the Key project of Natural Science Foundation of Chongqing, China under Grant No.CSTB2022NSCQ-MSX0493,
Chongqing Technology Innovation and Application Development under Grant No.cstc2021jscx-dxwtBX0018. (Corresponding authors: Z. Ma and X. Feng.)}
\thanks{Z. Li, M. Lu, and Z. Ma are with the School of Electronic Science and Engineering, Nanjing University, Nanjing, Jiangsu, China. M. Lu is also with the Interdisciplinary Research Center for Future Intelligent Chips (Chip-X), Nanjing University, Suzhou, Jiangsu, China. E-mails: lizhihao6@smail.nju.edu.cn, \{minglu, mazhan\}@nju.edu.cn}
\thanks{X. Zhang is with the University of Leeds, Leeds LS2 9JT, United Kingdom. Email: x.zhang15@leeds.ac.uk}
\thanks{X. Feng is with the Chongqing University of Technology, Chongqing, China. Email: xfeng@cqut.edu.cn}
\thanks{M. S. Asif is with the University of California Riverside, CA 92521. Email: sasif@ece.ucr.edu}
}

\newcommand{\Salnote}[1]{{\bf \color{blue} [SA:{#1}]}}
\newcommand{\MaZnote}[1]{{\color{blue} {#1}}}

\IEEEtitleabstractindextext{
\begin{abstract} 
Conventional cameras capture image irradiance (RAW) on a sensor and convert it to RGB images using an image signal processor (ISP). The images can then be used for photography or visual computing tasks in a variety of applications, such as public safety surveillance and autonomous driving.  One can argue that since RAW images contain all the captured information, the conversion of RAW to RGB using an ISP is not necessary for visual computing. In this paper, we propose a novel $\rho$-Vision framework to perform high-level semantic understanding and low-level compression using RAW images without the ISP subsystem used for decades.
Considering the scarcity of available RAW image datasets, we first develop an unpaired CycleR2R network based on unsupervised CycleGAN to train modular unrolled ISP and inverse ISP (invISP) models using unpaired RAW and RGB images. We can then flexibly generate simulated RAW images (simRAW) using any existing RGB image dataset and finetune different models originally trained in the RGB domain to process real-world camera RAW images. We demonstrate object detection and image compression capabilities in RAW-domain using RAW-domain YOLOv3 and RAW image compressor (RIC) on camera snapshots.
Quantitative results reveal that RAW-domain task inference provides better detection accuracy and compression efficiency compared to that in the RGB domain. Furthermore, the proposed $\rho$-Vision generalizes across various camera sensors and different task-specific models. An added benefit of employing the $\rho$-Vision is the elimination of the need for ISP, leading to potential reductions in computations and processing times.
% Additional advantages of the proposed $\rho$-Vision that eliminates the ISP are the potential reductions in computations and processing times.  

% \Salnote{I like $\rho$-Vision as a title, but be aware that it will be a pain to write it in ascii and it will be difficult to search for this paper. Consider changing to RAW-Vision} 

% \input{sections/Abstract}
\end{abstract}

}
\maketitle
\ifCLASSOPTIONpeerreview
\begin{center} \bfseries EDICS Category: 3-BBND \end{center}
\fi
\IEEEpeerreviewmaketitle

\maketitle

\section{Introduction}
\lettrine[lraise=0.1, nindent=0em, slope=-.5em]
{C}{onventional} cameras capture visual information in a scene and present it in the RGB (or equivalent YCbCr) format for subsequent visual computing (e.g., semantic understanding and communication). This pipeline is prevalent in a variety of applications~\cite{lin2023motion,wang2023image,yang2022multi}, such as %autonomous vehicles~\cite{533960},
smart communities~\cite{10.1145/3529511}, and surveillance systems~\cite{raty2010survey}. For instance, instantaneous RGB snapshots enable the detection of driving lanes or pedestrians in advanced driver assistance systems~\cite{okuda2014survey} to improve road safety and risk prevention. 

\begin{figure*}[ht]
\centering
\subfloat[Traditional RGB-Vision pipeline]{\includegraphics[width=0.48\linewidth]{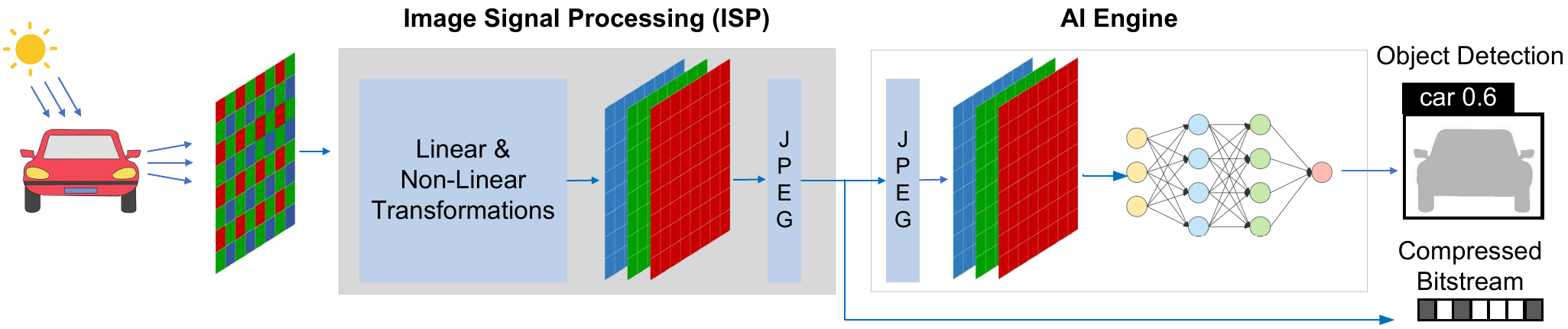} \label{fig:RGB-vision-w-ISP}}
\subfloat[The proposed $\rho$-Vision framework]{\includegraphics[width=0.48\linewidth]{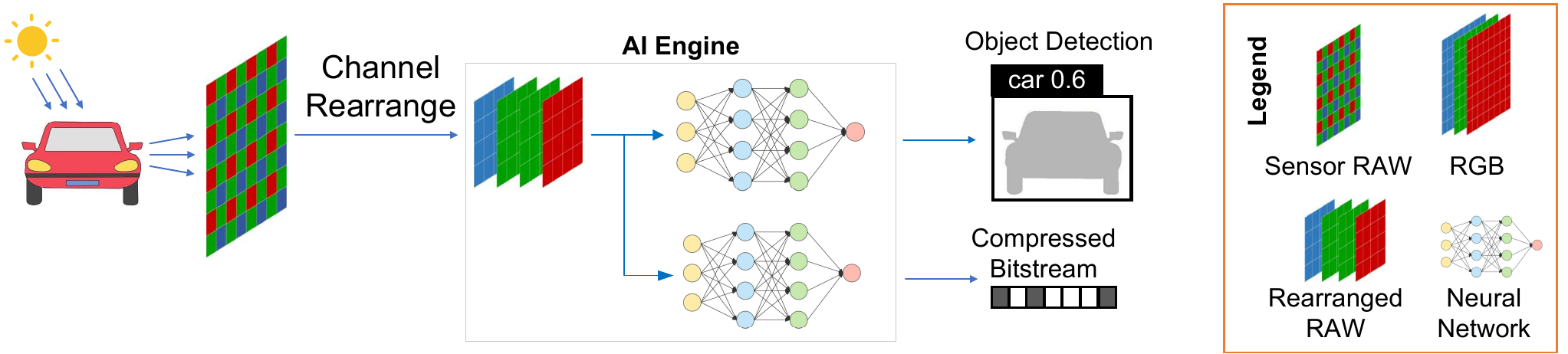} \label{fig:rho-vision-wo-ISP}}
%\end{center}
  \caption{{\bf Cameras for visual computing tasks.} (a) Traditional RGB-Vision framework (with in-camera ISP) executes tasks with RGB images; (b) Proposed $\rho$-Vision framework (without conventional ISP) executes with RAW images directly. 
%   ~\lm{(No reference for Fig1. b in the main paper.)}
  }
\label{fig:CV_RGB_RAW}
\end{figure*}

The camera sensor and image signal processor (ISP) are tightly coupled for traditional RGB-Vision, as illustrated in Fig.~\ref{fig:RGB-vision-w-ISP}. First, a CMOS or CCD sensor records color pixels in a Bayer pattern as a RAW image. Then, an ISP converts the RAW image to RGB representation through a series of linear and nonlinear steps (e.g., demosaicing, white balance, exposure control, gamma correction, and JPEG compression)~\cite{brady2020smart}. 
%,wallace1992jpeg
The compressed RGB images are then processed for various vision tasks and potentially stored for archival or review purposes.

The classic RGB-Vision pipeline used in cameras for decades has significant redundancies.
As for the surveillance video usage reported by leading video hosting firms, less than \SI{1}{\percent} of recorded videos are reviewed by human subjects~\cite{live_video_usage}. \edited{
The transformation process of the ISP is not just resource-intensive with additional hundreds of mW power consumption~\cite{buckler2017reconfiguring}—but also introduces additional processing delays. These delays are particularly detrimental in latency-sensitive applications, such as autonomous vehicles.
Moreover, domain discrepancy is inevitable if we train and test models using RGB images generated from different ISPs, adversely affecting the inference accuracy\footnote{Due to the space limitation, more details about this real-world experiment using commodity hardware are provided in the supplementary material.}. %The final model's inference accuracy is also highly contingent on the quality of the ISP.
% 
% To substantiate these claims, we have conducted comprehensive hardware implementations detailed in the supplementary material. These experiments provide a thorough comparison of accuracy, delay, and power costs associated with different ISPs, thereby demonstrating their substantial impact on system performance. This empirical evidence underlines the necessity to reconsider the role of ISPs in the current RGB-Vision pipeline, especially in applications where efficiency and speed are paramount.
}
This raises a question \textit{why do we need the ISP to convert RAW images to the human-perceivable RGB format?}
% 
% The transformation from RAW to RGB in ISP not only consumes resources (e.g., several hundred mW power consumption~\cite{buckler2017reconfiguring}), but also incurs additional processing delay that can be critical for time-sensitive services such as autonomous vehicles. 

In this work, we present a $\rho$-Vision framework\footnote{Greek letter $\rho$ represents the ``RAW'' for similar pronunciation.} to execute both high-level and low-level tasks directly on camera RAW images. The key steps of $\rho$-Vision pipeline are illustrated in Fig.~\ref{fig:rho-vision-wo-ISP}. % 
As an increasing number of artificial intelligence (AI) chips are installed on cameras~\cite{AI_camera}, it is feasible to utilize in-camera AI chips for RAW image processing in various tasks. 

\edited{One fundamental challenge in developing models for RAW-domain visual computing is the {\it lack sufficient RAW images and task-associated label annotations to train robust RAW-domain models,} since existing models are mainly developed for the RGB domain, large-scale, publicly accessible datasets consisting of RGB images (e.g., ImageNet~\cite{imagenet_cvpr09}).} On the other hand, directly applying pre-trained RGB-domain models to RAW images can lead to catastrophic performance degradation (see Table~\ref{table:without_labeled_detection}).

To address the challenges posed by dataset limitations, we developed the Unpaired CycleR2R model. This model leverages existing RGB images to generate simulated RAW (simRAW) images. These simRAW images are produced using inverse ISP (invISP) methods, with which fine-tuning the existing RGB-domain models for RAW-domain tasks with uncompromised performance is assured.
\edited{Training the Unpaired CycleR2R does not require paired RGB and RAW images. Instead, we can use the existing large-scale RGB dataset (e.g., 100,000 samples) and a much smaller RAW dataset (e.g., 7,000 images) to fulfill the purpose.  This is particularly valuable given the challenges and high costs of acquiring a large-scale RAW image dataset (samples and label annotations). The Unpaired CycleR2R not only transforms the existing RGB dataset to its RAW-domain counterpart but also allows us to reuse its labels for subsequent task models directly, completely avoiding the tedious and expensive labeling.} %By integrating a large, more easily accessible RGB dataset with a smaller RAW dataset, our method not only ensures a rich information base but also optimizes resource utilization.}
We then run a RAW-domain YOLOv3 to process RAW images, reporting better detection accuracy than the corresponding RGB-domain YOLOv3 used in several applications~\cite{yolo}. We also extend a variational autoencoder (VAE) based lossy/lossless RAW Image Compressor (RIC) from the TinyLIC~\cite{lu2022high} for RAW image compression. The resulting model shows superior performance to commercial approaches in both lossy and lossless modes. 

This paper makes the following contributions:
\begin{enumerate}[leftmargin=0.5cm]
\item {\bf Unpaired CycleR2R} for conversion between RAW and RGB images. We train a CycleGAN to train an ISP for RAW-to-RGB and an invISP for RGB-to-RAW transformation (R2R) using unpaired RGB and RAW images. Unsupervised learning using unpaired samples makes our approach much more accessible for practical implementation. In contrast, existing solutions (e.g., CycleISP~\cite{zamir2020cycleisp}, CIE-XYZ Net~\cite{afifi2020cie}, and MBISPLD~\cite{conde2022model}) are supervised models that require paired RAW and RGB images (from the same camera model). 
\begin{itemize}
% [leftmargin=0cm]
    \item Instead of training a generic deep network for ISP and invISP, we apply the modular unrolling to mimic functional steps in ISP and invISP subsystems, to which each step is motivated by imaging physics. 
    \item Since the same scene can be mapped into the different RAW samples by setting different brightness and color temperature levels, we characterize the probabilistic distribution of the illumination instead of using a fixed setting to best represent the practical conditions for the modeling of invISP/ISP.
\end{itemize}

\item {\bf RAW-domain models} (in principle) can be obtained by retraining corresponding RGB-domain models using the simRAW images generated using the proposed invISP. Such domain adaptation approaches~\cite{zhu2017unpaired, hnewa2021multiscale, li2022cross} need to be separately engineered for each individual task. 

\begin{itemize}
\item In our experiments, we demonstrate that YOLOv3 and RIC finetuned using simRAW samples provide outstanding performance for object detection and image compression in the RAW domain. We can consistently enhance their performance by further finetuning simRAW-tuned models with limited real RAW images from various camera models. %We observed  performance improvement across various camera models. 
%We also observe that 
Such a lightweight, few-shot fine-tuning method generalizes our method for camera-specific RAW-domain processing, which is attractive for practitioners.
\item To encourage the reproducible research, {a labeled MultiRAW dataset} that contains $>$7k RAW images acquired using multiple camera sensors is made publicly accessible for RAW-domain processing.
\end{itemize}

\end{enumerate}
\section{Related Work} \label{sec:related_work}

This section briefly reviews relevant approaches  for camera  ISP, RAW image processing, and simRAW generation.

\subsection{Camera ISP} \label{sec:camera_isp_review}
Modern ISP converts sensor RAW data to RGB images using a series of computations, as shown in Fig.~\ref{fig:CV_RGB_RAW}a. First, linear transformations, including demosaicing, white balance estimation, brightness adjustment, and color correction, are applied to map native RAW input to an intermediate format conforming to the CIE 1931 XYZ color space~\cite{karaimer2016software}. Subsequently, a sequence of nonlinear transformations translates the image from XYZ to RGB color space. \edited{Typical nonlinear operations include gamma correction and local transformations (e.g., denoising, sharpening, local tone mapping).} Then, a JPEG encoder is used to compress the RGB images for storage or transmission. The entire processing pipeline of such ISP subsystems is widely used in commodity cameras. A white paper on the ISP system can be found here~\cite{white_paper}.
% \footnote{\url{https://www.pathpartnertech.com/camera-tuning-understanding-the-image-signal-processor-and-isp-tuning/}}.
% \Salnote{Why not cite this as a proper reference?} 
To summarize, the ISP mainly converts sensor RAW samples to human-perceivable RGB images, which induces redundant computations, as discussed earlier. 

\subsection{RAW Image Processing}
Although most image processing algorithms have been developed for RGB images, recent explorations on RAW images have revealed superior performance for various tasks (e.g., denoising, deblurring)  \cite{nguyen2016raw,zamir2020cycleisp,conde2022model}.
%abdelhamed2020ntire,brooks2019unprocessing,
For instance, Brooks \etal.~\cite{brooks2019unprocessing} applied the UPI,  Zamir \etal.~\cite{zamir2020cycleisp} proposed a CycleISP, Conde \etal.~\cite{conde2022model} developed a learned dictionary-based model to convert an RGB image to its RAW format for denoising. 

One challenge with processing RAW images is their strong dependence on specific sensors and devices, which makes the aforementioned methods difficult to generalize to multiple sensors. Afifi \etal.~\cite{afifi2020cie} suggested processing images in device-independent CIE XYZ format that could be easily mapped from the sensor-specific RAW image via a colorimetric conversion. They not only reported performance improvement for various low-level tasks (e.g., denoising, deblurring, and defocusing) but also demonstrated the model generalization without requiring per-sensor supervision.  

Existing works mainly focus on the low-level processing of RAW images. In this paper, we explore high-level semantic understanding tasks (e.g., object detection and segmentation). We also investigate low-level RAW image compression (RIC) for two reasons: 1) RIC is a commodity feature vastly used in cameras to ensure efficient storage and exchange of image snapshots;  2) the studies of denoising and deblurring  in~\cite{afifi2020cie,zamir2020cycleisp,conde2022model,brooks2019unprocessing} can be easily extended in our framework. To the best of our knowledge, this work, together with our earlier work  in~\cite{zhihao_asilomar} offers the first study on the lossy and lossless compression of RAW images.

\subsection{simRAW Generation}
RAW-domain visual computing has promising prospects, as demonstrated by existing work, but training large neural networks for RAW-Vision requires large annotated datasets with RAW images. \edited{Collecting RAW images is somewhat easy and straightforward. However, the associated annotation labeling in the RAW domain is expensive and burdensome.}
\edited{Also, when we refer to a dataset used for the vision task, we generally assume the composite of the image samples and their label annotations.}
% \footnote{Both sensor RAW and CIE XYZ representations belong to the linear color space format. We categorize them together for discussions. \Salnote{I don't see the point of this footnote here.}} 
%
Thus, one approach is to generate RAW images from prevalent labeled RGB datasets, which requires reverse engineering the ISP, which we call invISP. Earlier algorithms, such as InvGamma~\cite{koskinen2019reverse}, assume the availability of spectral characterization of a target camera to train the reverse imaging pipeline. Recently, CycleISP~\cite{zamir2020cycleisp} and CIE-XYZ Net~\cite{afifi2020cie} suggested learning invISP module by fully leveraging the nonlinear representative capacity of underlying deep neural networks (DNNs). Training such models requires a large number of paired RAW and RGB images (e.g., DND dataset used by CycleISP~\cite{DND} and MIT-Adobe FiveK~\cite{MITFiveK} used in CIE-XYZ Net). 

% \zhihao{TBD: change all conde into MBISPLD according the notion in their latest paper"Reversed Image Signal Processing and RAW
% Reconstruction. AIM 2022 Challenge Report"}
DNNs trained to characterize invISP (and ISP, if applicable) are hardly interpretable. 
Brooks \etal.~\cite{brooks2019unprocessing} (UPI) and Conde \etal.~\cite{conde2022model} (MBISPLD) applied algorithm unrolling~\cite{AlgrithmUnrolling} to model modular components in the ISP subsystem by leveraging imaging physics. Such modular unrolling-based approaches were expected to require a small number of sample pairs for training~\cite{conde2022model}. Yet, UPI and MBISPLD still need paired RGB and RAW images, which limits the application in converting existing RGB datasets captured by unknown cameras, like BDD100K~\cite{yu2020bdd100k} into RAW format.

% which hinders the model generalization to diverse camera models. \Salnote{What is Conde? Author name or model name? Do they use large number of images? Why using large number of images hinders generalization?} 
% Capturing a sufficiently large number of RAW and RGB pairs is expensive and cumbersome; however, it is convenient to shoot RAW images using commodity cameras. On the other hand, image datasets with various RGB samples are also easy to obtain.
%\zhihao{When a substantial labeled RGB dataset is at hand, extensive annotation for a new RAW dataset becomes burdensome and unnecessary. With our approach, we can efficiently convert the existing RGB dataset into a RAW dataset using unpaired RGB to RAW conversion.}

We propose the ``Unpaired CycleR2R'' to properly model the invISP and ISP functions. We not only follow the CycleGAN structure~\cite{zhu2017unpaired} to characterize the mapping functions using unpaired RAW (instantaneously captured by  cameras) and RGB (obtained from existing datasets) images in an unsupervised manner but also enforce the modular unrolling approach for more robust model derivation. Though our method shares the general architecture of modular unrolling  for invISP/ISP modeling with state-of-the-art MBISPLD~\cite{conde2022model}, our method suggests the use of a probabilistic model to reflect non-bijective functional mapping in ISP modules. This is because the same RGB may come from a variety of RAWs acquired using different sensors or the same sensor with different illumination settings, while MBISPLD~\cite{conde2022model} strictly assumes the bijective mapping in the ISP subsystem.

\section{Unpaired CycleR2R}
\label{sec:gan}
\definecolor{g_inv_isp_color}{rgb}{0.97, 0.80, 0.68}
\definecolor{f_isp_color}{rgb}{0.85, 0.89, 0.95}
\definecolor{gcc_color}{rgb}{0.75, 0.0, 0.0}
\definecolor{fcc_color}{rgb}{0.44, 0.68, 0.28}

\begin{figure*}[ht]
\centering
\subfloat[Unpaired CycleR2R Framework.]{\includegraphics[width=0.95\linewidth]{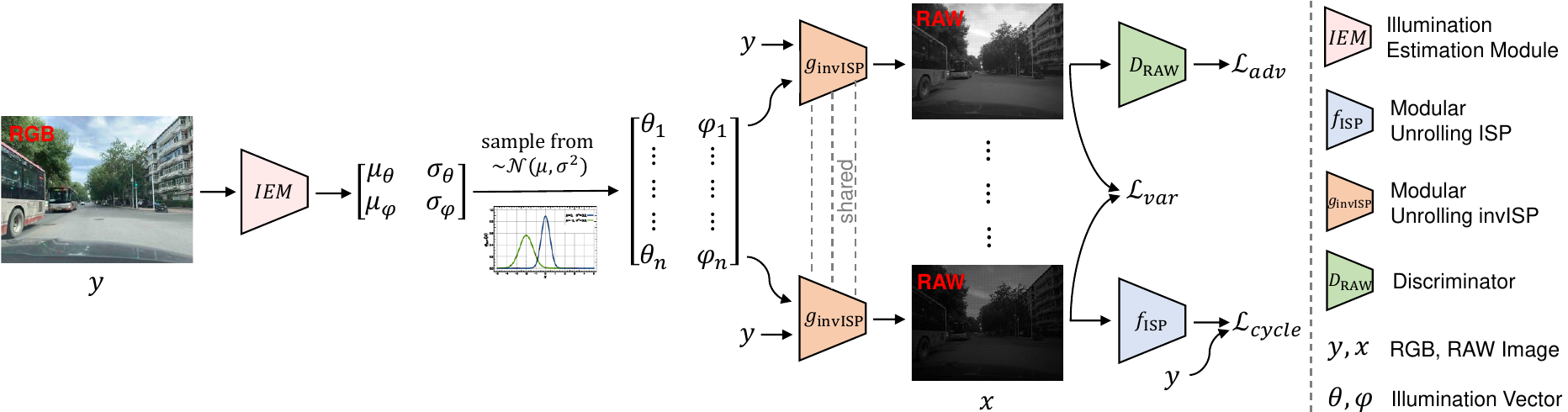} \label{fig:GAN-ISP-Framework}}

\subfloat[Detail architectures of 
\colorbox{g_inv_isp_color}{\color{black}$g_{\rm invISP}$}
 and 
\colorbox{f_isp_color}{\color{black}$f_{\rm ISP}$}
.]{\includegraphics[width=0.95\linewidth]{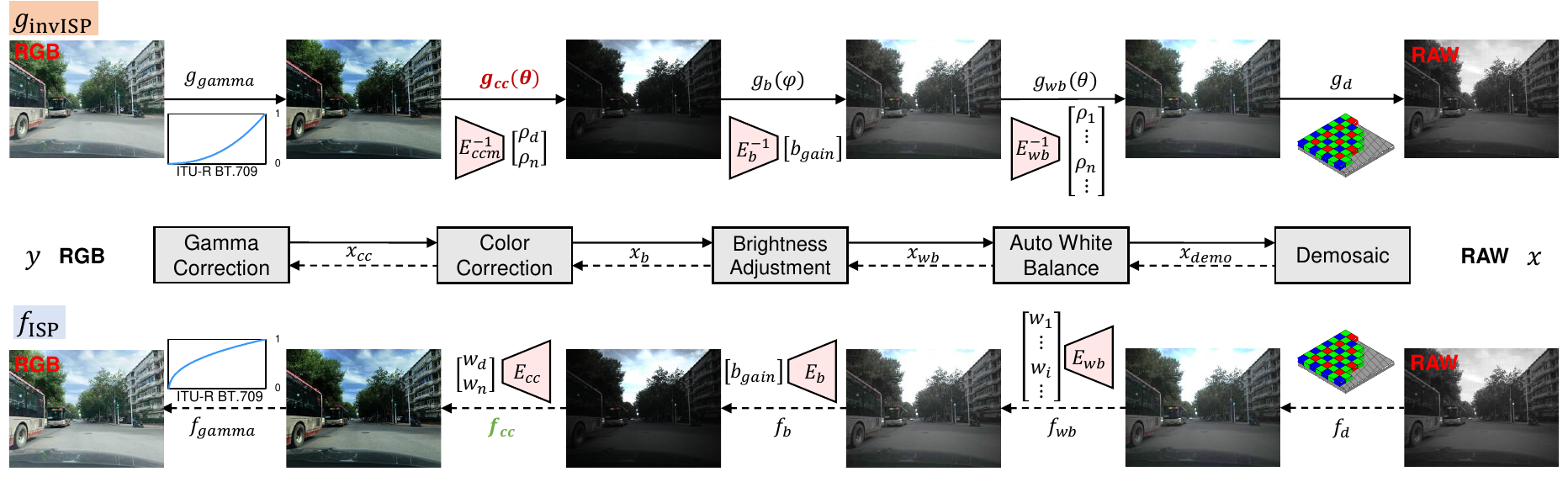} \label{fig:GAN-ISP-Details}}

\subfloat[Color correction \textcolor{fcc_color}{\rm $f_{cc}$} and inverse \textcolor{gcc_color}{\rm $g_{cc}(\theta)$}.]{\includegraphics[width=0.95\linewidth]{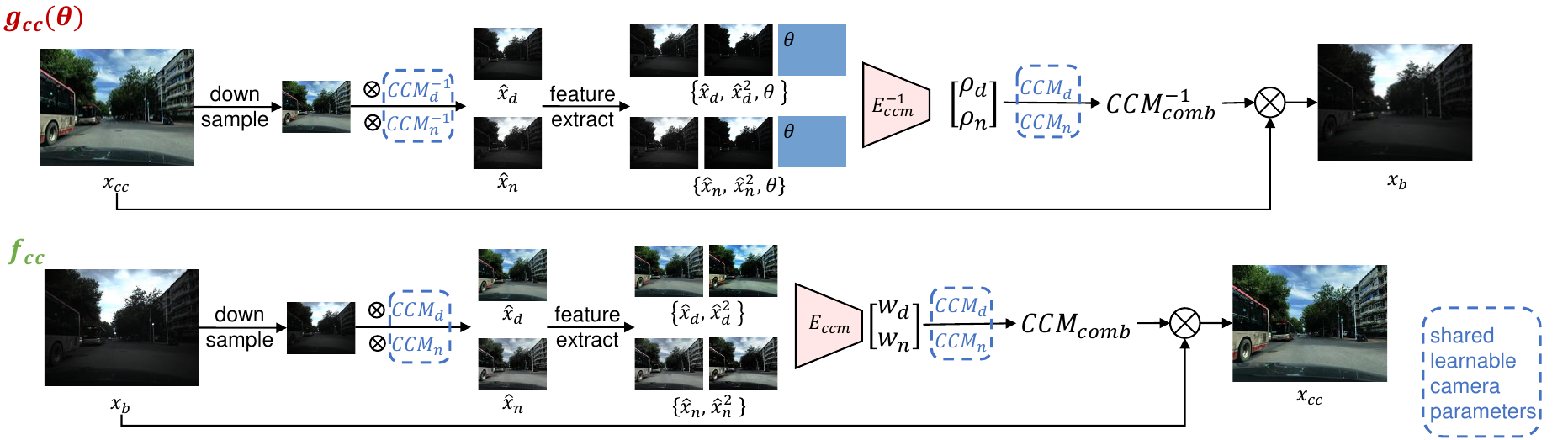} \label{fig:GAN-ISP-CC-Details}} 

     \caption{{\bf Unpaired CycleR2R.} (a) Our framework consists of an illumination estimation module (IEM) (Sec.~\ref{sec:iem}), modular unrolling based ISP/invISP (Sec.~\ref{sec:invISP}) and unsupervised loss functions (Sec.~\ref{sec:tranining_loss}). (b) The proposed framework simultaneously learns $g_{\rm invISP}$: RGB $\rightarrow$ RAW, and $f_{\rm ISP}$: RAW to $\rightarrow$ RGB with modular unrolling to best leverages the imaging physics in ISP/invISP for more robust model derivation. 
     % The forward ISP and is blue-shaded; while the inverse ISP is overlaid on the orange canvas. 
     (c) Color correction $f_{cc}$ in $f_{\textrm{ISP}}$ and its inverse process $g_{cc}(\theta)$ in $g_{\textrm{invISP}}$ are exemplified while other modules mostly follow the similar method.}
\label{fig:GAN_ISP}
\end{figure*}

This section presents details on how to easily simulate realistic RAW (simRAW) images from existing RGB image datasets.

\subsection{Problem Definition and CycleGAN Approach}
Existing work in~\cite{zamir2020cycleisp,abdelhamed2019noise,afifi2020cie} model the invISP process from RGB to RAW ($g:\mathcal{Y} \rightarrow \mathcal{X}$) as a one-to-one mapping in a supervised manner, for which a large number of paired RAW and RGB images (from the same camera) are required. Apparently, such a one-to-one mapping-based invISP does not reflect the imaging circumstances in practice. For example, the same scene may be acquired as two different RAW samples because of different illumination settings,  but the resultant RGB image (after the ISP) would be the same because the camera ISP is capable of making a proper estimation of those settings for realistic rendering.

The proposed Unpaired CycleR2R learns a one-to-many mapping of invISP through the introduction of the iillumination estimation module (IEM) that provides a distribution of color temperature $\theta$ and brightness $\phi$. Note that the use of unpaired RAW and RGB images avoids the collection of paired samples, making the solution more attractive and generally applicable in vast scenarios.
Figure~\ref{fig:GAN-ISP-Framework} provides an overview of the proposed Unpaired CycleR2R. We first use the IEM to estimate the illumination distribution of a scene. Then we sample a variety of $\theta$ and $\phi$ to simulate different illumination conditions used in practice (Sec.~\ref{sec:iem}). These $\theta$ and $\phi$ are then used to generate various simRAW samples for a given RGB input (Sec.~\ref{sec:invISP}). We use a set of loss functions to guide the generation of realistic simRAW images (Sec.~\ref{sec:tranining_loss}).

\subsection{Illumination Estimation Module (IEM)}
\label{sec:iem}
The same scene may appear differently in the RAW domain if different color temperature $\theta$ and brightness level $\phi$ are used in the acquisition. The ISP module estimates the $\theta$ and $\phi$ to generate properly-exposed RGB images under standard color temperature (D65 standard~\cite{ohta2006cie}).
This suggests the one-to-many mappings from RGB to RAW and unknown values of $\theta$ and $\phi$ used in the camera make the invISP an ill-posed problem. 

To tackle the ill-posed problem, we propose the IEM to estimate the original $\theta$ and $\phi$. We assume that the $\theta$ and $\phi$ follow the Gaussian distributions as $\mathcal{N} (\mu_{\theta},~\sigma_{\theta} )$ and $\mathcal{N} (\mu_{\phi}, ~\sigma_{\phi} )$~\cite{hollas2004modern}.
The IEM consists of {three} convolutional layers and two linear layers (see details in the supplemental material). During the training phase, the $\theta$ and $\phi$ are randomly sampled from Gaussian distribution and estimated through the guidance of $\mathcal{L}_{adv}$, $\mathcal{L}_{var}$ and $\mathcal{L}_{cycle}$ jointly to best simulate the real-world illumination.

\subsection{Modeling of the ISP (invISP) via Modular Unrolling}
\label{sec:invISP}
Although it is possible to build a black box neural network to represent the ISP/invISP functions, an interpretable network design is preferred for robust inference and wide generalization~\cite{conde2022model, AlgrithmUnrolling}. Therefore, we use modular unrolling to mimic the imaging physics involved in ISP and to build efficient $f_{\rm ISP}$/$g_{\rm invISP}$ used in the Unpaired CycleR2R framework.

Modern ISP systems in cameras are generally comprised of five major steps from RAW input to corresponding RGB output, as shown in Fig.~\ref{fig:GAN-ISP-Details}: demosaicing $f_{d}$, auto white balance $f_{wb}$, brightness adjustment $f_{b}$, color correction $f_{cc}$, and gamma correction $f_{g}$. Thus the ISP function can be expressed as:
\begin{align}
f_{\textrm{ISP}} &= f_{g} \circ f_{cc} \circ f_{b} \circ f_{wb} \circ f_{d},\\
\boldsymbol{y} &= f_{\textrm{ISP}} \circ \boldsymbol{x}, 
\end{align} where $\boldsymbol{y}\in \mathcal{Y}$ denotes an RGB image, $\boldsymbol{x}\in \mathcal{X}$ denotes a RAW image, and $\circ$ denotes the function composition operation. The invISP mirrors each step with illumination prior $\theta,~\phi$ as
\begin{align}
g_{\textrm{invISP}} & = g_{d} \circ g_{wb} ( \theta ) \circ g_{b} ( \phi ) \circ g_{cc} (\theta ) \circ g_{g}, \\
\boldsymbol{x} &= g_{\textrm{invISP}} \circ \boldsymbol{y},
\end{align} 
where we assume $g=f^{-1}$, for simplicity.

\subsubsection{Demosaicing}

Color (RGB) pixels on an image sensor are typically arranged in a Bayer pattern, half green, one-quarter red, and one-quarter blue (also called RGGB). 
To obtain a full-resolution color image, 
various demosaicing algorithms~\cite{li2008image} have been developed in the past. For simplicity, we bilinearly upsample same-color pixels in close proximity to obtain demosaiced image $\boldsymbol{x}_{demo}$. Similarly, in invISP, we reverse the process to mosaic $\boldsymbol{x}_{demo}$ for its RAW output as
\begin{align}
    \boldsymbol{x}_{demo} = f_{d}\circ \boldsymbol{x}, \mbox{~~~}
    \boldsymbol{x} = g_{d} \circ \boldsymbol{x}_{demo}.  \label{eq:demosaic}
\end{align}

\subsubsection{Auto White Balance (AWB)}
{ Cameras apply the white balance to ensure color constancy, which requires accurate approximation of the color temperature. Practical solutions often achieve this by augmenting the digital gains in Red and Blue channels~\cite{Arjan_ComputationalColorConstancy,brooks2019unprocessing}. For instance, various gain presets, $ \{(R_{gain}, B_{gain})\}$ = $\{ (r_{1}, ~b_{1}), \ldots,  (r_N, b_N) \}$, can be defined for specific color temperatures. These presets are linearly weighted for auto
white balance (AWB) because ambient illumination in real-life scenarios often mixes radiance from different light sources as
\begin{align}
    ( r_{gain}, b_{gain}) &= 
    W\cdot\{(R_{gain}, ~B_{gain})\}^T \nonumber\\
    & = (\sum_{i=1}^N w_i r_i, ~\sum_{i=1}^N w_i b_i ), \label{eq:white_balancing}
\end{align} where $W=\{w_1, \ldots, w_N\}$ is the weighting vector, and $N$ is total number of gain presets. 
As seen,  accurate AWB relies on the proper choice of the $w_i$, $r_i$ and $b_i$ in $f_{wb}$ to derive $[r_{gain}, 1, b_{gain}]$ which is then multiplied with every pixel in the R, G, and B channels. }

In general, the AWB function $f_{wb}$ in ISP maps the demosaiced image $\boldsymbol{x}_{demo}$ into a white-balanced image $\boldsymbol{x}_{wb}$. Given that AWB adjusts the global appearance of the image, we downscale the native input $\boldsymbol{x}_{demo}$ to $\boldsymbol{x}^{\downarrow}_{demo}$ at a size of 128$\times$128$\times$3 for processing, with which we can significantly reduce the space and time complexity. Specifically,
\begin{itemize}
    \item Starting with $\boldsymbol{x}^{\downarrow}_{demo}$, we generate a pre-AWB image $\hat{\boldsymbol{x}}_{wb_i}$ by multiplying every channel with a preset gain $[r_i,1,b_i]$. In training, we randomly initialize $r_i$, and $b_i$ following ~\cite{brooks2019unprocessing}. As suggested in~\cite{DBLP:conf/iccv/AfifiB19,Finlayson_ColorCorrection} for camera AWB or color correction, $\hat{\boldsymbol{x}}_{wb_i}$ is paired with $\hat{\boldsymbol{x}}^2_{wb_i}$ for learning. 
    \item Then, we propose the $E_{wb}( \cdot )$ that shares the same architecture with IEM, but with one-channel output, to process $\{\hat{\boldsymbol{x}}_{wb_i}, ~\hat{\boldsymbol{x}}^2_{wb_i}  \}$ for weight derivation as
\begin{align}\label{eq:w_i}
    w_i =  E_{wb} ( \{\hat{\boldsymbol{x}}_{wb_i}, ~\hat{\boldsymbol{x}}^2_{wb_i}  \} ),
\end{align} and  subsequently the final AWB gain as in \eqref{eq:white_balancing}.
\item Finally, instead of multiplying the AWB gain with every pixel of $\boldsymbol{x}_{demo}$ directly, highlight-preserving transformation $S(\boldsymbol{x}, ~\texttt{scalingFactor})$ is applied to avoid  highlight overflow~\cite{brooks2019unprocessing} as
\begin{align}
\label{equation:wb}
    &\boldsymbol{x}_{wb} = f_{wb} \circ \boldsymbol{x}_{demo} = S(\boldsymbol{x}_{demo}, ~[ r_{gain}, 1, b_{gain}]  ).
\end{align}
\end{itemize}

Correspondingly, for $g_{wb}$ in invISP, we reverse engineer the $f_{wb}$ to derive $[1/r_{gain}, ~1, ~1/b_{gain}]$. As mentioned before, the original color temperature is unknown in $g_{wb}$.
Therefore, we model the inverse weights $\{ \rho_1, \ldots, \rho_N\}$ using $E^{-1}_{wb}$ with the  color temperature prior $\theta$ estimated by the IEM.
Note that $E^{-1}_{wb}(\cdot)$ shares the same architecture with $E_{wb}(\cdot)$. The processing steps are as follows. 
\begin{itemize}
    \item  Apply preset inverse gain on downscaled input $\boldsymbol{x}_{wb}^{\downarrow}$ as $\hat{\boldsymbol{x}}_{demo_i} =  [ 1/r_{i}, ~1, ~1/b_{i} ] \cdot \boldsymbol{x}_{wb}^{\downarrow}$.
    \item Derive the weight and inverse AWB gain as
    \begin{align}
    \rho_i  &=  E^{-1}_{wb} ( \{\hat{\boldsymbol{x}}_{demo_i}, ~\hat{\boldsymbol{x}}^2_{demo_i}, ~\theta  \} ),
    \end{align} 
    and compute $( 1/r_{gain}, ~1/b_{gain}) = \sum_i \rho_i \cdot ( 1/r_{i}, ~1/b_{i} )$.
    \item  Generate $\boldsymbol{x}_{demo}$ as 
    \begin{equation}
   \label{equation:inv_wb}
   \boldsymbol{x}_{demo} = g_{wb} (\theta) \circ \boldsymbol{x}_{wb}  = [ \frac{1}{r_{gain}}, 1, \frac{1}{b_{gain}}] \cdot \boldsymbol{x}_{wb}.
    \end{equation}
\end{itemize}
Such derivations of $w_i$ and $\rho_i$ are also used in brightness adjustment and color correction to characterize the non-bijective mapping. 
 
\subsubsection{Brightness Adjustment (BA)}
Existing ISPs usually adjust the brightness of overexposed or underexposed RAW images by enforcing range-limited global gain  $b_{gain}$ to scale every pixel uniformly~\cite{zhou2014global}.
We use a neural network $E_{b}( \cdot )$, which shares the same architecture as $E_{wb}( \cdot )$, to derive $b_{gain}$. 
We use a downscale and grayscale version of $\boldsymbol{x}_{wb}$, denoted as $ \boldsymbol{x}_{wb}^{\downarrow G}$, because brightness adjustment is a global operation that does not require full-resolution spatial and spectral details. 

We compute $b_{gain} = {\beta + \alpha \cdot \textrm{tanh}( E_{b}( \boldsymbol{x}_{wb}^{\downarrow G} ) )}$,
in the range of $[\beta-\alpha, \beta+\alpha ]$ with $\alpha$ = 0.3 and $\beta$ = 1, following the suggestions in~\cite{brooks2019unprocessing}.
Finally, we also apply highlight-preserving transformation $S(\cdot, ~\cdot )$ used in ~\cite{brooks2019unprocessing} to have $\boldsymbol{x}_{b}$:
\begin{align}
\label{equation:brightness}
    \boldsymbol{x}_{b} = f_{b} \circ \boldsymbol{x}_{wb} = S(\boldsymbol{x}_{wb}, b_{gain} ).
\end{align}

For $g_{b}$ in invISP, we reverse the adjustment on well-exposed $\boldsymbol{x}_{b}$ using $1/b_{gain}$. Considering that a series of images captured in the same scene using different exposure levels could be rectified to the same well-exposed image after brightness adjustment, we introduce the brightness prior $\phi$ to recover $\boldsymbol{x}_{wb}$ as
\begin{align}
    b_{gain} = \beta + \alpha \cdot \textrm{tanh} ( E^{-1}_{b} ( \{  \boldsymbol{x}_b^{\downarrow G}, \phi \} ) ).
\end{align} 
 We multiply $1/b_{gain}$ with $\boldsymbol{x}_{b}$ to generate $\boldsymbol{x}_{wb}$ as
\begin{align}
\label{equation:inv_brightness}
    \boldsymbol{x}_{wb} = g_{b} \circ \boldsymbol{x}_b = 1/b_{gain} \cdot \boldsymbol{x}_{b}.
\end{align}
% \Salnote{Loud thinking here. This type of equation is elementary (like scalar multiplication). Do we really need to add it in a JOURNAL paper? Do we refer to this anywhere?}

\subsubsection{Color Correction (CC)}
In practice, a $3\times3$ color correction matrix (CCM) is used in camera ISPs to restore the colors of the acquired image to match the human perception~\cite{afifi2020cie}.
Similar to the AWB, camera vendors usually preset $\textrm{CCM}_{d}$ and $\textrm{CCM}_{n}$ for daytime and nighttime conversion, respectively~\cite{wolf2003color}. The final transformation $f_{cc}$ is often derived by linearly combining the presets as
\begin{equation}
    \textrm{CCM}_{mix} = \omega_d \cdot \textrm{CCM}_d + \omega_n \cdot \textrm{CCM}_n. \label{eq:ccm_mix}
\end{equation} 
As shown in Fig.~\ref{fig:GAN-ISP-CC-Details}, this work relies on a neural network $E_{cc}( \cdot )$ to produce respective $\omega_d$ and $\omega_n$. 
We compute $\omega_d$ as 
 \begin{align}
     \omega_{d} =  E_{cc} ( \{\hat{\boldsymbol{x}}_{d}, ~\hat{\boldsymbol{x}}^2_{d}  \} ),
     \end{align}
where $\hat{\boldsymbol{x}}_d$ is computed by multiplying daytime CCM preset with every color pixel in the down-sampled $\boldsymbol{x}_{b}^{\downarrow}$ of size $128\times 128 \times 3$ as $\hat{\boldsymbol{x}}_d= \textrm{CCM}_d \cdot \boldsymbol{x}_{b}^{\downarrow}$. The same procedure is applied to generate $\hat{\boldsymbol{x}}_n$, $\hat{\boldsymbol{x}}_n^2$ and compute $\omega_{n}$. 
{During the training process, both $\textrm{CCM}_d$ and $\textrm{CCM}_n$ are randomly initialized following the same setting defined for $r_i$ and $b_i$ in $f_{wb}$.} 

Then \eqref{eq:ccm_mix} is used to compute the $\textrm{CCM}_{mix}$ as 
\begin{align}
\label{equation:ccm}
    \boldsymbol{x}_{cc} = f_{cc} \circ \boldsymbol{x}_{b} = \textrm{CCM}_{mix} \cdot \boldsymbol{x}_{b}. 
\end{align}

\begin{comment}
\begin{align}
    \{ \hat{\boldsymbol{x}}_d, ~\hat{\boldsymbol{x}}_n \} &= \{ \textrm{CCM}_d, ~\textrm{CCM}_n \} \cdot \boldsymbol{x}_b^{\downarrow} \\
    %  x_{d} & =  x_{b} \cdot CCM_{d} \\
    %  x_{n} & =  x_{b} \cdot CCM_{n} \\
\begin{split}
     w_{d} &=  E_{cc} ( \{\hat{\boldsymbol{x}}_{d}, \hat{\boldsymbol{x}}^2_{d}  \} ) \\
     w_{n} &=  E_{cc} ( \{\hat{\boldsymbol{x}}_{n}, \hat{\boldsymbol{x}}^2_{n}  \} )
\end{split}
 \\
     \textrm{CCM}_{comb} &= w_{d} \cdot \textrm{CCM}_{d} + w_{n} \cdot \textrm{CCM}_{n} \\
    %  \boldsymbol{x}_{cc} = 
     f_{cc}( \boldsymbol{x}_b ) &=  \textrm{CCM}_{comb} \cdot \boldsymbol{x}_b
\end{align}
\end{comment}

The same $\boldsymbol{x}_{cc}$ may be produced by different $\boldsymbol{x}_{b}$ and different CCM scaling. Thus, following the discussions for $g_{wb}$, we use $E_{cc}^{-1}(\cdot)$ with color temperature prior $\theta$ to derive inverse weights $\rho_d$ and $\rho_n$ of $g_{cc}$,
as $ \rho_d = E^{-1}_{cc} ( \{ \hat{\boldsymbol{x}}_{d}, ~\hat{\boldsymbol{x}}_{d}^2,~ \theta\})$ and $ \rho_n = E^{-1}_{cc} ( \{ \hat{\boldsymbol{x}}_{n}, \hat{\boldsymbol{x}}_{n}^2, ~\theta \})$.
Thereafter we can easily obtain $\textrm{CCM}_{mix}^{-1}$ as
\begin{align}
\textrm{CCM}_{mix}^{-1}=( \rho_d \cdot \textrm{CCM}_d + \rho_n \cdot \textrm{CCM}_n )^{-1}.
\end{align} 
Finally, we  have
\begin{align}
\label{equation:inv_ccm}
    \boldsymbol{x}_{b} = g_{cc} \circ \boldsymbol{x}_{cc} = \textrm{CCM}_{mix}^{-1} \cdot \boldsymbol{x}_{cc}.
\end{align}

\subsubsection{Gamma Correction}
Gamma correction is used to match the non-linear characteristics of a display device or human perception~\cite{farid2001blind}. 
We adopt the correction function recommended in ITU-R BT.~709 standard~\cite{stokes1996standard}, noted as $f_{g}$, which is widely used in commodity ISPs today~\cite{drago2003adaptive}. Additional details are provided in the supplementary material.

\subsection{Training Loss}
\label{sec:tranining_loss}
We use three loss functions, denoted as $\mathcal{L}_{adv}$, $\mathcal{L}_{cycle}$ and $\mathcal{L}_{var}$, to train the Unpaired CycleR2R. The overall loss used to train  our Unpaired CycleR2R can be written as
\begin{equation}
    \mathcal{L} = \mathcal{L}_{adv} + \mathcal{L}_{cycle} + \mathcal{L}_{var}.
\end{equation}

First, a discriminator $D_\textrm{RAW}$ is applied to measure the similarity between generated and real images. The discriminator can be further decomposed as $D_\textrm{color}$ and $D_\textrm{bright}$, where $D_\textrm{color}$ discriminates color discrepancy using 2D log-chroma histogram~\cite{barron2017fast} and $ D_\textrm{bright}$ discriminates brightness discrepancy using 1D grayscale histogram. 
$D_\textrm{color}$ stacks five convolutional layers with Leaky ReLU~\cite{xu2015empirical} and $D_\textrm{bright}$ uses five linear layers. The outputs of $D_\textrm{color}$ and $D_\textrm{bright}$ are added together as the output of $D_\textrm{RAW}$. We update the parameters of $g_\textrm{invISP}$ and $D_{\textrm{RAW}}$ by minimizing the adversarial loss $\mathcal{L}_{adv}$ given as 
\begin{align}
    \mathcal{L}^{G}_{adv} &= \Vert 1 - D_\textrm{RAW}( g_\textrm{invISP}({\boldsymbol{y}}) )  \Vert_2, \\
    \mathcal{L}^{D}_{adv} &=  \Vert 1 - D_\textrm{RAW}( {\boldsymbol{x}}  ) \Vert_2 + \Vert D_\textrm{RAW} ( g_{\rm invISP}({\boldsymbol{y}})  ) \Vert_2,
    \\
    \mathcal{L}_{adv} &= \mathcal{L}^G_{adv} + \mathcal{L}^D_{adv}. 
\end{align}

A cycle-consistency loss $\mathcal{L}_{cycle}$ is used to indirectly optimize $\theta$ and $\phi$ considering the assumption that RAW images captured under all possible illumination settings of the scene shall be converted into  the same RGB sample. Thus the reconstructed RGB from the simRAW, given as $\overline{\boldsymbol{y}} = f_{\rm ISP}\circ g_{\rm invISP}\circ \boldsymbol{y}$,  should be as close to the original RGB as possible, which gives us the following loss: 
\begin{equation}
    \mathcal{L}_{cycle} = \Vert \overline{\boldsymbol{y}} - \boldsymbol{y} \Vert_{1}. 
\end{equation}

The cycle-consistency constraint often leads to the one-to-one mapping of $g_{\rm invISP}$ in optimization~\cite{park2020contrastive}.
To assure the one-to-many mapping of  $g_{\rm invISP}$ in practice,   another variant loss $\mathcal{L}_{var}$ is used  with which  we wish to enlarge the distance between two simRAW images $\boldsymbol{x}_1$ and $\boldsymbol{x}_2$ under two different illumination settings: $(\theta_1, ~\phi_1)$ and $(\theta_2, ~\phi_2)$. To independently evaluate the distance of color and brightness attributes, we measure the loss $\mathcal{L}_{var}$ in YUV space ($\boldsymbol{x}_1 \rightarrow \{\textrm{y}_1, ~\textrm{u}_1, ~\textrm{v}_1\}$, $\boldsymbol{x}_2 \rightarrow \{\textrm{y}_2, ~\textrm{u}_2, ~\textrm{v}_2\}$) as  
\begin{align}
    \mathcal{L}_{var} &= -\frac{\Vert \{ \textrm{u}_1, \textrm{v}_1 \} -  \{ \textrm{u}_2, \textrm{v}_2 \} \Vert_2}{\Vert \theta_1 - \theta_2 \Vert_2} -
    \frac{\Vert\textrm{y}_1 - \textrm{y}_2\Vert_2}{\Vert \phi_1 - \phi_2 \Vert_2}.
\end{align}
\section{RAW-domain Task Execution} \label{sec:rho_CV}
\begin{figure*}
% [t]
\centering
\subfloat[The impact on Conv layer.]{\includegraphics[width=0.98\textwidth]{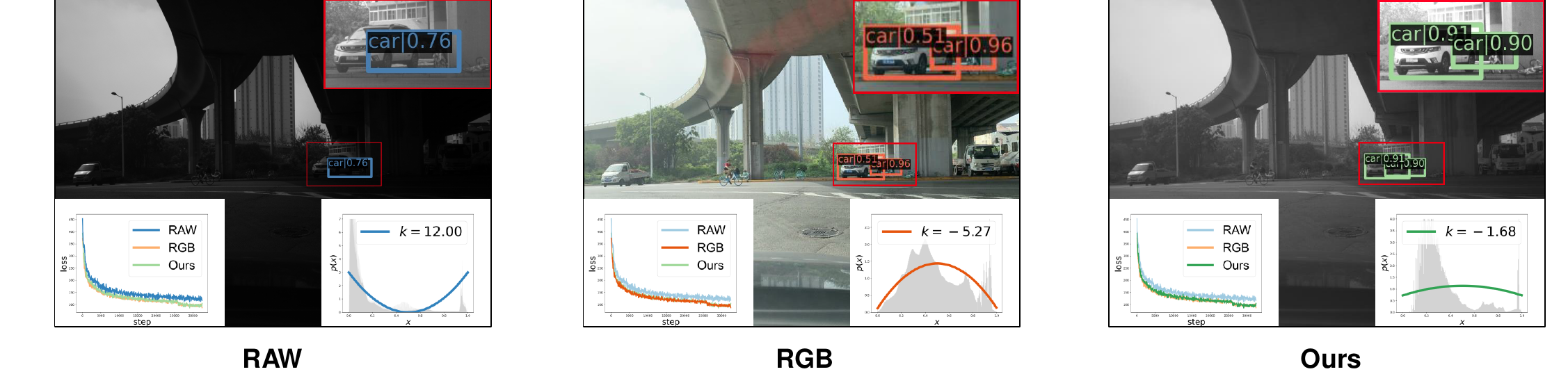} \label{fig:RAW-CV-Conv-Impact}}

\subfloat[The impact on BN layer.]{\includegraphics[width=0.98\textwidth]{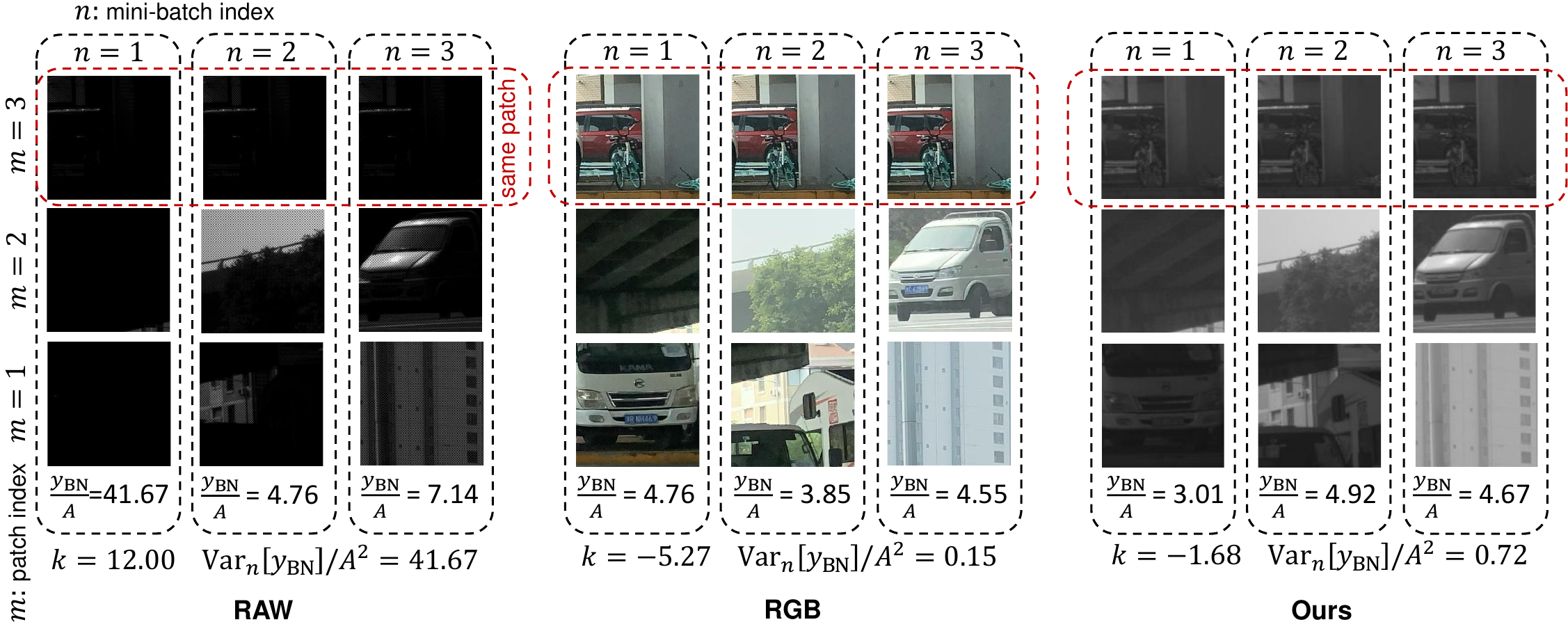}\label{fig:RAW-CV-BN-Impact}}

\caption{\textbf{Distribution impact of samples in native RAW, RGB, and Gamma-corrected (Ours) spaces.}
YOLOv3 is refined specifically for each domain space to quantify the distribution impact.
% \zhihao{
(a) The fitted histogram distribution using~\eqref{eq:k_quard} is shown in the lower right corner, where the RAW image has the largest value of $k$. As can be seen from the training loss plotted in the lower left corner, the convergence speed of the RAW image is the slowest, leading to the missing of critical objects during detection; (b)
The variance of features across mini-batches with the same input patch is negatively correlated to the convergence of a neural network~\cite{liu2021adadm}. For the investigation of feature variance after a BN layer, $\Var_n[\boldsymbol{y}_{\tt BN}]$, patches with index $m$ from mini-batches with index $n$ are randomly sampled in each domain. As seen, a large $k$ also leads to imbalanced variance across mini-batches, causing an unstable training process. The $\boldsymbol{y}_{\tt BN}$ is calculated using~\eqref{eq:y_bn}}

% Content distribution of images in different domains is modeled using~\eqref{eq:k_quard}. 
% {(a)}  (b) Large $k$ also leads to unstable batch normalization with imbalanced variance across mini-batches that have same patch sample used in training. \Salnote{This caption is not very informative. You just straight into $k$ without explaining what are the different figures. I do not understand what is happening in the figure. ALSO please do not capitalize random words. Avoid $\left( \right)$ when you can just use $()$.}
\label{fig:RAW_CV}
\end{figure*}

The generated simRAW images can be used to train RAW-domain models for various tasks. {We discuss high-level object detection and low-level image compression tasks, for which we refine well-known models originally developed for RGB images. } 

\subsection{High-Level Object Detection}
Object detectors~\cite{yolo,lin2017feature,chen2021you,girshick2015fast, ren2015faster} have achieved great success in detecting objects in RGB images. The performance of off-the-shelf RGB-domain models such as YOLOv3\footnote{We use YOLOv3 because it is vastly used in products; but our method can be easily extended to other object detectors.} Fig.~\ref{fig:RAW-CV-Conv-Impact} presents an example where object is not detected on RAW image. The ``Naive Baseline'' in Table~\ref{table:without_labeled_detection},~\ref{table:ablation_hdr_split} also show a sharp loss in performance. This is counter-intuitive because sensor RAW data with a larger dynamic range shall contain more information on the physical scene than its ISP-processed RGB sample.  

We argue that the vastly diverse distribution of RAW images significantly undermines the representation capacity of DNNs originally trained using RGB samples. We first use analytical approximation to show the impact of RAW data distribution on both convolution (Conv) and batch normalization (BN) layers commonly adopted in learned detectors~\cite{yolo, wu2018group}; and then apply the gamma correction to regularize RAW image distribution for RAW-domain task inference. 

\subsubsection{Distribution Analysis of RAW Images}
\label{sec:raw_detect_arch}

\textbf{Distribution approximation using patches.}
\label{sec:model_distribution}
In practice, an input image $\boldsymbol{x}$ is often divided into non-overlapping small patches $\mathcal{P} = \{ \boldsymbol{P}_1,\boldsymbol{P}_2, \ldots \}$ to train desired models for task inference~\cite{yolo, wang2020deep}. For a patch $\boldsymbol{P} \in \mathcal{P}$, its histogram can be approximated using a Gaussian distribution $\mathcal{N}(\mu, ~\sigma^2 )$. To simplify modeling the distribution of pixel intensity $p(\vx)$ over the entire image using patches, we treat each patch $\boldsymbol{P}$ as a superpixel with intensity $\mu$ and assume that the $p(\vx)$ can be approximated by $p(\mu)$.

% To model the distribution of pixel intensity $p(\vx), \vx \in \boldsymbol{x}$ over the entire image, we treat each patch $\boldsymbol{P}$ as a superpixel with the intensity $\mu$ inferred by the mean of this patch, and assume $p(\vx) \approx p( \mu )$. \Salnote{I do not understand what you mean by $p(\vx)$ here. Is this distribution of pixel intensity over the entire image? If so, what does $p(\vx) \approx p(\mu)$ mean? }

Referring to the RAW snapshot depicted in the leftmost subplot of Fig.~\ref{fig:RAW-CV-Conv-Impact},
most pixel patches are clustered in dark and bright regions, which yield histogram peaks near 0 and 1\footnote{Sensor RAW images are normalized to the range of [0,1] for processing.}. Therefore, $p(\mu)$ can be possibly hypothesized using a U-shaped function.
In the meantime, without losing generality for image with proper exposure control, the mean of the entire image shall be close to the middle level of the dynamic range, which, in other words, for normalized RAW image, $\mathbb{E}[p(\mu)]$ = 0.5. It then leads us to approximate $p(\mu)$ using a quadratic function:
\begin{equation}
    p ( \vx ) 
    \approx p ( \mu )
    \approx k \mu^2 - k \mu + \frac{k}{6} + 1, \label{eq:k_quard}
\end{equation}
with $0 < k \leq 12$ to guarantee $\min\{p(\mu)\}\geq0$. Coincidentally, \eqref{eq:k_quard} also models the histogram of normalized RGB image very well but has the $k<0$. As seen, $k$ can be used to characterize the distribution of the input image (RAW or RGB). We next show that $k$ impacts the performance of Conv and BN layers analytically, which in turn, explains why native YOLOv3 trained for RGB images cannot be directly used to process RAW samples.

\noindent \textbf{Effect of $k$ on convergence.}
Suppose $\boldsymbol{w} \in \mathbb{R}^{S\times S}$ denotes the Conv kernel weights randomly initialized with $\mathcal{U}( -\eta, ~\eta ), ~\eta \to 0$, and $\vb$ is the Conv bias that is randomly initialized in same manner as the weights. Then a single layer of CNN can be represented as 
\begin{equation}
    \boldsymbol{y} = \boldsymbol{w} \ast 
    \tilde{\boldsymbol{x}} + \vb,
\end{equation}
where $\tilde{\boldsymbol{x}} = \boldsymbol{x}-0.5$ shifts the center to zero. As for the bounding box regression of the input patch $\boldsymbol{P}\sim \mathcal{N}(\mu, \sigma^2)$ in YOLOv3~\cite{yolo}, the loss function $\mathcal{L}$ is
\begin{equation}
    \mathcal{L} = \frac{1}{H\times W} ( \boldsymbol{w} \ast ( \boldsymbol{P}-0.5 ) + \mathbf{b} - \hat{\vy} )^2,
\end{equation} having $\hat{\vy}$ as the ground truth label, and $H$/$W$ as the height/width of $\boldsymbol{P}$.

% \Salnote{What is the need for defining a new symbol $\vw$ here that is a sample from $\boldsymbol{w}$??? Not a good set of notations.}\zhihao{Fixed}

For each training iteration, the weight $w \in \boldsymbol{w}$ is updated with learning rate $\alpha$, i.e.,
$ w \gets w - \alpha \frac{\partial \mathcal{L}}{\partial w}. $ 
The stability of parameter update is directly related to the variance of the gradient $\Var[\frac{\partial \mathcal{L}}{\partial w}]$ as shown in~\cite{wu2018group}. This variance is approximately given as 
\begin{align}
    \Var[ \frac{\partial \mathcal{L}}{\partial w} ]
    &\approx  C^2 \Var[ \tilde{\mu}^2 ] + 
    D^2 \Var[ \tilde{\mu} ] + 
    \textrm{const},  \label{eq:var_gradient_weights}
\end{align}
where $\tilde{\mu} = \mu-0.5$, $C = 2 \sum w = 2 S^2 \mathbb{E}[ w ]$, and $D$ is a constant. Since $w$ is randomly initialized with a small value close to zero, $\mathbb{E}[ w ]\approx 0$, which simplifies \eqref{eq:var_gradient_weights} to:
\begin{align}
    \begin{split}
    \Var[ \frac{\partial \mathcal{L}}{\partial w} ]
    &\approx 
    D^2 \Var[ \tilde{\mu} ] + 
    \textrm{const}
    =
    \frac{D^2}{180} k  + \textrm{const}.
    \end{split} \label{eq:simp_var_gradient_weights}
\end{align}
This shows that the variance of the gradient is directly related to the parameter $k$. A larger $k$ provides larger $\Var[ \frac{\partial \mathcal{L}}{\partial w} ]$, making the convergence of the Conv weights more difficult and the CNN model unreliable. The proofs of \eqref{eq:var_gradient_weights} and \eqref{eq:simp_var_gradient_weights} are provided in the Supplemental Material.

\noindent\textbf{Effect of $k$ on batch normalization.}
\label{sec:impact_on_bn_layer}
\begin{figure}
     \centering
     \includegraphics[width=0.9\linewidth]{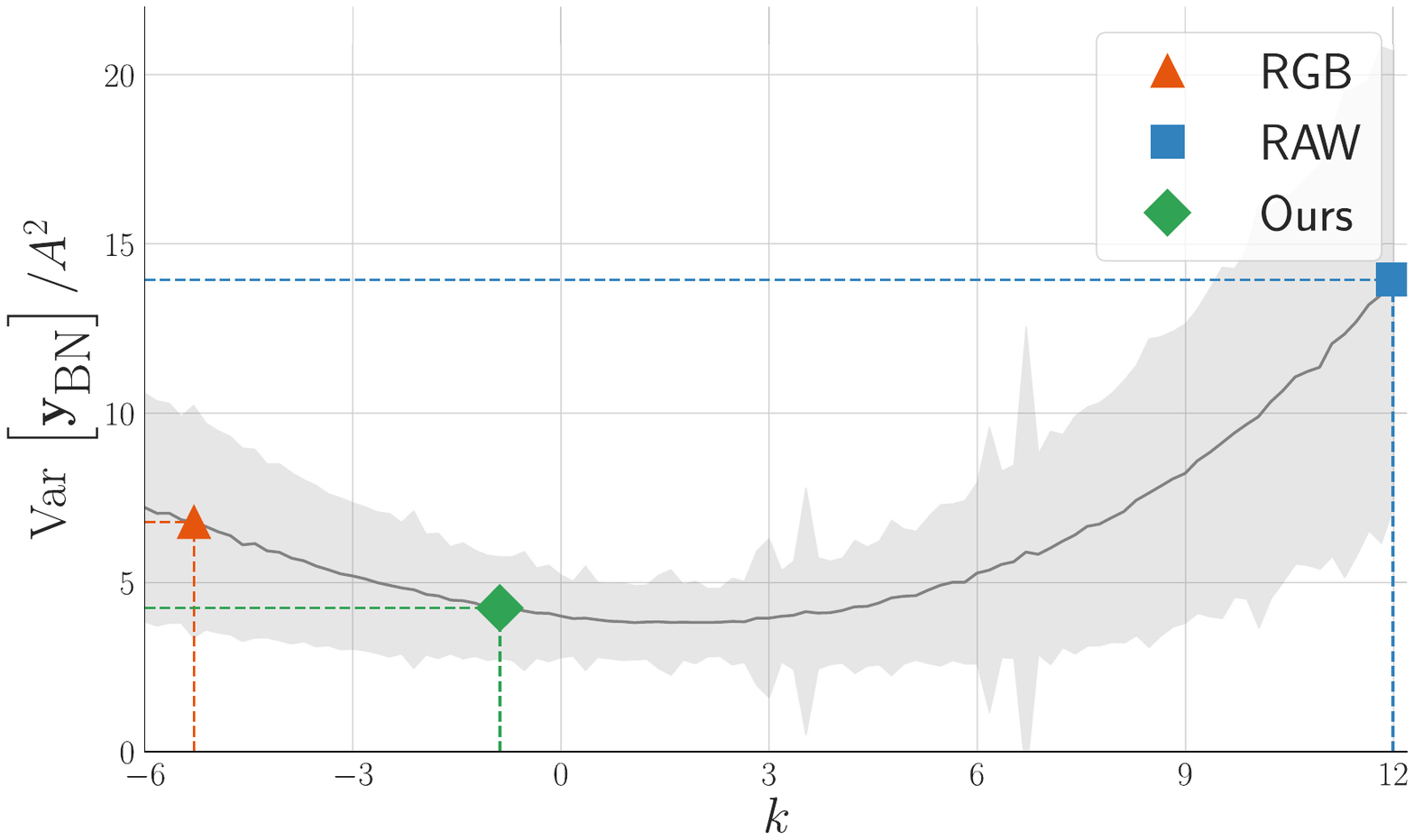} 
     \caption{{ $\Var_n\left[ \boldsymbol{y}_{\tt {BN}} \right]/A^2$ {\bf vs.} $k$.}}
     \label{fig:VAR_Y_K}
\end{figure}

Batch Normalization~\cite{ioffe2015batch} (BN) randomly splits a batch of training samples into mini-batches during training iteration to assure stability. A BN layer is usually placed after a convolutional layer in order to normalize the output features. Given an input patch with index $m$ from a mini-batch with index $n$ of size $M$, the output of the BN layer is given by:
\begin{align}
\boldsymbol{y}_{\tt BN} = \textrm{BN} ( \boldsymbol{y}^{\ell m n} ) = \gamma \frac{\boldsymbol{y}^{\ell m n} - \mathbb{E}_m[ \boldsymbol{y}^{\ell m n} ]} {\sqrt{ \Var_m[ \boldsymbol{y}^{\ell m n} ]}} + \beta,
\end{align}
where $\gamma$ and $\beta$ are learnable parameters, and $\boldsymbol{y}^{\ell m n}$ is the feature of layer $\ell$.

It is vital for the convergence of a neural network that the mean and variance of features with the same input patch are similar across mini-batches during training~\cite{liu2021adadm}. Thus, we can measure the convergence speed of a model by the cross-batch variance $\Var_n[\boldsymbol{y}_{\tt BN}]$. The larger $\Var_n[\boldsymbol{y}_{\tt BN}]$ leads to the imbalance among mini-batches and thus yields an unstable training gradient.
In the following part, we will demonstrate that $\Var_n[\boldsymbol{y}_{\tt BN}]$ increases significantly for RAW images having larger $k$, which degrades the training stability of the RAW-domain detector.

We will start by modeling the relationship between the input patch $\boldsymbol{P}^{m n}$ and $\boldsymbol{y}_{\tt BN}$. As proven in~\cite{liu2021delving}, we have that $\mathbb{E}_m [ \boldsymbol{y}^{\ell m n}]$ is approximately 0 and its variance is approximately $\alpha_l \cdot \Var_m [ \boldsymbol{y}^{\ell-1 m n} ]$, where $\alpha_l$ is a constant when the weight of layer $l$ is fixed. Therefore, the relationship can be recursively derived as follows:
\begin{align}
\boldsymbol{y}_{\tt BN} \approx \frac{A}{ \sqrt{\Var_m[ \boldsymbol{P}^{m n} - 0.5 ]}} + \beta, \label{eq:y_bn}
\end{align}
where $A = \gamma\cdot \boldsymbol{y}^{\ell m n} / \prod_{i=1}^{\ell} \sqrt{\alpha_i}$ is a constant when the input patch $\boldsymbol{P}^{m n}$ is fixed.

Next, we will model the variance $\Var_n[ \boldsymbol{y}_{\tt {BN}}]$ across mini-batches. Given ~\eqref{eq:y_bn}, we simplify the variance as:
\begin{align}
\Var_n[ \boldsymbol{y}_{\tt BN} ]
 \approx A^2 \cdot \mathop{\Var}_n[ \frac{1}{
\sqrt{ \mathop{\Var}_m
[ \boldsymbol{P}^{m n}-0.5 ]}} ].
\label{eq:var_n_ybn}
\end{align}
Due to the limited batch size $M$, it is difficult to obtain a closed-form probability distribution of $\boldsymbol{P}^{m n}$. Therefore, we resort to sampling simulations for approximation. To simplify the problem, we neglect the variance in a patch and represent it with its mean value $\mu$. We obtain the value of $\Var_n[ \boldsymbol{y}_{\tt BN} ]/A^2$ using $5000$ randomly-sampled mini-batches that contain the same patch $\boldsymbol{P}^{m n}$ characterized by $p(\mu )$ in \eqref{eq:k_quard}. The whole process is repeated $5000$ times using different randomly-sampled $\boldsymbol{P}^{m n}$ to obtain the average value of $\Var_n[\boldsymbol{y}_{\tt {BN}} ]/A^2$.
Figure~\ref{fig:VAR_Y_K} shows the quantitative approximation between $\Var_n[ \boldsymbol{y}_{\tt {BN}} ]/A^2$ and $k$. It is clear that $\Var_n[ \boldsymbol{y}_{\tt BN} ]$ increases significantly for larger values of $k$.

\textbf{Remark.} 
As seen, larger $k$ induced by the RAW input slows the convergence of convolutional weights and yields unstable batch normalization, making the underlying model unreliable and inefficient, which requires us to regularize the distribution of RAW images for robust performance.

\definecolor{BlueLine}{RGB}{115, 149, 211}
\definecolor{OrangeLine}{RGB}{237, 125, 49}

\begin{figure*}
% [t]
     \centering
     \subfloat[RAW Image Compression (RIC) Architecture]{\includegraphics[width=0.99\linewidth]{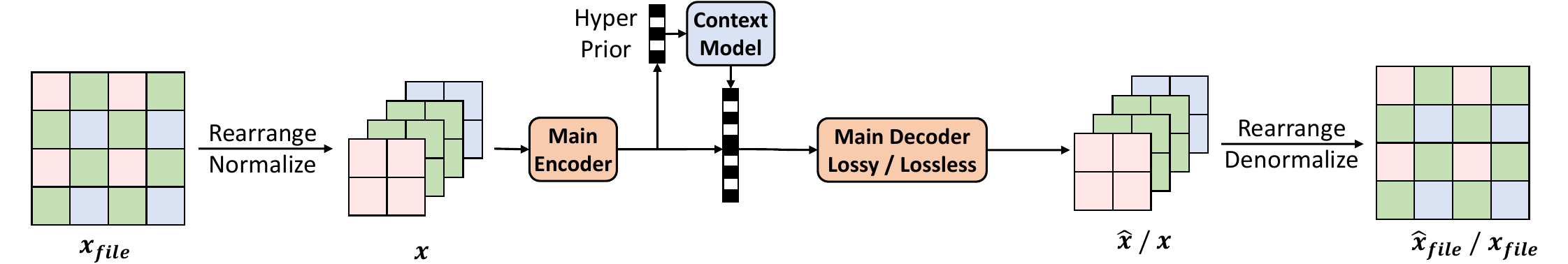}  \label{fig:lossy_lossless_arch}}
     
     \subfloat[Lossless RIC Decoder]{\includegraphics[width=0.99\linewidth]{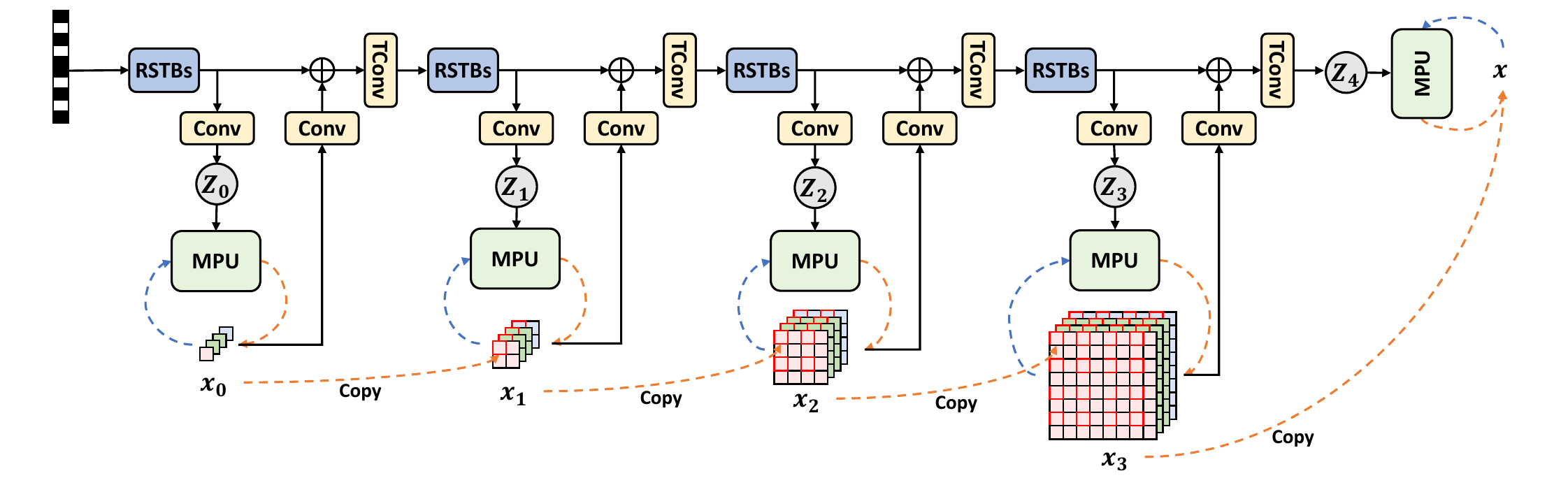}  \label{fig:lossless_RIC_decoder}}
     
     \subfloat[Multichannel Processing Unit (MPU)]{\includegraphics[width=0.98\linewidth]{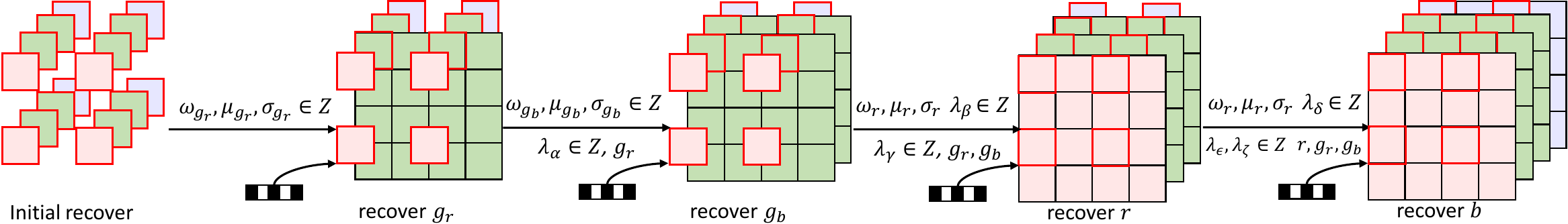}  \label{fig:lossless_RIC_details}}
     
     \caption{ {\bf RAW Image Compression (RIC).} (a) Similar TinyLIC architectures are used for both lossy and lossless RIC. More details about the lossy pipeline are in~\cite{lu2022high}.
     (b) The detailed architecture of the lossless decoder.  $\boldsymbol{Z}_i$, $i\in[0,4]$ aggregates low-resolution information to characterize the logistic distribution for element probability derivation. The probability is used for both encoding ({\color{BlueLine}{blue}}) and decoding ({\color{OrangeLine}{orange}}). In decoding, $\boldsymbol{x}$ are gradually recovered with progressive decoding; (c) Multichannel Processing Unit (MPU) is devised to process the element in $\boldsymbol{x}_i$ with the probability derived from the $\boldsymbol{Z}_i$ under a predefined order, \eg, $g_r\rightarrow g_b\rightarrow r \rightarrow b$. Arithmetic decoding is omitted. Residual Swin Transformer Blocks (RSTBs) are stacked to aggregate and embed necessary information. The Conv and TConv stand for the convolutional and transposed convolutional layer respectively. 
     % \Salnote{Same thing as above. What is the purpose of making a complicated figure and not explaining what is going on here. If you need to refer to [??] for details, then do not add this figure.}
     }
     
     %Specifically, $\boldsymbol{x}_0, \boldsymbol{x}_1, \boldsymbol{x}_2, \boldsymbol{x}_3$ in each stage are nearest down sampled $\boldsymbol{x}$ from original resolution. During the decoding process in every stage $\#i\in \left\{1, 2, 3 \right\}$ expect the stage $\#0$, we just need to recover the three-quarter region of $\boldsymbol{x}_i$, since the rest part has been recovered during the previous stage. ({\color{BlueLine}{Blue}} lines depict the encoding process, by contrast, {\color{OrangeLine}{orange}} lines represent for decoding steps. Black lines are paths duplicated in both encoding and decoding.)
     %(c) A demonstration of decode $\boldsymbol{x}_i$ uses predicted parameters $\boldsymbol{Z}_i$ and bit-streams. Arithmetic decoding is omitted. Lossless RIC applies the multichannel processing following the Bayer pattern to further reduce the bit-streams of $\boldsymbol{x}$ in arithmetic coding.}
\end{figure*}

\subsubsection{Distribution Regularization Using Gamma Correction}
\label{sec:hdr_split}
To eliminate the negative impact of a large $k$ of a RAW image for RAW-domain detection, we use a simple-yet-efficient  gamma correction defined in ITU-R BT.709~\cite{stokes1996standard} to reduce the $k$ by brightening the dark area and darkening the light area.
As shown in the rightmost subplot of Fig.~\ref{fig:RAW_CV}, those sub-images generated by the gamma correction effectively reduce the $k$ of the original RAW input, making the convergence of RAW-domain detector faster and the resultant model more robust with better performance (see Table~\ref{table:ablation_hdr_split}).

\subsection{Low-Level RAW Image Compression}
\newcommand{\R}{
\begin{tikzpicture}[baseline]
\draw (0,-0.2em) -- (0em,0.8em) -- (1em,0.8em) -- (1em,-0.2em) -- (0,-0.2em);
\draw (0, 0.3em) -- (1em, 0.3em);
\draw (0.5em, -0.2em) -- (0.5em, 0.8em);
\draw[fill=red] (0em, 0.3em) rectangle (0.5em,0.8em);
\end{tikzpicture}
}
\newcommand{\Gr}{
\begin{tikzpicture}[baseline]
\draw (0,-0.2em) -- (0em,0.8em) -- (1em,0.8em) -- (1em,-0.2em) -- (0,-0.2em);
\draw (0, 0.3em) -- (1em, 0.3em);
\draw (0.5em, -0.2em) -- (0.5em, 0.8em);
\draw[fill=green] (0.5em, 0.3em) rectangle (1em,0.8em);
\end{tikzpicture}
}
\newcommand{\Gb}{
\begin{tikzpicture}[baseline]
\draw (0,-0.2em) -- (0em,0.8em) -- (1em,0.8em) -- (1em,-0.2em) -- (0,-0.2em);
\draw (0, 0.3em) -- (1em, 0.3em);
\draw (0.5em, -0.2em) -- (0.5em, 0.8em);
\draw[fill=green] (0em, -0.2em) rectangle (0.5em,0.3em);
\end{tikzpicture}
}
\newcommand{\B}{
\begin{tikzpicture}[baseline]
\draw (0,-0.2em) -- (0em,0.8em) -- (1em,0.8em) -- (1em,-0.2em) -- (0,-0.2em);
\draw (0, 0.3em) -- (1em, 0.3em);
\draw (0.5em, -0.2em) -- (0.5em, 0.8em);
\draw[fill=blue] (0.5em, -0.2em) rectangle (1em,0.3em);
\end{tikzpicture}
}
Applications often mandate the archival of images for after-action review and analysis, leading to the desire for high-efficiency image compression.
Existing image codecs mainly deal with RGB, monochrome, or YCbCr color spaces. In this section, we explore the feasibility of encoding RAW images directly. We suggest using learned image compression for this purpose, not only because of its superior coding efficiency~\cite{chen2019neural,lu2022transformer, lu2022high} but also its flexibility to support the coding of various image sources. 

\subsubsection{Lossy RAW Image Compression (RIC)}
We extend the TinyLIC proposed in~\cite{lu2022high} for lossy RIC with the following amendments shown in Fig.~\ref{fig:lossy_lossless_arch}:
\begin{enumerate}
% [leftmargin=1.2em]
    \item {\it Rearrangement:} First, we follow \cite{liu2019learning} to rearrange Bayer RAWs to RGGB presentation. At each pixel position, its spectral components, e.g., (R, G, G, B), are stacked for compression, which is similar to the pixel of (R, G, B) used in default TinyLIC;
    \item {\it Normalization:} 
    We normalize the original Bayer RAW files through linearization using 
\begin{align}
    \boldsymbol{x} = \frac{\boldsymbol{x}_{file} - \texttt{blackLev}}{\texttt{saturationLev}- \texttt{blackLev}}, \label{eq:linearization}
\end{align}
where $\boldsymbol{x}_{file}$ is the Bayer RAW collected by an image sensor, and $\boldsymbol{x}$ is normalized RAW in the range of [0, 1]. $\texttt{saturationLev}$ and $\texttt{blackLev}$ indicate the dynamic range of image pixels, which can be directly retrieved from the EXIF metadata of the RAW input.
\end{enumerate}

The $\texttt{saturationLev}$ is related to the bit depth supported by the specific camera for RAW acquisition. It is the same for all RAW images captured by the same camera. 
 We choose to embed the $\texttt{blackLev}$ coefficients globally (e.g., $1\times16=16~\textrm{bits}$ present $<$ \SI{0.03}{\percent} of the total bits used by a $512\times512$ image) to ensure the encoder-decoder consistency. Stacking RGGB components could let the lossy RIC explicitly learn the inter-spectral or inter-color correlations. As will be shown later, our lossy RIC significantly outperforms the traditional methods on RAW encoding. For other imaging patterns, e.g., RYYB used in the Huawei P30 Pro can be processed similarly.

\subsubsection{Lossless RIC}

Numerous applications need to cache RAW images losslessly for future processing (e.g., professional photography, safety-critical event record, etc.). We further extend the lossy RIC to support the lossless mode. 

{\bf Multiresolution Decoding.} In lossy RIC, the processing steps in the main encoder-decoder pair are symmetrically mirrored~\cite{lu2022high}. In lossless RIC, we redesign the main decoder while keeping the same main encoder as in lossy RIC.
Instead of decoding the full-resolution reconstruction in one shot, we gradually decode the pixels to refine the image resolution from $\boldsymbol{x}_{0}$ to $\boldsymbol{x}_{4}$ ($\boldsymbol{x}_{4}$ = $\boldsymbol{x}$) in Fig.~\ref{fig:lossless_RIC_decoder}. As seen, the compression performance is improved by exploiting the correlations from previously-decoded neighbors. Such progressive decoding enables the low-resolution preview that is not available in existing approaches~\cite{l3c,srec} but is a highly-desired tool in commercial cameras.

The lossless decoding of $\boldsymbol{x}_{i}$, $i\in[1, 4]$ is organized as:
\begin{enumerate}
    \item The $\boldsymbol{x}_{i}$ is dyadically upsampled from $\boldsymbol{x}_{i-1}$ with 2$\times$ scaling at each dimension, i.e., each pixel in $\boldsymbol{x}_{i-1}$ is expanded to four pixels arranged in a local 2$\times$2 patch in $\boldsymbol{x}_{i}$;
    \item As  in Fig.~\ref{fig:lossless_RIC_decoder}, the upper-left pixel of each 2$\times$2 patch of $\boldsymbol{x}_{i}$ is directly filled using the corresponding pixel of $\boldsymbol{x}_{i-1}$ (highlighted with red box), while the other three pixels in each 2$\times$2 patch of $\boldsymbol{x}_{i}$ is decoded using the conditional probability of logistic distribution that is characterized by the $\boldsymbol{Z}_i$;
\end{enumerate}
    
    Note that a special case is made for the processing of $\boldsymbol{x}_0$ since there are no available pixels from a lower resolution scale. As a result, $\boldsymbol{Z}_0$ is generated by parsing the compressed bitstream only.

{\bf Multichannel Processing.} As seen, $\boldsymbol{Z}$\footnote{The subscript index $i$ is omitted in $\boldsymbol{Z}_i$ for general assumption.}  thoroughly aggregates information from the lower resolution scale, which is then input into the Multichannel Processing Unit (MPU) to encode/decode pixels in a predefined order to fully exploit the cross-channel correlations. Here, we use $r, g_r, g_b, b$ to represent four Bayer pattern channels. For the MPU in Fig~\ref{fig:lossless_RIC_details}, only $\boldsymbol{Z}$ is used to derive the  probability of $g_{r}$ samples for its encoding and decoding first. Then both $\boldsymbol{Z}_i$ and previously-processed samples are used to process $g_{b}$, $r$, and $b$ sequentially. 
Here we parameterize the sample distribution $P$ as a mixture of logistic distribution $P_{l}$~\cite{mentzer2019practical}:
\begin{equation}
    \begin{split}
        P ( g_{r} ~ | ~\boldsymbol{Z} ) &= \sum\nolimits_{k=1}^K \omega_{g_rk} \cdot P_l(g_r ~ | ~ \mu_{g_rk}, \sigma_{g_rk} ) \\
        P ( g_{b} ~ | ~\boldsymbol{Z}, g_r ) &= \sum\nolimits_{k=1}^K \omega_{g_bk} \cdot P_l(g_b ~ | ~ \tilde{\mu}_{g_bk}, \sigma_{g_bk} ) \\
        P ( r ~ | ~\boldsymbol{Z}, g_r, g_b ) &= \sum\nolimits_{k=1}^K \omega_{rk} \cdot P_l(r ~ | ~ \tilde{\mu}_{rk}, \sigma_{rk} ) \\
        P ( b ~ | ~\boldsymbol{Z}, g_r, g_b, r ) &= \sum\nolimits_{k=1}^K \omega_{bk} \cdot P_l(b ~ | ~ \tilde{\mu}_{bk}, \sigma_{bk} ), \\
    \end{split}
\end{equation}
where the $\tilde{\mu}$ builds the cross channels dependency:
\begin{equation}
\begin{split}
    \tilde{\mu}_{g_bk} &= \mu_{g_rk} + \lambda_{\alpha k} \cdot g_r, \\
    \tilde{\mu}_{rk} &= \mu_{rk} + \lambda_{\beta k} \cdot g_r + \lambda_{\gamma k} \cdot g_b, \\
    \tilde{\mu}_{bk} &= \mu_{bk} + \lambda_{\delta k} \cdot g_r + \lambda_{\epsilon k} \cdot g_b + \lambda_{\zeta k} \cdot r.
\end{split}
\end{equation}
The logistic function $P_{l}(x|\mu, \sigma)$ can be easily calculated using \{$\textrm{sigmoid}(\frac{x-\mu+1/2s}{\sigma})$ - $\textrm{sigmoid}(\frac{x-\mu-1/2s}{\sigma})$\} with $s$ = $\texttt{saturationLev}$-$\texttt{blackLev}$. Parameters $\mu, \sigma, \omega$ and $\lambda$ are a part of the prior $\boldsymbol{Z}$ and we set $K=10$ as in~\cite{cao2020lossless, mentzer2019practical}. In training, we only need to minimize the entropy loss $\mathcal{L}=-\mathbb{E}\left[ \log P( r, g_{r}, g_{b}, b ~|~\boldsymbol{Z} ) \right]$ for lossless compression.
% \Salnote{I do not see where $g_r,g_b,r,b$ are defined above.} \zhihao{Fixed}
\section{Experimental Results}
This section presents experiments where we execute tasks using RAW images directly.

\begin{table*}
    % [htbp]
    \small
    \setlength{\tabcolsep}{5.2pt}
    \centering
    \caption{\edited{{\bf MultiRAW Dataset:} A collection of 7,469 labeled RAW images from five diverse camera sensors. The dataset covers various settings (e.g., Bayer pattern, bit depth) and application scenarios.}}
    \label{tab:Multi-RAW-Datasets}    
    \begin{tabular}{llcclll}
    \toprule
        \textbf{Camera} 
        & \textbf{Usage} 
        & \textbf{Bayer Pattern} 
        & \textbf{Bit Depth} 
        & \textbf{Task} 
        & \textbf{Scenarios}
        & \textbf{No. of Images} \\ \midrule
        iPhone XSmax   
        & Mobile                
        & RGGB                   
        & 12 
        & \edited{Detection, Segmentation} 
        & \edited{Day}
        & 1153           
        \\ \midrule
        Huawei P30pro 
        & Mobile                
        & RYYB                   
        & 12   
        & \edited{Detection}
        & \edited{Day, Night} 
        & 3004           
        \\ \midrule
        asi 294mcpro   
        & Industrial 
        & RGGB                   
        & 14   
        & \edited{Detection}
        & \edited{Day, Night}
        & 2950           
        \\ \midrule
        Oneplus 5t      
        & Mobile                
        & RGGB             
        & 10   
        & \edited{Detection}
        & \edited{Day}
        & 101            
        \\ \midrule
        \edited{LUCID TRI054S}      
        & \edited{Autopilot}      
        & \edited{RGGB}             
        & \edited{24}
        & \edited{Detection}
        & \edited{Day, Tunnel}  
        & \edited{261}
        \\ \bottomrule
    \end{tabular}
\end{table*}

\subsection{Experiment Setup}
\label{sec:simu_gen}
We first list the RAW and RGB image datasets used in experiments, and then discuss the setup of the proposed Unpaired CycleR2R to generate simRAW images.

\newcommand{\RGBcity}{$\text{RGB}_{\text{b}}$}
\newcommand{\RGBipone}{$\text{RGB}_{\text{i}}$}
\newcommand{\RAWipone}{$\text{RAW}_{\text{i}}$}
\newcommand{\pseudoRAWcity}{$\text{RAW}_{\text{b}}$}
\newcommand{\Baselines}{\textcolor{blue}{$\spadesuit$~}}
\newcommand{\InvISP}{\textcolor{orange}{$\blacksquare$~}}
\newcommand{\DomainAdap}{\textcolor{green}{$\blacklozenge$~}}
\newcommand{\Ours}{\textcolor{red}{$\bigstar$~}}

\begin{table*}
    \small
    \setlength{\tabcolsep}{8pt}
    \centering
    \caption{
    {\textbf{Recall and Average Precision (AP) of Object Detection on the testing set of iPhone RAWs.} RGB detector is trained using original RGB images in \textit{BDD100K}~\cite{yu2020bdd100k} (e.g., \RGBcity); Various simRAW datasets associated with \RGBcity~are generated using different methods which are simply marked as sim\pseudoRAWcity~to train the RAW detector. 
    The testing RAW images in iPhone RAW \RAWipone~and their paired RGB images in \RGBipone~converted using built-in iPhone ISP are tested accordingly.
    All detectors are shared with the same head, \ie, YOLOv3~\cite{yolo} and the same backbone, \ie, MobileNetV2~\cite{sandler2018mobilenetv2} for a fair comparison.
    \Baselines Baselines, \DomainAdap Domain Adaptation Solutions, \InvISP invISP Methods, \Ours Ours.}
    }
    \label{table:without_labeled_detection}
    \begin{tabular}{lrrrrr}
    \toprule
            \textbf{Method} & \textbf{ invISP Training} & \textbf{Detector Training} & \textbf{Detector Testing} & \textbf{Recall} & \textbf{AP}    \\ \midrule
         \Baselines Naive Baseline & -  & \RGBcity & \RAWipone & 67.5 & 51.1 \\ 
         \Baselines RGB Baseline  & -  & \RGBcity & \RGBipone & 74.7 & 55.6 \\ \midrule
        % \DomainAdap CycleGAN (ICCV'17)~\cite{zhu2017unpaired} &  -  &  \RGBcity,~\RAWipone & \RAWipone  & 10.8 & 3.9 \\
        \DomainAdap DA-Faster (CVPR'18)~\cite{chen2018domain} &  -  &  \RGBcity,~\RAWipone & \RAWipone  & 13.7 & 12.9 \\
        \DomainAdap MS-DAYOLO (ICIP'21)~\cite{hnewa2021multiscale} &  -  & \RGBcity,~\RAWipone & \RAWipone  & 31.2 & 29.7 \\
        \DomainAdap AT (CVPR'22)~\cite{li2022cross} &  -  & \RGBcity,~\RAWipone & \RAWipone  & 68.9 & 53.2 \\\midrule
        \InvISP InvGamma (ICIP'19)~\cite{koskinen2019reverse} &  \RGBipone,~\RAWipone  & sim\pseudoRAWcity & \RAWipone  & 68.6 & 48.7 \\
        \InvISP CycleISP (CVPR'20)~\cite{zamir2020cycleisp} & \RGBipone,~\RAWipone  & sim\pseudoRAWcity & \RAWipone & 71.6 & 52.7 \\ 
        % \InvISP Invertible-ISP (CVPR'21)~\cite{xing2021invertible} & \RGBipone,~\RAWipone & \pseudoRAWcity & \RAWipone & - & - \\ 
        \InvISP CIE-XYZ Net (TPAMI'21)~\cite{afifi2020cie} & \RGBipone,~\RAWipone & sim\pseudoRAWcity & \RAWipone & 72.2 & 53.0 \\ 
        % % Marcos~\cite{conde2022model} & simu RAW & RAW & 0.091 & 0.010 \\ \midrule
        \InvISP MBISPLD (AAAI'22)~\cite{conde2022model} & \RGBipone,~\RAWipone & sim\pseudoRAWcity & \RAWipone & 73.0 & 53.7 \\ \midrule
        \Ours \textbf{Unpaired CycleR2R} & \RGBcity,~\RAWipone & sim\pseudoRAWcity & \RAWipone & \textbf{76.1}  & \textbf{59.1} \\
    \bottomrule
    \end{tabular}
\end{table*}

\begin{figure*}[ht]
\centering
\subfloat[iPhone XSmax.]{\includegraphics[width=0.32\linewidth]{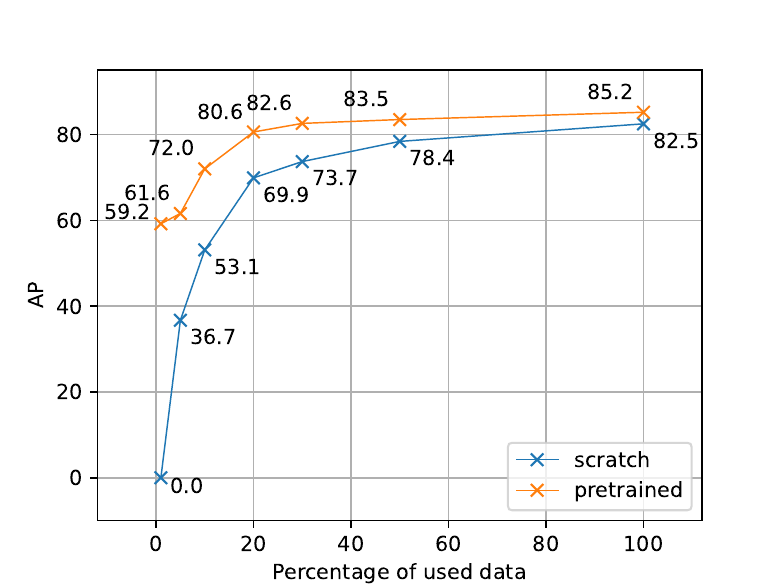} \label{fig:iphone-few-shot}}
\subfloat[Huawei P30pro.]{\includegraphics[width=0.32\linewidth]{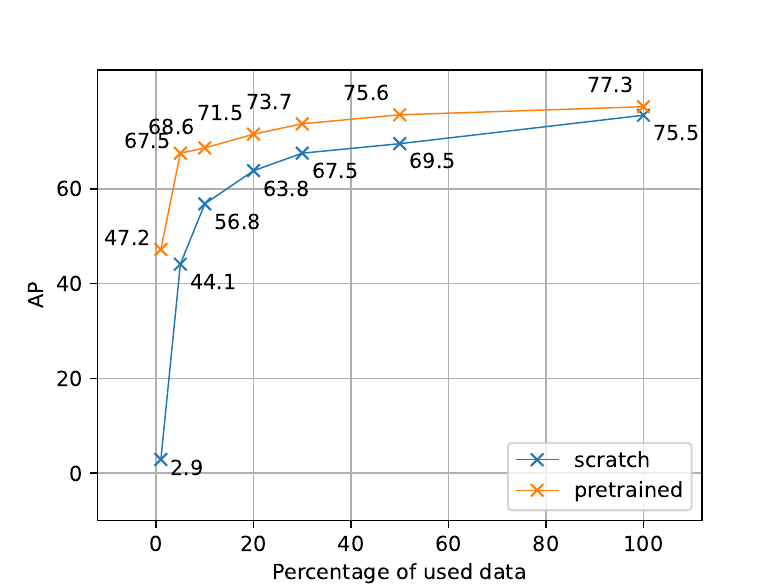} \label{fig:huawei-few-shot}}
\subfloat[asi 294mcpro.]{\includegraphics[width=0.32\linewidth]{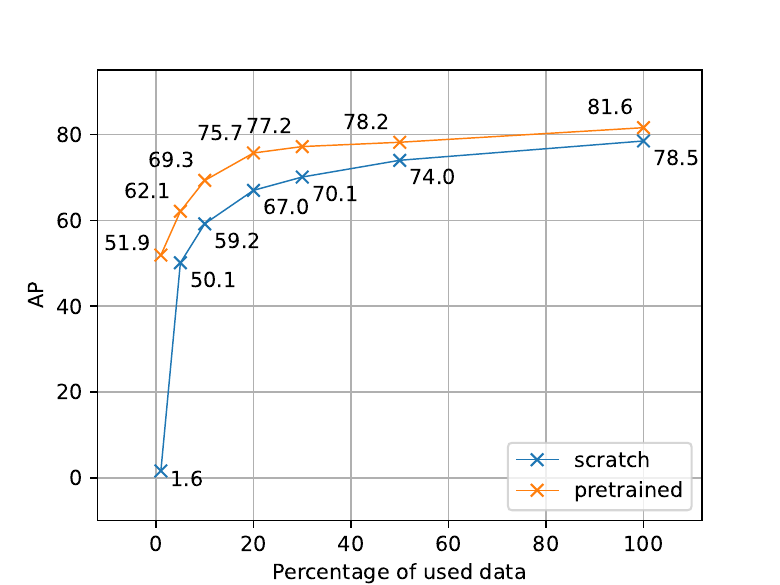} \label{fig:asi-few-shot}}
%\end{center}
  \caption{{\bf Few-shot model finetuning using limited camera RAWs.} The ``pretrained'' YOLOv3 is pretrained using samples in simRAW$_\text{b}$ generated by our invISP, and the ``scratch'' model is first randomly initialized and then directly trained using labeled real RAW images from a specific camera model.}
\label{fig:Few_Shot_Detection}
\end{figure*}

\subsubsection{Datasets}
% \Salnote{Who collected this dataset?}
\edited{MultiRAW is a high-resolution RAW image dataset 
% \Salnote{that we???}\zhihao{Fixed} 
acquired using popular camera sensors. Specifically, the entire dataset was shot using five cameras fixed on the car dashboard at different times (day and night) and geolocations (rural, tunnel, and urban areas) to simulate real-life autonomous driving.} Having RAW images from different cameras allows us to validate the generalization of the proposed method. 

\edited{Table~\ref{tab:Multi-RAW-Datasets} provides details about 7,469 RAW images that cover a variety of application scenarios (mobile/ industrial/autopilot), imaging Bayer patterns ( RGGB/RYYB), and bit depths (e.g., $10$/$12$/$16$/$24$). All RAW images could be converted to RGB counterparts using the corresponding in-camera ISP.}
Fine-grained detection and segmentation bounding boxes for cars, persons, traffic lights, and traffic signs are labeled manually by a third-party professional image labeling firm. 

Among the RAW images captured by iPhone XSmax, Huawei P30pro, asi 294mcpro, and the LUCID TRI054S, we randomly selected \SI{30}{\percent} samples as the test set and the remaining ones as the training set. For RAW images acquired by Oneplus 5t, we use all of them for testing.

\textit{BDD100K}~\cite{yu2020bdd100k} is one of the largest autonomous driving datasets. It contains \SI{100}{k} RGB images taken in diverse scenes such as city streets, residential areas, and highways, making object detection more challenging and close to real-life scenarios. We follow the official data splitting of \SI{70}{k}, \SI{10}{k}, and \SI{20}{k} images for training, validation, and testing, respectively. 
{To avoid the discrepancy of traffic signs captured in {\it BDD100K} (\eg, collected across tens of different countries) and MultiRAW (\eg, collected mainly in mainland China) datasets, we only conducted training and testing on the ``car'' category that has the largest number of objects (\eg, $>$\SI{55}{\percent}) and presents the smallest differences.}
     
\textit{Flicker2W}~\cite{liu2020unified} is widely used to train learned image compressors~\cite{lu2022high}.
Although we collected more than 7k samples in MultiRAW, its size is less than that of RGB image datasets (\eg, 100k in {\it BDD100K}) used in popular tasks. To train robust models, we apply our  Unpaired CycleR2R to generate simRAW images using popular RGB datasets. For example, all RGB samples in {\it BDD100K} and {\it Flickr2W} are converted to  RAW images to retrain existing RGB-domain models.

\subsubsection{Training  Unpaired CycleR2R for simRAW Generation}
{We first train our Unpaired CycleR2R to generate simRAW images to finetune/retrain existing RGB-domain models for RAW-domain tasks. For example,  unpaired RGB and iPhone RAW images that were respectively chosen from the {\it BDD100K}~\cite{yu2020bdd100k} (without knowledge of camera sensors) and MultiRAW datasets are used to train the Unpaired CycleR2R for the high-level detector.
We also use the original {\it Flicker2W}~\cite{liu2020unified} along with the random iPhone RAW images to train the Unpaired CycleR2R for the low-level compressor.

We train the model using randomly selected patches of size $512\times512$ and batch size $8$. We applied random scaling and reflection to augment the training data. A single NVIDIA 3090Ti was used for $20,000$ iterations of an Adam optimizer with a learning rate $5\cdot10^{-4}$. The discriminators were set with a learning rate at $5\cdot10^{-5}$. The momentum for Adam is set to $\left\{0.9, 0.99\right\}$. Finally, $g_{\rm invISP}$ from the Unpaired CycleR2R framework was used to produce simRAW$_{\rm b}$ and simRAW$_{\rm f}$ images using RGB images in {\it BDD100K}~\cite{yu2020bdd100k} and {\it Flicker2W}~\cite{liu2020unified}}. 

\subsection{RAW-domain Object Detection}
\label{sec:exp_obj_det}
This section shows that the object detector trained on simRAW images can directly process real-life RAW images and provide good accuracy. Then, we discuss how the performance of a simRAW-pretrained detector can be further improved by the few-shot finetuning using a limited number of labeled real RAW images from a camera used in a specific application scenario. Our results show that such a few-shot fine-tuned detector consistently outperforms the model trained from scratch using the same set of real RAW images.

{\bf Training RAW-domain detector.}
We chose the popular YOLOv3 as our baseline object detector~\cite{yolo} and the prevalent MobileNetv2~\cite{sandler2018mobilenetv2} as the backbone. 

\begin{itemize}
    \item RAW-domain YOLOv3 can be trained using simRAW$_{\rm b}$ set generated from the RGB samples (RGB$_{\rm b}$) in {\it BDD100K} using various invISP methods~\cite{koskinen2019reverse, zamir2020cycleisp, afifi2020cie, conde2022model} (see Table~\ref{table:without_labeled_detection}). We trained YOLOv3 using SGD with the batch size at $8$, a momentum of $0.9$, and a weight decay of $10^{-4}$. A learning rate of $0.02$ was used in training for $30$ epochs.
    \item  RAW-domain YOLOv3 can also be retrained using unpaired RGB$_{\rm b}$ and RAW$_{\rm i}$ sets of respective {\it BDD100K} and iPhone RAW (a subset of MultiRAW) through the use of domain adaptation (DA) techniques following their official settings~\cite{chen2018domain, hnewa2021multiscale, li2022cross}. 
\end{itemize}
%On one hand, 
%On the other hand,
We tested the trained models on the test set of RAW$_{\rm i}$.

{\bf Remark.} For the invISP methods listed in Table~\ref{table:without_labeled_detection} that strictly required RGB-RAW pairs from a specific camera model, we fine-tuned their pre-trained models using our MultiRAW dataset by applying the same training settings used to train our Unpaired Cycle R2R model (\ie epochs, training patches).

\begin{figure}
     \centering
     \includegraphics[width=0.8\linewidth]{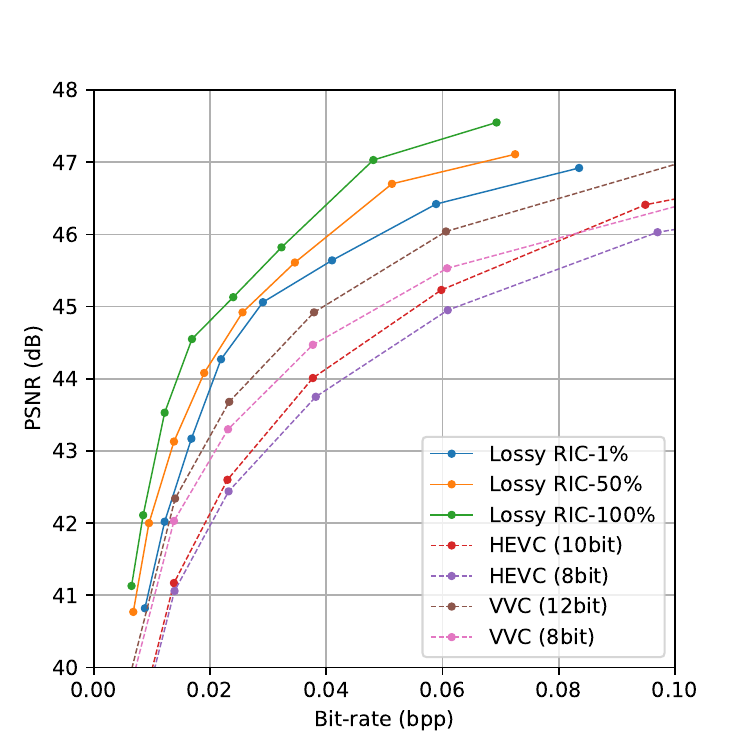} 
     \caption{{{\bf Rate-distortion Performance of Lossy RIC.}  These testing RAW images are from our MultiRAW dataset  and distortion is measured in the RAW domain. All RICs are pre-trained with simRAW images and fine-tuned using a limited number of real captured RAWs. Lossy RIC-\SI{1}{\percent}, Lossy RIC-\SI{50}{\percent} and Lossy RIC-\SI{100}{\percent} denote \SI{1}{\percent}, \SI{50}{\percent} and \SI{100}{\percent} real RAW samples are used for fine-tuning. HEVC and VVC Intra coders are evaluated for comparison.}}
     \label{fig:Lossy-RIC-Results}
\end{figure}

\begin{table}
    \small
    \setlength{\tabcolsep}{2.5pt}
    \centering
    \caption{\textbf{The lossless compression results on MultiRAW.}}
    \label{table:lossless_compression}
    \begin{tabular}{ccccr}
    \toprule
\multirow{2}{*}{\textbf{Method}} &
    \multirow{2}{*}{\textbf{Camera}} &
    \multicolumn{2}{c}{\textbf{Latency (s)}} &
    \multirow{2}{*}{\textbf{BPP}} \\ \cmidrule{3-4}
 &
    &
    Encoder &
    Decoder &
     \\ \midrule

CinemaDNG &  
    \multirow{6}{*}{\makecell[c]{iPhone XSmax\vspace{+0.1em}\\(RGGB/\SI{12}{bits})\vspace{+0.1em}\\  $4032\times3024$}}  &
    0.49 &
    0.52 &
    9.51 \\ 

JPEG XL &  
    &
    19.73 &
    6.38 & 
    5.46 \\ 

FLIF    &
    &
    38.67 & 
    8.76 &
    5.46 \\ 

PNG &
    & 
    0.44 &
    0.24 & 
    7.98        \\ 

\textbf{Lossless RIC-\SI{1}{\percent}}  &  
    &
    \multirow{2}{*}{1.35} &
    \multirow{2}{*}{0.78} &
    5.29 \\

\textbf{Lossless RIC-\SI{100}{\percent}}  &  
    &
    &
    &
    \textbf{5.21} \\ \midrule

CinemaDNG &  
    \multirow{6}{*}{\makecell[c]{Huawei P30pro\vspace{+0.1em}\\(RYYB/\SI{12}{bits})\vspace{+0.1em}\\  $4032\times3024$}}  &
    0.42 &
    0.46 &
    9.52 \\ 

JPEG XL &  
    &
    21.37 &
    6.67 & 
    5.62 \\

FLIF    &
    &
    41.67 & 
    9.56 &
    5.60 \\ 

PNG &
    & 
    0.39 &
    0.22 & 
    7.93  \\ 
    
\textbf{Lossless RIC-\SI{1}{\percent}}  &  
    &
    \multirow{2}{*}{1.32} &
    \multirow{2}{*}{0.79} &
    5.16 \\

\textbf{Lossless RIC-\SI{100}{\percent}}  &  
    &
    &
    &
    \textbf{5.12} \\ \midrule

CinemaDNG &  
    \multirow{6}{*}{\makecell[c]{asi 294mcpro\vspace{+0.1em}\\(RGGB/\SI{14}{bits})\vspace{+0.1em}\\  $4144\times2822$}}  &
    0.40 &
    0.44 &
    18.85 \\ 

JPEG XL &  
    &
    9.37 &
    2.91 & 
    1.46 \\ 

FLIF    &
    &
    18.83 & 
    4.26 &
    1.36 \\ 

PNG &
    & 
    0.32 &
    0.17 & 
    2.86  \\ 

\textbf{Lossless RIC-\SI{1}{\percent}}  &  
    &
    \multirow{2}{*}{1.83} &
    \multirow{2}{*}{0.81} &
    0.77 \\

\textbf{Lossless RIC-\SI{100}{\percent}}  &  
    &
    &
    &
    \textbf{0.75} \\ \bottomrule

% CinemaDNG &  
%     \multirow{5}{*}{\makecell[c]{Oneplus 5t\vspace{+0.1em}\\(RGGB/10bits)\vspace{+0.1em}\\  $4656\times3496$}}  &
%     - &
%     - &
%     10.14 \\ 

% JPEG XL &  
%     &
%     - &
%     - & 
%     4.43 \\ 

% FLIF    &
%     &
%     - & 
%     - &
%     6.44 \\ 

% PNG &
%     & 
%     - &
%     - & 
%     6.53  \\ 

% \textbf{Ours}  &  
%     &
%     - &
%     - &
%     \textbf{4.38} \\ \bottomrule
    \end{tabular}
\end{table}

{\bf Comparative Studies.}
Table~\ref{table:without_labeled_detection} reports the object detection accuracy in the RAW domain. 
%
% \Salnote{We MUST discuss the results in Tables and Figures in the text. We need to walk through all the rows/columns of the table at a high level and then draw conclusions about what is the main takeaway from the results. The text does not adequately discuss that right now. }
%
% \zhihao{
In general, our model achieves SOTA performance with a large margin (more than \SI{5.4}{\percent}).  DA approaches in~\cite{chen2018domain, hnewa2021multiscale} failed to learn the effective mapping between RAW and RGB samples. This is mainly because the domain difference between RAW and RGB samples is fundamentally different from that between two RGB sets studied in~\cite{chen2018domain, hnewa2021multiscale}.  A self-supervised learning approach was proposed~\cite{li2022cross} by introducing more generic representation features, which then boosted the performance noticeably. However, the sizeable training resources (\SI{96}{\giga\relax} GPU memory for five days) limit its adoption in practical applications. And as we mentioned above, previous invISP methods~\cite{conde2022model, afifi2020cie, zamir2020cycleisp, koskinen2019reverse} modeled a known camera could not convert the RGB from an unknown camera properly. 
% }
% In general, our model achieves SOTA performance
% with \SI{59.1}{\percent} AP and outperforms the most recent invISP method---MBISPLD~\cite{conde2022model} by \SI{5.4}{\percent} and the DA approach---AT~\cite{li2022cross} by \SI{5.9}{\percent}. More gains are observed in comparison to other methods. A similar trend of improvement appears for the Recall metric as well.
{More importantly, our model is the only one exceeding the RGB baseline that is widely deployed in practical applications. These results offer promising prospects for RAW-domain task inference, for which  existing RGB-domain models are retrained using samples in simRAW$_{\rm b}$ that are generated using the Unpaired CycleR2R. In contrast, directly feeding RAW images to the RGB-domain detector presents inferior performance as exemplified in the Naive Baseline that lacks any RAW-domain knowledge. }  

% {
The performance of simRAW-pretrained YOLOv3 can be further improved by few-shot learning using a limited number of labeled real RAW images from a specific camera model. As shown in Fig.~\ref{fig:Few_Shot_Detection}, the detection accuracy is  improved consistently for three cameras. We also provide the model accuracy when training YOLOv3 from scratch, for which the model is first randomly initialized and then trained using the same labeled real RAWs. The results show that fine-tuning simRAW-pretrained YOLOv3 consistently outperforms training the model from scratch. More specifically, \SI{2.7}{\percent} - \SI{59.2}{\percent} AP improvement for iPhone XSmax, \SI{1.8}{\percent} - \SI{44.3}{\percent} for Huawei P30pro, and \SI{3.1}{\percent} - \SI{50.4}{\percent} for asi 294mcpro clearly reveals the advantages of pretraining a model using simRAW images. 
% Note that the testing conditions are kept the same. } 

% {The results in \Salnote{Figure ??/Table ??} show that fine-tuning simRAW-pretrained YOLOv3 consistently outperforms training the model from scratch.}

\begin{figure*}
% [h]
\begin{center}
\begin{tabular}[b]{c@{ } c@{ }  c@{ } c@{ } c@{ } c@{ }	}\hspace{-4mm}
    \multirow{4}{*}{\includegraphics[width=.305\textwidth, trim=0 0 138 0, clip, valign=t]{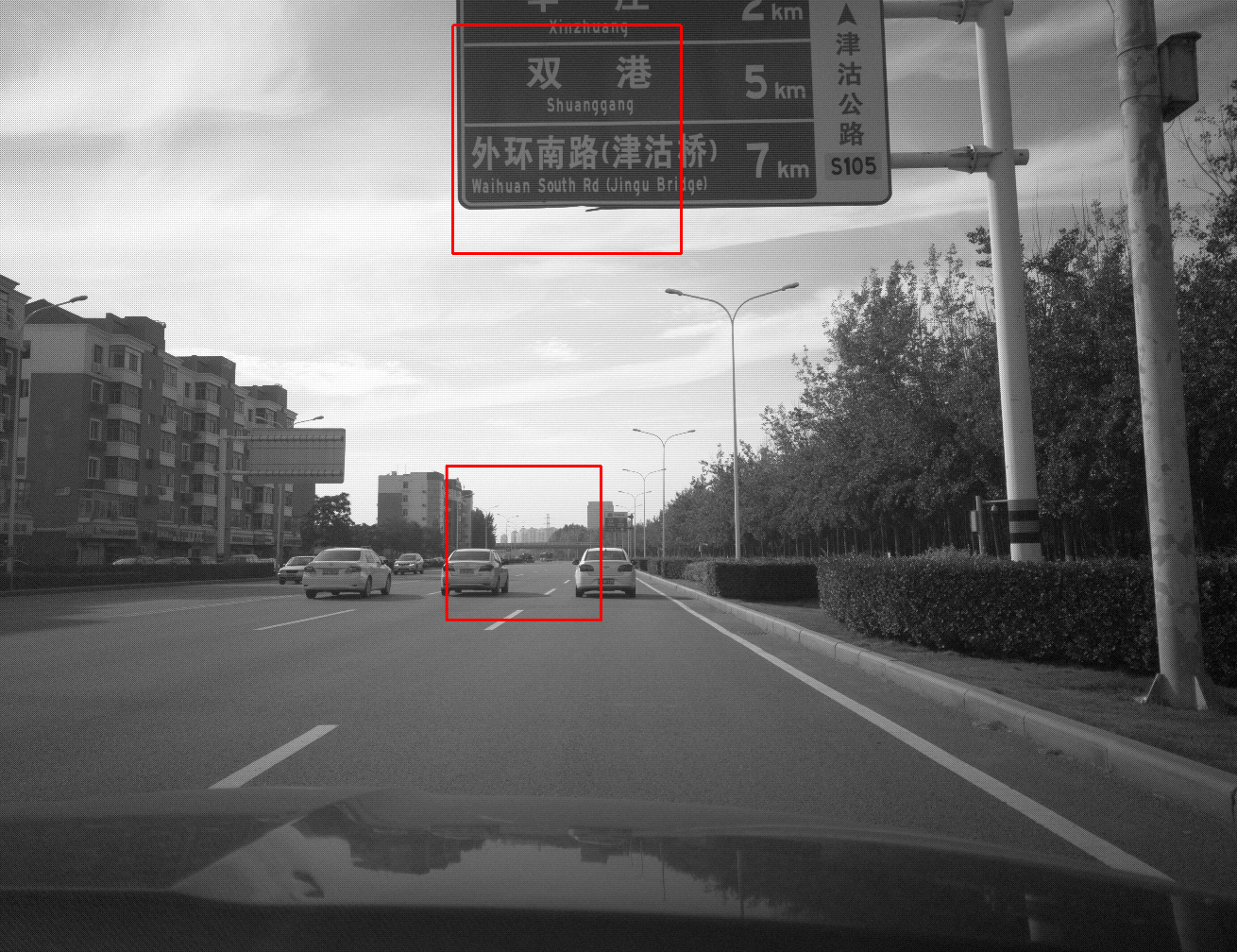}} &   
    \includegraphics[width=.125\textwidth,valign=t]{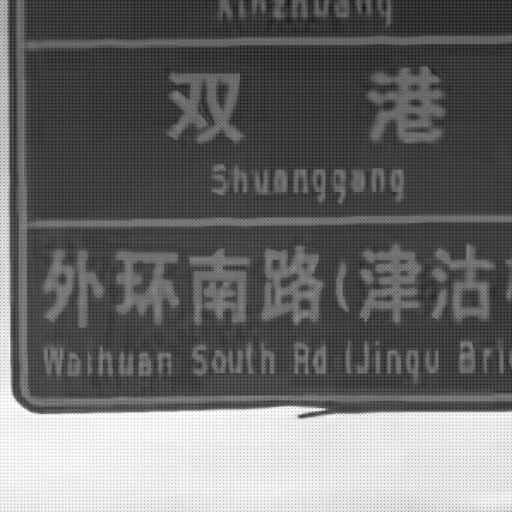}&
  	\includegraphics[width=.125\textwidth,valign=t]{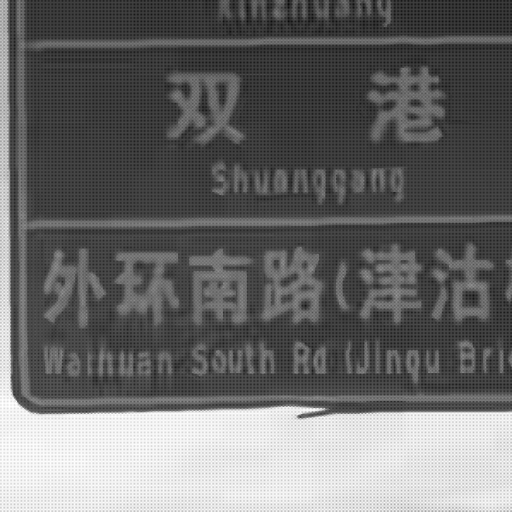}&   
    \includegraphics[width=.125\textwidth,valign=t]{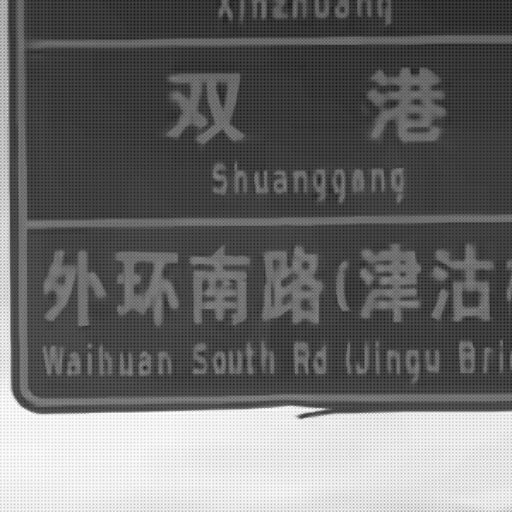}&
    \includegraphics[width=.125\textwidth,valign=t]{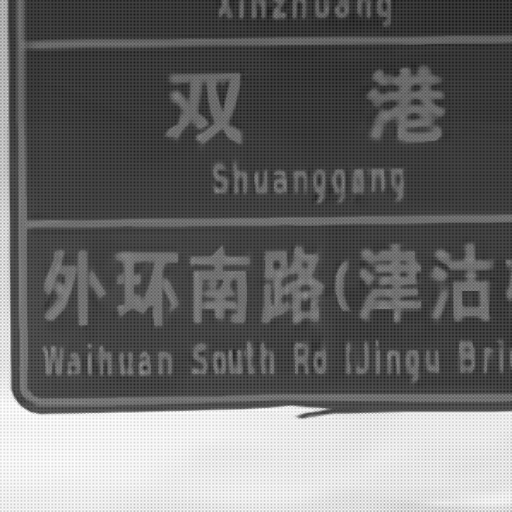}&
     \includegraphics[width=.125\textwidth,valign=t]{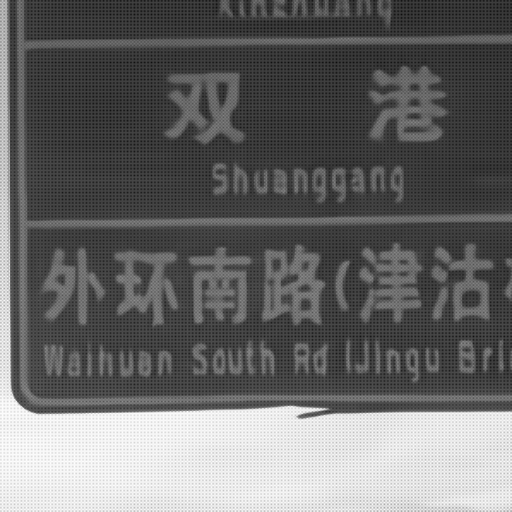}
     
\vspace{+0.2cm}
\\

    &
    \includegraphics[width=.125\textwidth,valign=t]{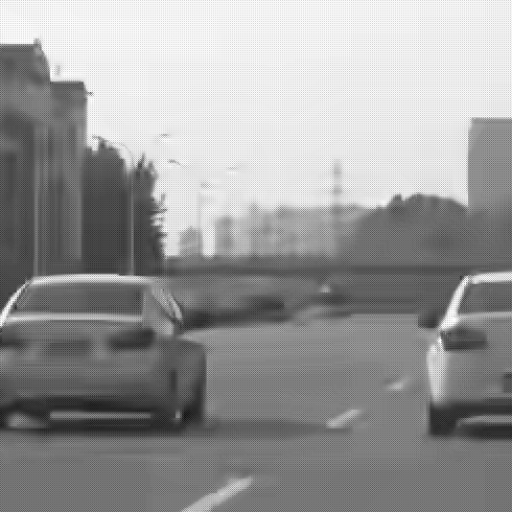}&
  	\includegraphics[width=.125\textwidth,valign=t]{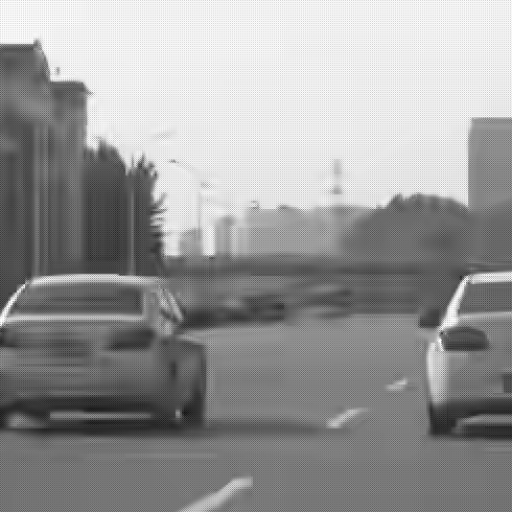}&   
    \includegraphics[width=.125\textwidth,valign=t]{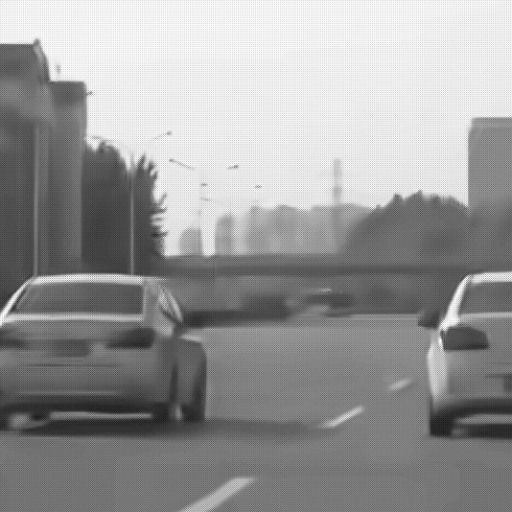}&
    \includegraphics[width=.125\textwidth,valign=t]{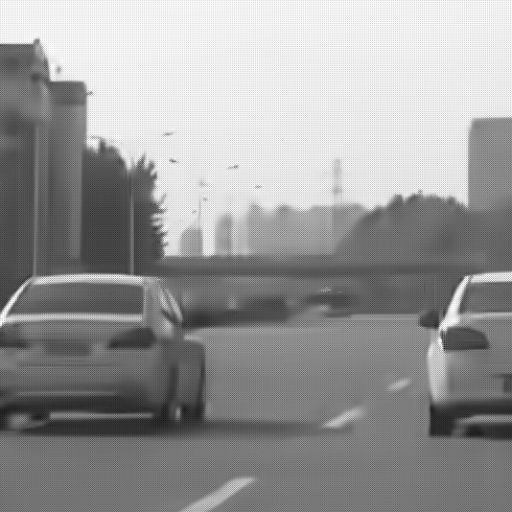}&
     \includegraphics[width=.125\textwidth,valign=t]{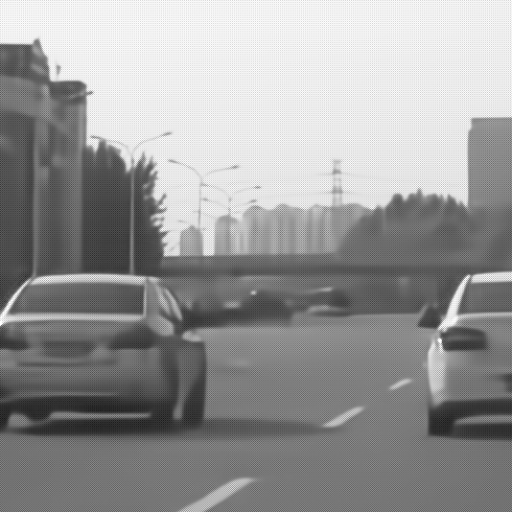}

\vspace{+0.2cm}    
\\

     \SI{12}{bpp} & \SI{0.015}{bpp} & \SI{0.015}{bpp} & \SI{0.016}{bpp} & \SI{0.017}{bpp} & {\bf \SI{0.015}{bpp}}\\
     - & \SI{44.15}{\dB} & \SI{44.28}{\dB} & \SI{44.79}{\dB} & \SI{45.14}{\dB} & {\bf \SI{46.35}{\dB}}\\
    Ground Truth  & HEVC-8bits & HEVC-10bits & VVC-8bits & VVC-12bits & {\bf Ours} \\
    
\end{tabular}
\end{center}
\caption{{\bf Qualitative Visualization.} Reconstructions and close-ups of the HEVC, VVC, and our method. Corresponding bpp and PSNR are marked. Gamma correction and brightness adjustment have been applied for a better view. {\it Zoom for better details.}}
\label{figure:lossy_compression_visual}
\end{figure*}

{\bf Implementation convenience.} Given that our method does not require paired RAW and RGB samples to train the invISP for the generation of simRAW images, it is much easier for practitioners to use our method to promote RAW-domain tasks. Furthermore, generated simRAW images can also be used to train/finetune models for other tasks, such as the segmentation method discussed in the supplementary  material.
 
Note that most of the domain adaptation approaches~\cite{chen2018domain, hnewa2021multiscale} need to modify the underlying model for each individual task manually. 
% \zhihao{
For example, Chen et al.~\cite{chen2018domain} changed the predictor head of bounding box (bbox)  and Li et al.~\cite{li2022cross} added bbox relevant loss functions which makes the migration to other tasks without bbox prediction impractical.
% } 
%
In contrast, our method just retrains existing RGB-domain models (deployed in practice) to process RAW inputs. This makes our method suitable for various tasks, including detection and segmentation. Please see the supplementary material for more details.

\subsection{RAW Image Compression (RIC)}
{This section presents results for RAW Image Compression (RIC) as a typical low-level task. Similar to Sec.~\ref{sec:exp_obj_det}, we first demonstrate the feasibility and performance of simRAW-pretrained RIC, and then illustrate further improvement by few-shot finetuning using real RAW images.}

{\bf Training RAW-domain RIC.} We use unpaired RGB and RAW samples from Flicker2W and iPhone RAW datasets to train the Unpaired Cycle R2R and generate simRAW images. Given that we do not need to label the semantic cues for high-level tasks, we  use less than \SI{1}{\percent} (\ie, $10$) real and random RAW images to train our model. We present this challenging setting to demonstrate that our method can be fine-tuned using a small number of real images, which can be especially useful in real-world settings with limited training data. 

Then we follow the same procedure in Sec.~\ref{sec:simu_gen} to convert native RGB images in {\it Flicker2W}~\cite{liu2020unified} to simRAW images (\ie simRAW$_f$) for the training of lossy and lossless RIC models, marked as Lossy RIC-\SI{1}{\percent} and Lossless RIC-\SI{1}{\percent}.  We also fine-tune these simRAW-pretrained models with more real RAW data for further improvement (see Lossy and Lossless RIC-\SI{100}{\percent}).

All RIC training threads run on a single NVIDIA 3090Ti GPU for a total of $400$ epochs. Adam optimizer with the learning rate of $10^{-4}$ and batch size of 8 for each iteration. For lossy RIC, eight models are trained to provide $8$ different bit rates (or quality levels). A pre-trained high-rate model is used to initialize the weights and to train a model for the lossless mode.

{{\bf Comparative studies of Lossy RIC.}} We compare our Lossy RIC with HEVC Intra and VVC Intra through the compression of MultiRAW test set. Because traditional video codecs cannot handle the RAW images directly, we decompose the spectral channels (RG$_r$G$_b$B) of a RAW image into an RGB image \ie, R$G_r$B and a residual image r$_g$ = G$_b$ - G$_r$ following the Apple ProRes RAW setting,  which then can be encoded by HEVC Intra, VVC Intra, respectively. Here, we apply the reference software models of HEVC and VVC for intra-compression (i.e., HEVC 16.22\footnote{\url{https://vcgit.hhi.fraunhofer.de/jct-vc/HM}} and VTM 11.0\footnote{\url{https://vcgit.hhi.fraunhofer.de/jvet/VVCSoftware_VTM}}).
Note that we perform the linearization on the RAW image by scaling the pixels in [0, 1] range. We then scale them to different bit depths before feeding them into the HEVC or VVC Intra coder. 
Since both HEVC and VVC can support higher bit depth beyond 8-bit precision, we have also tried out the 10-bit and 12-bit precision. As such, we can alleviate the quantization loss as much as possible. 
All other parameters in HEVC and VVC intra-encoders are kept same as default.

Figure~\ref{fig:Lossy-RIC-Results} shows the rate-distortion performance of RAW image compression. As seen, even simRAW-pretrained lossy RIC (i.e., Lossy RIC-\SI{1}{\percent}) largely outperforms state-of-the-art 12-bit VVC. 
When we use more real RAW images to finetune the pretrained lossy RIC, the compression efficiency is further improved as reported for Lossy RIC-\SI{50}{\percent} and Lossy RIC-\SI{100}{\percent} that provide 2--3 dB PSNR gains over \SI{8}{bits} HEVC Intra, 1--2 dB gains over \SI{12}{bits} VVC Intra across a wide bit rate range. 

Figure~\ref{figure:lossy_compression_visual} presents the reconstructions and closeups generated by the HEVC Intra, VVC Intra, and our Lossy RIC-\SI{100}{\percent}. As seen, our method noticeably improves the subjective quality with sharper textures and lesser  noise.

\begin{table}
    \small
    \setlength{\tabcolsep}{1.6pt}
    \centering
    \caption{\edited{\textbf{Ablation of modular components of Unpaired CycleR2R}. KL divergence was assessed between real and simulated RAW images. Additionally, Recall and Average Precision (AP) metrics were evaluated using various RAW-domain detectors. These detectors were trained on simRAW images, generated by adapting modular settings in Unpaired CycleR2R, including Auto White Balance (AWB), Brightness Adjustment (BA) and Color Correction (CC), Illumination Estimation Module (IEM) and variance loss $\mathcal{L}_{var}$.}}

    \label{table:ablation_r2r_component}
    \begin{tabular}{lcccccccc}
    \toprule
        \textbf{Method} & \textbf{AWB} & \textbf{BA} & \textbf{CC} & \edited{\textbf{IEM+$\mathcal{L}_{var}$}}
        & \edited{\textbf{KL}} & \textbf{Recall} & \textbf{AP}    \\ \midrule
        \edited{InvGamma}~\cite{koskinen2019reverse} 
        & - 
        & - 
        & - 
        & - 
        & \edited{0.25}
        & 68.6
        & 47.7
        \\ \midrule
        \edited{CycleISP}~\cite{zamir2020cycleisp} 
        & - 
        & - 
        & - 
        & - 
        & \edited{0.08}
        & 71.6
        & 52.7
        \\ \midrule
        \edited{CIE-XYZ Net}~\cite{afifi2020cie} 
        & - 
        & - 
        & - 
        & - 
        & \edited{0.13}
        & 72.2
        & 53.0
        \\ \midrule
        \edited{MBISPLD}~\cite{conde2022model} 
        & - 
        & - 
        & - 
        & - 
        & \edited{0.11}
        & 73.0
        & 53.7
        \\ \midrule
        % \checkmark & & & & & 70.2 & 52.3\\ \midrule
        \multirow{4}{*}{Unpaired CycleR2R} 
        & \checkmark 
        & 
        & 
        & 
        & \edited{0.12}
        & 71.9 
        & 53.6\\ 
        
        & \checkmark 
        & \checkmark 
        & 
        & 
        & \edited{0.10}
        & 72.5 
        & 54.9 \\  
        
        & \checkmark 
        & \checkmark 
        & \checkmark 
        & 
        & \edited{0.09}
        & 74.9 
        & 56.1 \\
        
        & \checkmark 
        & \checkmark 
        & \checkmark 
        & \checkmark
        & \edited{\textbf{0.06}}
        & \textbf{76.1}  
        & \textbf{59.1} \\
    \bottomrule
    \end{tabular}
\end{table}

{{\bf Comparative studies of Lossless RIC.}} {Table~\ref{table:lossless_compression} shows a comparison of our approach and other lossless codecs for the test samples in MultiRAW set, in terms of the bits per pixel (BPP). 
For PNG, JPEG XL, and FLIF designed for RGB images, we use the same evaluation strategy by splitting it into two images for compression. 
First, our Lossless RIC-\SI{1}{\percent} trained on simRAW$_\textrm{f}$ only already outperforms PNG (a widely-used RGB compressor) with 35--\SI{74}{\percent} reduction in BPP and CinemaDNG (a professional RAW compressor) with 45--\SI{96}{\percent} reduction in BPP, for three different cameras. Compared with JPEG XL and FLIF, our method offers up to \SI{48}{\percent} reduction in BPP, but our method runs much faster for both encoding and decoding. 
Furthermore, the small gap of ($<\SI{2}{\percent}$) between Lossless RIC-\SI{100}{\percent} and Lossless RIC-\SI{1}{\percent} reveals that our Unpaired CycleR2R is capable of characterizing and embedding sufficient RAW-domain knowledge using a very small amount of real RAW images. Furthermore, the simRAW images produced by our invISP are able to train a generic lossless RAW image compressor. This makes our solution very attractive for practical situations with limited real training data.}

\section{Ablation Studies}
{In this section, we first provide additional discussion on the modular components of invISP. Then we study the impact of gamma correction and illumination on RAW-domain detection. Finally, we presents the progressive decoding ability of our lossless RIC and $\rho$-Vision's hardware implementation.}

\subsection{Modular Components of Unpaired CycleR2R}

\edited{Table~\ref{table:ablation_r2r_component} employs the Kullback-Leibler (KL) divergence alongside Recall and Average Precision (AP) to evaluate the impact of modular components within the Unpaired CycleR2R framework. The KL divergence assesses the distribution similarity between the color channels of simulated (simRAW) and real captured RAW images, with lower values indicating higher resemblance. This statistical measure is crucial for determining the realism of simRAW images utilized for training vision models.}
% Assuming a discrete histogram distribution, the KL divergence is computed as follows:}
% \edited{
%     \begin{align}
%     D_{KL}(P || Q) = \frac{1}{N} \sum_{k=1}^{N} \sum_{x} P_k(x) \log\left(\frac{P_k(x)}{Q_{i(k)}(x)}\right), \label{eq:avg_KL_divergence} 
%     \end{align} {with } $i(k)$ = $\argmin\nolimits_i \left\lVert P_k - Q_i \right\rVert$. 
% }
% \edited{$P$ represents the histogram distributions of simulated RAW images, while $Q$ corresponds to those of real RAW images. The index $i(k)$ is determined for each simRAW image $k$ by finding a real RAW image whose histogram $Q_i$ is closest to $P_k$. The formula calculates the KL divergence for each simRAW image against its closest-matched real RAW image. The average of these KL divergence values across all images ($N$) is then computed, providing an overall realism measure of simRAW images to the real RAW data, which is crucial in assessing whether simRAW images are suitable for training downstream tasks for practical inference.}
\edited{In addition to the KL divergence, Recall, and AP metrics are calculated by executing the task upon the same real RAW dataset. Here, various detectors are trained on simRAW images by activating or deactivating certain modules of our model.} These metrics demonstrate that each module significantly influences task accuracy. The Illumination Estimation Module (IEM), in particular, markedly increases the AP from \SI{56.1}{\percent} to \SI{59.1}{\percent}. This improvement is credited to IEM's advanced simulation of diverse lighting conditions, which is more representative of real-world scenarios than traditional methods that often rely on fixed illumination settings for ISP/invISP modeling. We give some visualization samples in the Supplementary, showcasing the module's contribution to producing training data that closely mirrors genuine imaging conditions.

\begin{figure*}
\begin{center}
\begin{tabular}[b]{c@{ } c@{ } c@{ } c@{ } c@{ }}
    \includegraphics[width=.19\textwidth,valign=t]{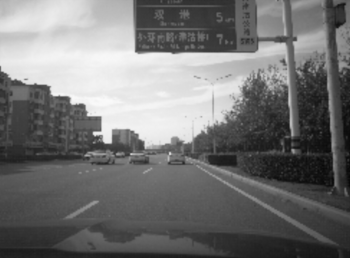}&
  	\includegraphics[width=.19\textwidth,valign=t]{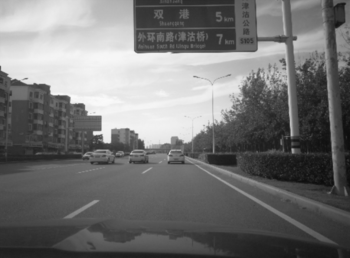}&   
    \includegraphics[width=.19\textwidth,valign=t]{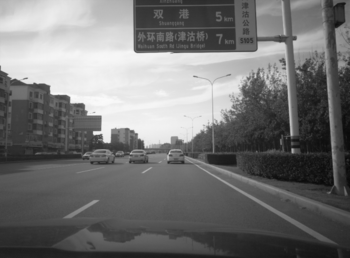}&
    \includegraphics[width=.19\textwidth,valign=t]{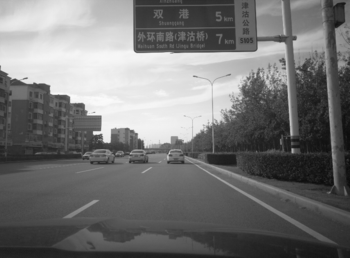}&
     \includegraphics[width=.19\textwidth,valign=t]{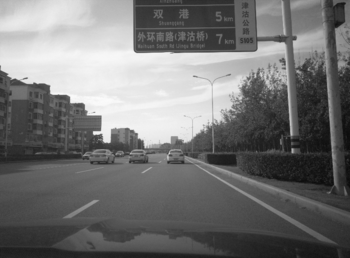} \\

\SI{0.08}{bpp} / \SI{36.77}{\dB} &
\SI{0.14}{bpp} / \SI{40.07}{\dB} &
\SI{0.38}{bpp} / \SI{44.13}{\dB} &
\SI{1.34}{bpp} / \SI{49.96}{\dB} &
\SI{5.13}{bpp} / GT 
\vspace{+0.3cm}
\\
    \includegraphics[width=.19\textwidth,valign=t]{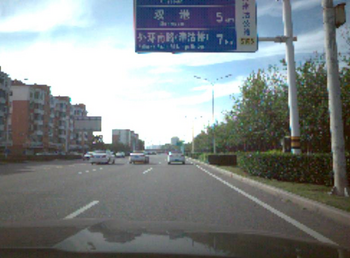}&
  	\includegraphics[width=.19\textwidth,valign=t]{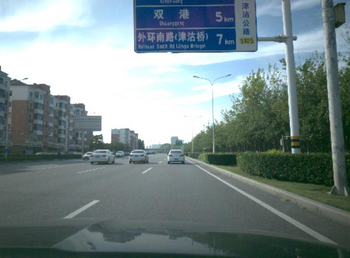}&   
    \includegraphics[width=.19\textwidth,valign=t]{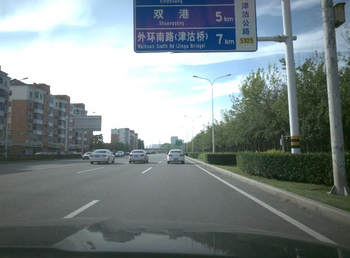}&
    \includegraphics[width=.19\textwidth,valign=t]{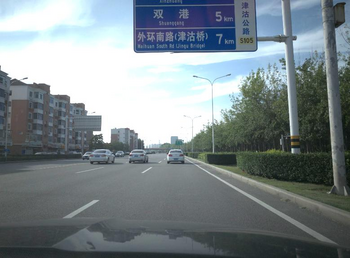}&
     \includegraphics[width=.19\textwidth,valign=t]{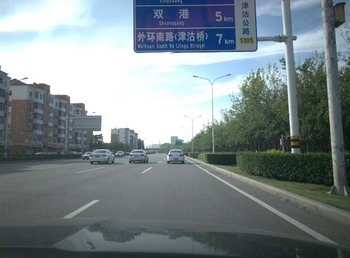} \\
      
\SI{0.15}{\second} / \SI{28.67}{\dB} &
\SI{0.26}{\second} / \SI{31.43}{\dB} &
\SI{0.37}{\second} / \SI{34.60}{\dB} &
\SI{0.52}{\second} / \SI{38.24}{\dB} &
\SI{0.68}{\second} / GT \\
    
\end{tabular}
\end{center}
\caption{{\bf Progressive Decoding.} The gradual reconstruction of RAW images and their corresponding RGB images converted by an in-camera ISP. Bits per pixel (bpp) / PSNR (dB) is shown under RAW images. Decoding latency (s) / PSNR (dB) is also listed below RGB images. PSNR is derived against the GT (ground truth). Gamma correction and brightness adjustment have been applied for a better view. {\it Zoom for more details.}}
\label{figure:progressive_decoding}
\end{figure*}

\subsection{Gamma Correction}
\begin{table}
    \small
    \setlength{\tabcolsep}{4.6pt}
    \scriptsize
    \caption{\textbf{Impact of Gamma Correction (GC) on object detection for different cameras and different object classes.}}
    \label{table:ablation_hdr_split}
    
    \begin{threeparttable}
    \begin{tabular}{lccccccc}
    \toprule
\textbf{Domain} & 
    \textbf{Camera} &
    \textbf{Vehicle} &
    \textbf{Person} &
    \textbf{Tr. Sign} &
    \textbf{Tr. Light} &
    \textbf{mAP} \\ \midrule
    RGB & 
    \multirow{3}{*}{\makecell[c]{i-12}} &
    82.6 &
    34.9 &
    72.7 &
    54.0 &
    \textbf{61.1}  \\
    RAW w/o GC &  
    &
    79.2 &
    25.3 &
    70.2 &
    48.6 &
    55.8 \\
    RAW w/ GC &                  
    & 
    81.1 &
    33.5 &
    73.8 &
    55.0 &
    60.8 \\
    \midrule
% RAW (w/ HDR-split) & 
%     &
%     82.1 & 30.3 & 73.6 & 56.0 & 60.5 \\
% RAW (learnt Adaptive Gamma) & 
%     &
%     81.3 & 34.1 & 72.8 & 58.8 & \textbf{61.7} \\ \midrule
    RGB &
    \multirow{3}{*}{\makecell[c]{HW-12 }} &
    75.3 &
    38.4 &
    62.2 &
    49.5 &
    56.4  \\
    RAW w/o GC &  
    &
    74.3 &
    33.7 &
    61.4 &
    49.4 &
    54.7 \\
    RAW w/ GC &
    & 
    75.5 &
    39.4 &
    62.0 &
    50.9 &
    \textbf{57.0} \\
    \midrule
% RAW (w/ HDR-split) &  
%     &
%     75.5 & 37.7 & 60.8 & 49.5 & 55.9 \\
% RAW (learnt Adaptive Gamma) & 
%     &
%     75.7 & 38.3 & 62.2 & 51.2 & 56.8 \\ \midrule
    RGB & 
    \multirow{3}{*}{\makecell[c]{asi-14 }} &
    76.2 &
    34.0 &
    66.4 &
    60.7 &
    58.3  \\
    RAW w/o GC &  
    &
    60.2 &
    18.5 &
    49.7 &
    46.8 &
    43.8 \\
    RAW w/ GC&                  
    & 
    79.0 &
    42.2 &
    71.8 &
    63.8 &
    \textbf{64.2} \\
% RAW (w/ HDR-Split)&                  
%     & 
%     77.5 & 38.6 & 67.5 & 60.0 & 60.9 \\
% RAW (learnt Adaptive Gamma) & 
%     &
%     78.2 & 45.3 & 71.3 & 62.0 & 64.2 \\ 
    \bottomrule
    \end{tabular}
    
    \begin{tablenotes}
        \footnotesize
        \item i-12 $\leftarrow$ iPhone XSmax (\SI{12}{bits}); HW-12 $\leftarrow$ Huawei P30pro (\SI{12}{bits}); asi-14 $\leftarrow$ asi 294mcpro (\SI{14}{bits}).
        \item \textbf{Tr. Sign} $\leftarrow$ Traffic Sign; \textbf{Tr. Light} $\leftarrow$ Traffic Light.
    \end{tablenotes}
    
    \end{threeparttable}
    
\end{table}
\normalsize
\label{sec:ablation_gamma}
{We prove analytically in Sec.~\ref{sec:hdr_split} that directly feeding linear RAW images into the detector typically leads to inferior detection performance. We then suggest the gamma correction approach adjusts the input distribution and subsequently boosts the performance of RAW-domain detection. Here we offer a quantitative evaluation of the gamma correction. 

For a fair comparison, we train three detectors sharing the same head (YOLOv3~\cite{yolo}) and backbone (MobileNetV2~\cite{sandler2018mobilenetv2}) using the training samples in the proposed MultiRAW dataset. Specifically, we train an RGB-domain detector (i.e., RGB in Table~\ref{table:ablation_hdr_split}) using the RGB images that are converted from the RAW samples with in-camera ISP, and use it to test RGB images as well. We also train two RAW-domain detectors where one option, RAW w/ GC in Table~\ref{table:ablation_hdr_split}, applies the gamma correction to adjust training RAW samples prior to the training, and the other one, RAW w/o GC in Table~\ref{table:ablation_hdr_split}, keeps using the same RAW samples without change.

As seen, without gamma correction, the performance of the RAW-domain detector is inferior to that of the RGB-domain detector with a noticeable gap. The gamma correction significantly improves the detection results even exceeding the RGB-domain model, further confirming the analytical proof in Sec.~\ref{sec:hdr_split}. 
The performance improvement is larger for asi 294mcpro camera sensor that has a 14-bit dynamic range, suggesting that fine-grained spatial details in shadow and highlight parts can be well retained in high-bit-precision RAW samples for better object detection.}

\subsection{\edited{Illumination Influence on Detection Accuracy}}
\label{sec:illumination-impact}
\begin{table}
    \small
    \setlength{\tabcolsep}{3.5pt}
    \centering
    \caption{\edited{\textbf{Comparative analysis of object detection performance under different illumination conditions across various object sizes.}}}
    \label{table:ablation_illumination}
    
    \edited{
    \begin{tabular}{lccccc}
    \toprule
    \textbf{Scenarios} &
    \textbf{Domain} & 
    $\textbf{mAP}_\text{small}$ &
    $\textbf{mAP}_\text{medium}$ &
    $\textbf{mAP}_\text{large}$ &
    \textbf{mAP} \\ \midrule
    \multirow{2}{*}{Day} 
    & RGB
    & 19.1
    & 54.9
    & 73.6
    & 58.5 \\
    
    & RAW
    & \textbf{21.2} 
    & \textbf{57.8}
    & \textbf{74.7}
    & \textbf{60.5} \\
    \midrule
    \multirow{2}{*}{Night} 
    & RGB 
    & 17.0
    & 49.1
    & 63.1
    & 56.1 \\

    & RAW
    & \textbf{28.2}
    & \textbf{54.5}
    & \textbf{67.7}
    & \textbf{59.8} \\
    \bottomrule
    \end{tabular}
    }
\end{table}

% \newcolumntype{M}[1]{>{\centering\arraybackslash}p{#1}}

\newcommand{\AddDetImg}[2]{\includegraphics[width=#1\textwidth]{#2}}
\newcommand{\AddDetImgs}[2]{
\AddDetImg{0.28}{Figs/detection_visual_compare/#1/#2/ldr_raw.png} &
\AddDetImg{0.28}{Figs/detection_visual_compare/#1/#2/hdr_raw.png} &
\AddDetImg{0.28}{Figs/detection_visual_compare/#1/#2/rgb.png} \vspace{+6pt}
}

\begin{figure*}
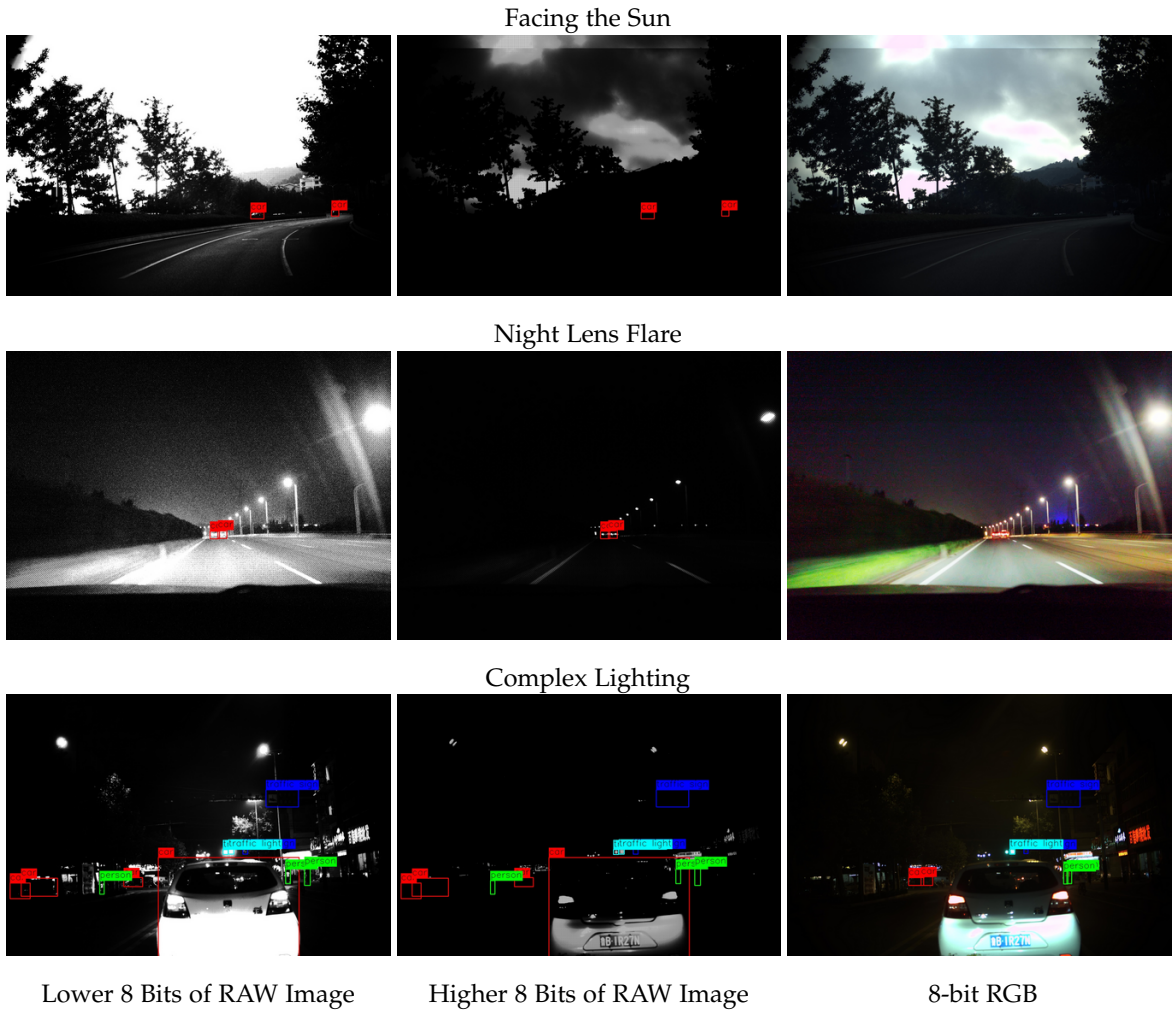

\begin{center}
% \begin{tabular}{p{0.155\textwidth}@{ } p{0.155\textwidth}@{ } p{0.155\textwidth}@{ } p{0.155\textwidth}@{ } p{0.155\textwidth}@{ } p{0.155\textwidth}@{ }}
\begin{tabular}{c@{ } c@{ } c@{ }}
\multicolumn{3}{c}{\edited{Facing the Sun}} \\
\AddDetImgs{face_sun}{2021-09-11-0800_0-CapObj_0000} \\
\multicolumn{3}{c}{\edited{Night Lens Flare}} \\
\AddDetImgs{glare}{IMG_20210911_185558} \\
\multicolumn{3}{c}{\edited{Complex Lighting}} \\
\AddDetImgs{incorrect_exposure}{2021-09-11-1146_8-CapObj_0000} \\
Lower 8 Bits of RAW Image & Higher 8 Bits of RAW Image & 8-bit RGB 
\\
\end{tabular}
\end{center}
\vspace{-5pt}
\caption{\edited{\textbf{Visualizing the detection results under challenging illumination conditions for the RAW domain and RGB domain methods.}  The first column displays the lower 8 bits of the RAW data, which retains the shadow details often lost in standard 8-bit RGB images due to brightness clipping. The second (middle) column presents the higher 8 bits of the RAW data, which contains the details in the bright areas. The third column uses the 8-bit RGB for detection. Detection results are overlaid for comparative study. }}%{\it Zoom for better details.}}}}
\vspace{-10pt}

\label{figure:ablation_detection_compare_visual}
\end{figure*}

\edited{The influence of illumination conditions on object detection accuracy was systematically evaluated across various object scales. The performance metrics, detailed in Table~\ref{table:ablation_illumination}, reveal notable improvements when using RAW data for detection, especially under complex nighttime conditions (+3.7). This improvement is particularly pronounced for small objects (+11.2), where the RAW format's extended dynamic range facilitates the discernment of details often lost in standard 8-bit RGB images due to brightness clipping or overexposure.}

\edited{Figure~\ref{figure:ablation_detection_compare_visual} further visualizes the detection results under various challenging illumination scenarios, such as direct sunlight, night lens flare, and complex lighting with multiple sources. 
% We split the lower 8-bit and higher 8-bit intensity values of the RAW image for visualization. In contrast, results using standard 8-bit RGB images are also provided. Note that even for an HDR RGB image, it would be tone-mapped to 8-bit for display on commodity platforms. 
As seen, RAW-domain processing can better preserve details in both bright and shadow areas which are typically overlooked in standard 8-bit RGB images due to brightness clipping or overexposure. 
%data to emphasize details in the bright areas, which. 
% The side-by-side representation showcases the enhanced dynamic range present in RAW images and its crucial role in improving detection robustness.
}

\subsection{Progressive Decoding}
 Our lossless RIC model supports progressive decoding to refine the image resolution gradually, which enables the low-resolution preview of high-resolution RAW and RGB images (converted by the in-camera ISP). Such a prompt low-resolution preview (see the visualization in Fig.~\ref{figure:progressive_decoding}) is a useful add-on function for applications like professional photography.  As also shown in Fig.~\ref{figure:progressive_decoding}, we can observe the restoration of high-frequency details gradually. Another useful takeaway of such progressive decoding is its inherent network transmission friendliness, with which we can still decode partial bitstream received at the client for display or consumption~\cite{1037998}.  

\subsection{Hardware Implementation}
We deployed our $\rho$-Vision framework on the Axera-Tech AX620A SoC, which facilitated a direct comparison between RAW-domain visual computing and conventional ISP processing methods. Utilizing YOLOv8-S and the MultiRAW dataset as our benchmark, the experimental outcomes underscore the superiority of $\rho$-Vision. Specifically, we observed a 3\% increase in detection accuracy, coupled with substantial reductions in latency (72\%), power consumption (62\%), and memory usage (36\%). Notably, these enhancements were achieved without necessitating complex modifications to the network model architecture. The benefits of our approach become even more evident under challenging conditions, such as in low-light and high dynamic range scenarios. Detailed results and further analyses can be found in our supplementary materials.

 \section{Conclusion}\label{sec:discussion_conclusion}
%\xzhang{
%In this paper, we explore and verify the feasibility of executing high-level computer vision tasks and low-level image compression in RAW domain.  Specifically, the proposed Unpaired CycleR2R based on the GAN model can learn and capture the cross-camera characteristics, by which abundant RGB image datasets can be utilized for RAW-domain model training towards any target camera deployed in practice. Extensive simulations have demonstrated its encouraging prospect. For example, RAW-domain-based object detection has provided competitive accuracy to the conventional methods in the RGB domain. The proposed solution is attractive to the applications with stringent latency and bandwidth requirements, such as the autonomous driving.
%}

%\iffalse
In this paper, we demonstrate that performing high-level vision tasks and low-level image compression on the camera RAW images is practically feasible and efficient.  In this way, the ISP modules, which are incorporated into cameras for decades, can be completely bypassed, thus promising an alternative and encouraging paradigm for image/video acquisition, processing, and display.
Our Unpaired CycleR2R is able to effectively characterize, embed, and transfer necessary knowledge between unpaired RGB and camera RAW samples to build the mapping between RGB and RAW spaces in an unsupervised manner. This enables us to conveniently generate sufficient simRAW images to retain/finetune popular RGB-domain neural models deployed in existing products for RAW-domain task executions. We present extensive experiments on high-level object detection and low-level image compression tasks in the RAW domain, which show better performance can be achieved in RAW domain compared to the RGB domain. 
One potential future direction is to develop similar models for processing RAW videos. 
% We will make our dataset and code publicly accessible for reproducible research upon the acceptance of this work at 
% \Salnote{Which other material will you release upon acceptance?} \zhihao{Fixed}
% {
We will make our MultiRAW dataset and Unpaired CycleR2R code publicly accessible for reproducible research at \url{https://njuvision.github.io/rho-vision} upon the acceptance of this work.
% Pretrained Unpaired CycleR2R models and our MultiRAW dataset are made available to the public immediately at \url{https://njuvision.github.io/rho-vision}.\url{https://njuvision.github.io/rho-vision}.
% }.

%through the use of modular unrolling based ISP and invISP models,

%abundant simRAW images can be generated using popular RGB image datasets 

%domain. Extensive simulations have revealed an encouraging prospect along this direction: for example, RAW-domain-based object detection has provided competitive accuracy to the conventional methods in RGB-domain.

%On one hand, our proposed Unpaired CycleR2R utilizes the GAN model to learn and capture the cross-camera characteristics, by which we can make use of abundant RGB image datasets for RAW-domain model training towards any target camera deployed in practice.

%On the other hand, by enabling such RAW-domain processing, we can completely bypass the ISP modules which are incorporated into cameras for decades. This promises an alternative and encouraging paradigm for image/video acquisition, processing, and display. Our proposed solution is attractive to those applications having stringent latency and bandwidth requirement, such as the autonomous driving.
%\fi

\section{Acknowledgement}

We are very grateful for those pioneering explorations in~\cite{afifi2020cie,conde2022model,zamir2020cycleisp} which inspire this work.

\bibliographystyle{IEEEtran}
\bibliography{tpami.bib}

\begin{IEEEbiography}[{\includegraphics[width=1in,height=1.25in,clip,keepaspectratio]{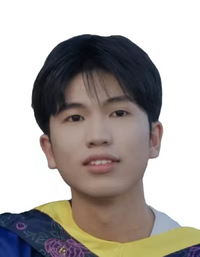}}]{Zhihao Li} (Member, IEEE) received his B.S. and Master degrees in the School of Electronic Science and Engineering from Nanjing University, Jiangsu, China, in 2020 and 2023 respectively. His current research focuses on raw image signal processing and event camera. He is a co-recipient of the 2nd International Illumination Estimation Challenge and the CVPRW 2022 Night Image Rendering Challenge Runner-up solution.
\end{IEEEbiography}

\begin{IEEEbiography}[{\includegraphics[width=1in,height=1.25in,clip,keepaspectratio]{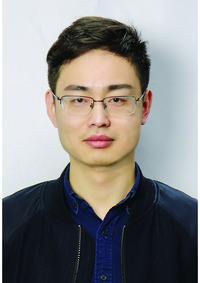}}]{Ming Lu} (Member, IEEE) is an Associate Researcher in the School of Electronic Science and Engineering, Nanjing University, Jiangsu, China. He received his B.S. and Ph.D. degrees in the School of Electronic Science and Engineering from Nanjing University, Jiangsu, China, in 2016 and 2023 respectively. His current research focuses on deep learning-based image/video coding. He is a co-recipient of the 2018 ACM SIGCOMM Student Research Competition Finalist, the 2020 IEEE MMSP Image Compression Grand Challenge Best Performing Solution, and the 2023 IEEE WACV Best Algorithms Paper Award.
\end{IEEEbiography}

\begin{IEEEbiography}[{\includegraphics[width=1in,height=1.25in,clip,keepaspectratio]{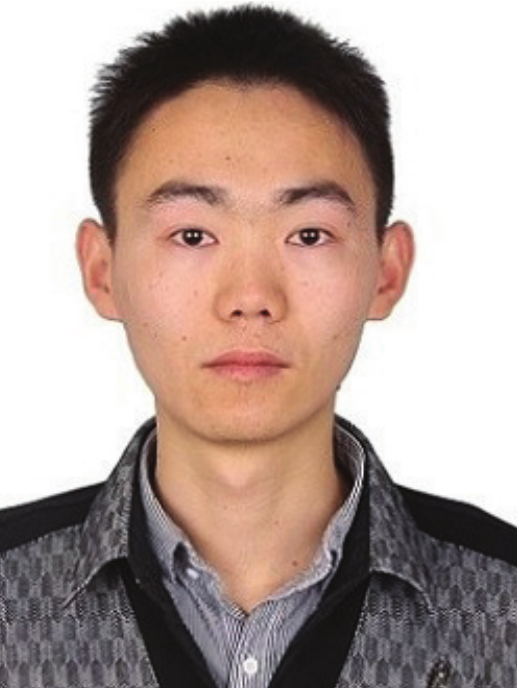}}]{Xu Zhang} (Member, IEEE) is a Lecturer with the School of Computing, Faculty of Engineering and Physical Sciences, University of Leeds, LS2 9JT, United Kingdom. He received the BS degree in communication engineering from the Beijing University of Posts and Telecommunications, China, in 2012 and the Ph.D. degree in computer science from the Department of Computer Science and Technology, Tsinghua University, China, in 2017. 
He worked at Nanjing University and the University of Exeter before joining the University of Leeds.
His research interests include AI for computing and networking, multimedia communication, and Internet of Things. He was a recipient of the EU Marie Skłodowska-Curie Individual Fellowships and a co-recipient of the 2019 IEEE Broadcast Technology Society Best Paper Award.
\end{IEEEbiography}

\begin{IEEEbiography}[{\includegraphics[width=1in,height=1.25in,clip,keepaspectratio]{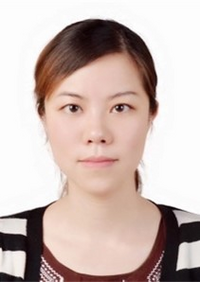}}]{Xin Feng} (Senior Member, IEEE) received her B.S. degree in computer science and technology from Chongqing University in 2004. She conducted her postgraduate study from 2004 to 2006 and got the Ph.D. degree in computer applications from Chongqing University in 2011. She is currently an associate professor at Chongqing University of Technology. She studied at New York University as a postdoctoral from 2014 to 2016. Her research falls in the area of computer vision, image, and video processing. Her current research interests cover object detection, and multi-object tracking for visual perception in auto-driving.
\end{IEEEbiography}

\begin{IEEEbiography}[{\includegraphics[width=1in,height=1.25in,clip,keepaspectratio]{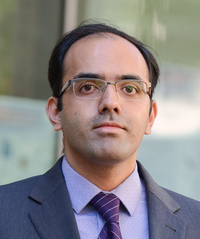}}]{M. Salman Asif} (Senior Member, IEEE) received the B.Sc. degree from the University of Engineering and Technology, Lahore, Pakistan, and the M.S and Ph.D. degrees from the Georgia Institute of Technology, Atlanta, GA, USA. He is currently an associate professor with the University of California Riverside, Riverside, CA, USA. Prior to that, he was a Postdoctoral Researcher with Rice University, Houston, TX, USA, and a Senior Research Engineer with Samsung Research America, Dallas. His research interests include computational imaging, signal/image processing, computer vision, and machine learning. He was the recipient of the NSF CAREER Award, Google Faculty Award, Hershel M. Rich Outstanding Invention Award, and UC Regents Faculty Fellowship and Development Awards.
\end{IEEEbiography}

\begin{IEEEbiography}[{\includegraphics[width=1in,height=1.25in,clip,keepaspectratio]{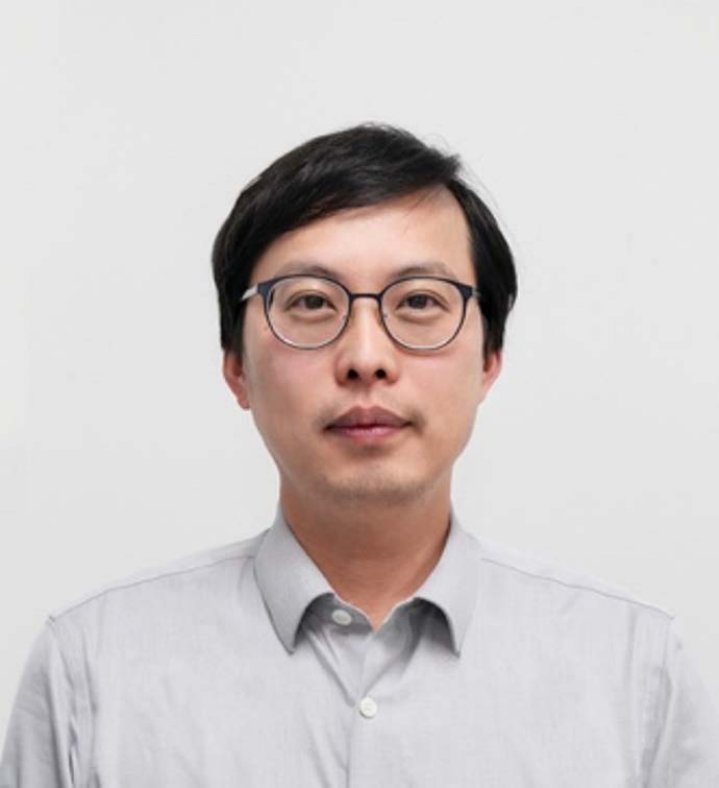}}]{Zhan~Ma} (Senior Member, IEEE) is a Professor in the School of Electronic Science and Engineering, Nanjing University, Jiangsu, 210093, China. He received his Ph.D. from New York University, New York, in 2011 and his B.S. and M.S. from the Huazhong University of Science and Technology, Wuhan, China, in 2004 and 2006 respectively. From 2011 to 2014, he has been with Samsung Research America, Dallas, TX, and  Futurewei Technologies, Inc., Santa Clara, CA, respectively. His research focuses include learned image/video coding and computational imaging. He was awarded the 2018 PCM Best Paper Finalist, the 2019 IEEE Broadcast Technology Society Best Paper Award, the 2020 IEEE MMSP Grand Challenge Best Image Coding Solution, the 2023 IEEE WACV Best Algorithms Paper Award, and the 2023 IEEE Circuits and Systems Society Outstanding Young Author Award.
\end{IEEEbiography}

\end{document}

\makeatletter\@input{supp_material_aux.tex}\makeatother
\title{
Supplemental Materials - Efficient Visual Computing with Camera RAW Snapshots}

\author{Zhihao Li,
        Ming Lu,
        Xu Zhang,
        Xin Feng,
        M. Salman Asif,
        and
        Zhan Ma}

\maketitle

\begin{abstract}
In this supplementary material, we provide additional information to further evidence the generalization of the proposed \textit{{$\rho$-Vision}} for various functionalities. \edited{Specifically, we first compare the RGB-Vision and $\rho$-Vision frameworks using a real-world hardware implementation in Sec.~\ref{sec:hardware_implementation}. Then, we provide details of our Unpaired CycleR2R in Sec.~\ref{sec:details_of_nn} and give proofs of some equations in Sec.~\ref{sec:proofs}. In addition, we demonstrate the advantages of running classification and segmentation in the RAW domain directly in Sec.~\ref{sec:raw_cls} and Sec.~\ref{sec:raw_seg}, respectively. At last, we show more visualization results in Sec.~\ref{sec:more_figures}.}
\end{abstract}

\begin{IEEEkeywords}
Camera RAW, RAW-domain Object Detection, RAW Image Compression
\end{IEEEkeywords}

\IEEEpeerreviewmaketitle
\maketitle

\begin{figure}[ht]
\centering
\edited{
\subfloat[Hardware Platform]{\includegraphics[width=0.32\linewidth]{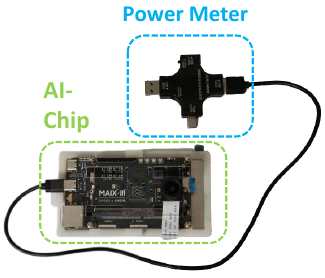} \label{fig:hardware_platform}}
\subfloat[$\rho$-Vision Framework] {\includegraphics[width=0.58\linewidth]{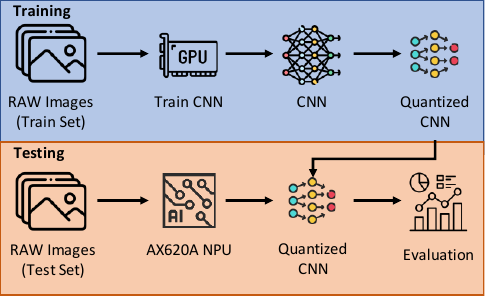} \label{fig:hardware_raw_pipeline}}\\
\subfloat[RGB-Vision Framework]{\includegraphics[width=0.92\linewidth]{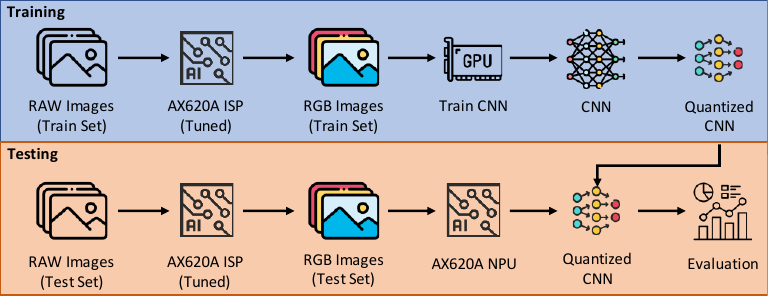} \label{fig:hardware_rgb_pipeline}}\\
\subfloat[Average Gains]{\includegraphics[width=0.9\linewidth]{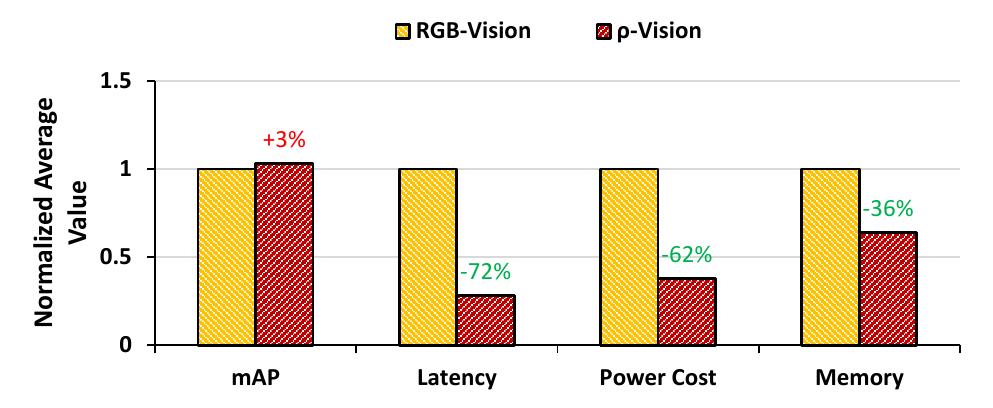} \label{fig:supp_hardware_results}}
%\end{center}
%\caption{}
\caption{\edited{{\bf RGB-Vision vs. $\rho$-Vision.} (a) The hardware system uses AX620A AI SoC. A UC96B power meter is connected for measurement; (b) $\rho$-Vision framework trains and tests models using RAW images directly, completely bypassing the ISP; (c)  Traditional RGB-Vision framework requires the ISP to generate RGB images for model training and testing; (d) {\bf Average Gains of $\rho$-Vision to RGB-Vision.} Metrics are normalized to the results generated by the RGB-Vision pipeline.}} %with mAP scores tested using the same testing samples. Latency refers to the total time elapsed when executing the entire pipeline, power cost is measured using a power meter, and memory usage is recorded using the default monitoring application.}
} % end of edited
\label{fig:supp_hardware_implementation}
\end{figure}

\section{\edited{A Real-World Hardware Implementation}}
\label{sec:hardware_implementation}
\subsection{\edited{Hardware System for Comparative Benchmark}}
\label{sec:hardware_platform}

\edited{A commodity hardware platform is used to assess the efficiency of RAW-domain visual computing as illustrated in Fig.~\ref{fig:hardware_platform}. It is built upon the Axera-Tech AX620A SoC with a quad-core Arm Cortex-A7 processor, an NPU (Neural Processing Unit), an ISP (Image Signal Processor), and other subsystems. This AX620A SoC is primarily used to process images and videos for vision tasks. Its ISP has two modes: one is the Standard mode (AX620A ISP), and the other is the AI mode (AX620A AI ISP). When using AX620A AI ISP, onboard NPU is utilized to run various neural algorithms like NN (Neural Network) denoising, by which AX620A SoC claims its outstanding performance for low-light imaging.

%for image/video tasks. 

%which is designed for various image and video tasks. This automotive quality chip has a Quad A7 CPU, a powerful 3.6 Tops (int-8) NPU capable, and two types of image signal processing (ISP) units. While one ISP unit focuses on standard processes, the other, known as the AI-ISP, uses neural algorithms for low-light scenery. When AI-ISP is used, it shares the NPU computation resource with vision tasks. 

We use the same RAW samples in the MultiRAW dataset for a fair evaluation. The YOLOv8-S, recommended by the AX620A SoC specification, exemplifies the detection task. Its default settings are assumed for consistency and reproducibility. Upon completing the training of YOLOv8-S, its model is quantized into INT-8 precision using AX620A's official quantization tool and subsequently deployed on AX620A's NPU for inference.

Metrics such as mAP, latency, power consumption, and memory usage are collected for quantitative comparison. With this aim, when executing the YOLOv8-S, a UC96B power meter is connected to the AX620A SoC to collect the power usage, latency is measured using a timer library (C++), and the memory consumption is reported using the default memory monitoring tool provided by the AX620A SoC.

%documented to facilitate a comparative analysis of accuracy and computational resource expenditure between the RGB-Vision and $\rho$-Vision frameworks.

\begin{figure*}
\small
\centering
\edited{
\begin{tabular}{p{0.22\linewidth} p{0.22\linewidth} p{0.22\linewidth} p{0.22\linewidth}}
  \centering\includegraphics[width=\linewidth]{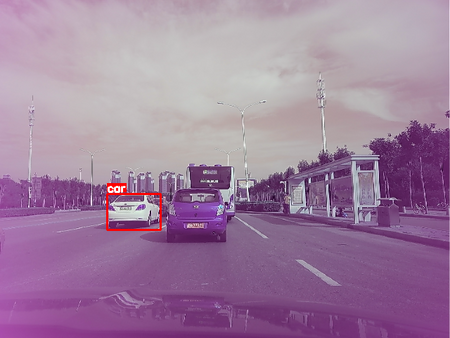} &
  \centering\includegraphics[width=\linewidth]{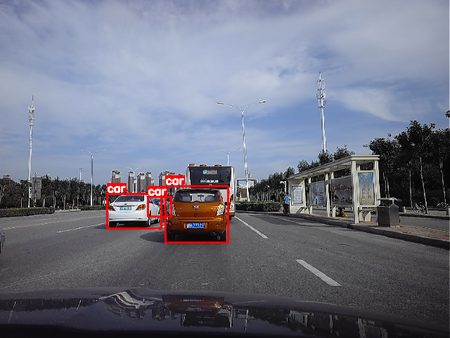} &
  \centering\includegraphics[width=\linewidth]{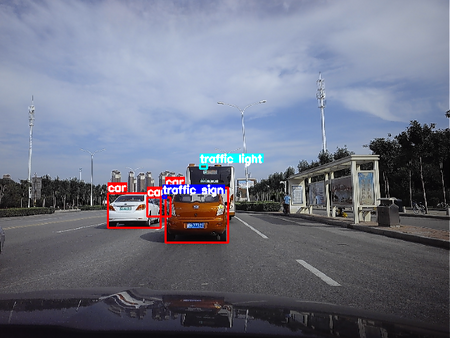} &
  \centering\includegraphics[width=\linewidth]{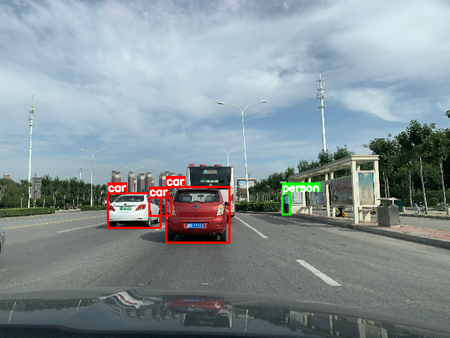} \tabularnewline
  
  \centering\includegraphics[width=\linewidth]{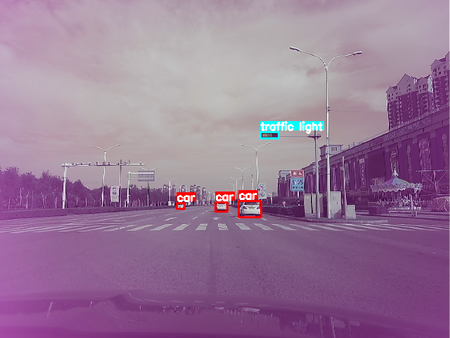} &
  \centering\includegraphics[width=\linewidth]{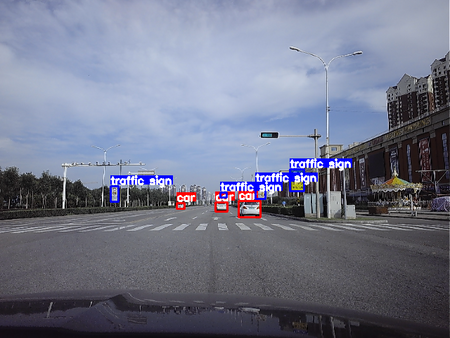} &
  \centering\includegraphics[width=\linewidth]{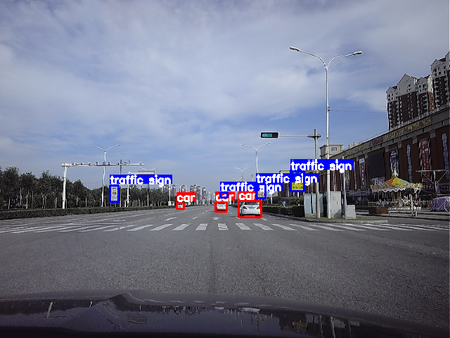} &
  \centering\includegraphics[width=\linewidth]{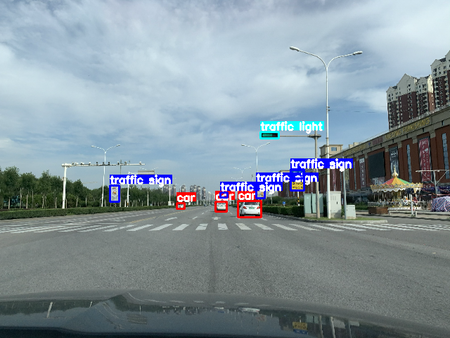} \tabularnewline

    \centering\includegraphics[width=\linewidth]{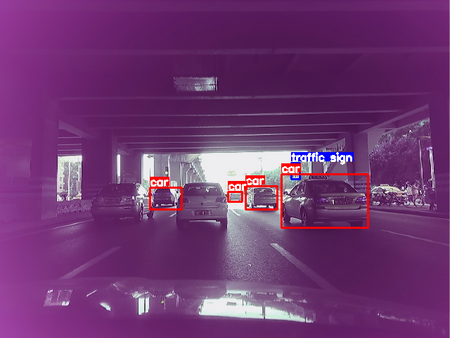} &
  \centering\includegraphics[width=\linewidth]{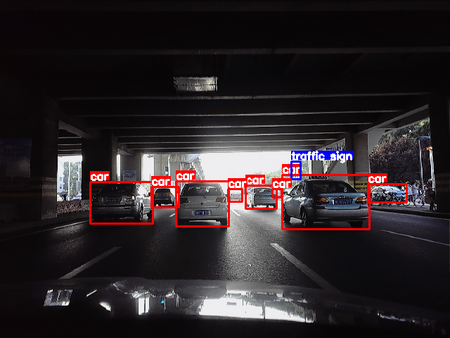} &
  \centering\includegraphics[width=\linewidth]{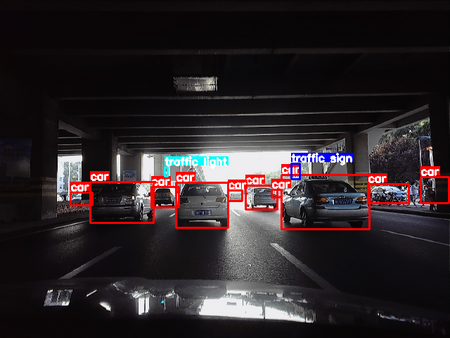} &
  \centering\includegraphics[width=\linewidth]{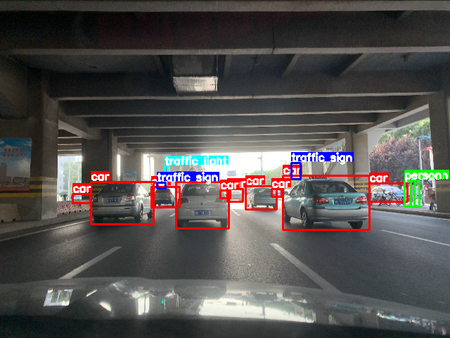} \tabularnewline
  \centering iPhone $\rightarrow$ AX620A (D) &
  \centering iPhone $\rightarrow$ AX620A (T) &
  \centering AX620A (T) $\rightarrow$ AX620A (T) &
  \centering iPhone $\rightarrow$ iPhone
\end{tabular}
}
\caption{\edited{{\bf Impact of ISP used in RGB-Vision on the detection task.} The setup of ``Training ISP$\rightarrow$Testing ISP'' indicates the ``Training ISP'' used to generate RGB images for training and the ``Testing ISP'' used to generate RGB images for testing respectively. Default parameters used by the ISP are marked with ``(D)'' and expert-tuned parameters used by the ISP are annotated with ``(T)''. The first two columns illustrate domain discrepancies when training and testing using different ISPs, while the last two columns demonstrate how ISP quality (with expert tuning) affects object detection accuracy. {\it Zoom for better details.}}}
\label{fig:supp_ax620a_iphone_results}
\end{figure*}

\begin{table*}[t]
    \normalsize
    \setlength{\tabcolsep}{6pt}
    \centering
    \caption{\edited{Detection performance for various ISP combinations.}}
    \label{table:supp_AX620A_iphone_results}
    
    \edited{
    % \begin{threeparttable}
    \begin{tabular}{lllccccc}
    \toprule
     \textbf{Domain}&\textbf{Training ISP} & 
    \textbf{Testing ISP} &
    \textbf{Car} &
    \textbf{Person} &
    \textbf{Traffic Light} &
    \textbf{Traffic Sign} &
    \textbf{mAP} \\ \midrule
     \multirow{4}{*}{RGB-Vision} &iPhone & AX620A (Default) & 0.324 & 0.022 & 0.134 & 0.213 & 0.173   \\
     &iPhone & AX620A (Tuned) & 0.696 & 0.108 & 0.523 & 0.253 & 
     0.397  \\
     &AX620A (Tuned) & AX620A (Tuned) & 0.788 & 0.225 & 0.661 & 0.443 & 0.529  \\
     &iPhone & iPhone & \textbf{0.798} & 0.219 & \textbf{0.693} & 0.474 & \textbf{0.546}  \\
     \midrule
     $\rho$-Vision &-& - & 0.796 & \textbf{0.241} & 0.655 & \textbf{0.490} & 
    \textbf{0.546}   \\
    \bottomrule
    \end{tabular}%\\
    % mAP(Q) stands for the results using the quantized model.
    % \end{threeparttable}
    }
\end{table*}

\begin{itemize}
    \item {\bf $\rho$-Vision} trains YOLOv8-S using RAW samples (from the iPhone XSmax, a subset of the MultiRAW dataset). Then, such a RAW-domain YOLOv8-S is quantized using the abovementioned rules and deployed on the NPU for detection. For task inference, RAW images are fed directly to the neural model (without requiring ISP computations). Following the common practice, \SI{70}{\percent} RAW images are used to train RAW-domain YOLOv8-S, and the remaining \SI{30}{\percent} RAW images are tested using quantized YOLOv8-S on NPU. Fig.~\ref{fig:hardware_raw_pipeline} plots the processing steps in $\rho$-Vision.
    \item {\bf RGB-Vision} applies the AX620A ISP onboard to convert RAW images to their corresponding RGB formats for subsequent computations. The training and testing split is the same as in the $\rho$-Vision paradigm. The RGB-vision processing pipeline is pictured in Fig.~\ref{fig:hardware_rgb_pipeline}.
\end{itemize}

%For the RGB-Vision framework hardware configuration (see Fig.~\ref{fig:hardware_rgb_pipeline}), the AX620A chip's integrated ISP was employed to generate training and testing image sets with uniformity. This strategy was specifically selected to mitigate the domain variability that distinct ISP architectures may introduce. A comprehensive discussion on this is presented in Sec.~\ref{sec:impact_of_isp}. 

{All associated hardware drivers, system images, benchmark code, and datasets will soon be available at \url{https://njuvision.github.io/rho-vision} to encourage reproducible research.}
} % end of edited 

%We use the MultiRAW dataset (iPhone RAW) for both training and evaluation. Unlike the RGB-Vision framework configuration, the $\rho$-Vision framework shown in Fig.~\ref{fig:hardware_raw_pipeline} eliminates the ISP stage, inputting RAW images directly into the system.
%\input{sections/figures/supp_hardware_results}

\edited{
\subsection{Experimental Analysis}

{\bf Overall Evaluation.} Fig.~\ref{fig:supp_hardware_results} showcases the efficacy of the proposed $\rho$-Vision paradigm. Compared to RGB-Vision, it provides a notable \SI{3}{\percent} detection accuracy increase. The same YOLOv8-S is just retrained using RAW images without any dedicated network model engineering. It reduces the latency by \SI{72}{\percent}, a critical advancement for autonomous driving applications. Furthermore, the \SI{62}{\percent} reduction in power consumption presents significant advantages of $\rho$-Vision for AIoT devices, where energy efficiency is crucial. The \SI{36}{\percent} decrease in memory usage also enables the deployment of $\rho$-Vision on lower-cost embedded devices. {The performance improvement owes to better-preserving scene information in the RAW domain. The skipping of ISP generally avoids the extra computations and memory caching, leading to a noticeable cost and latency reduction.} These promise the encouraging potential of $\rho$-Vision in advancing computer vision applications for better task performance, faster response, and less cost.

%with  combining precision with cost and energy efficiency.

{\bf Impact of ISP used in RGB-Vision Paradigm.} In Fig.~\ref{fig:hardware_rgb_pipeline}, the AX620A ISP is expert-tuned. This is because default settings used in AX620A ISP cannot provide a decent result, which motivates us to study the impact of various ISP configurations on task efficiency. The ISP used in the iPhone XSmax is also evaluated as Apple experts deliberately calibrate it for outstanding quality. Note that the ISP is only required in the RGB-Vision framework.

Similarly, we use iPhone RAW images from the MultiRAW dataset in experiments. We have different ISP combinations for RGB-Vision to train and test RGB images (converted from the same set of iPhone RAWs). The training and testing split is the same for either RGB-domain or RAW-domain processing. 

As in Table~\ref{table:supp_AX620A_iphone_results} for the RGB-Vision category, the training ISP converts iPhone RAW images to the corresponding RGB samples to train YOLOv8-S, while the testing ISP is used to generate RGB samples (from iPhone RAW images) for testing previously trained YOLOv8-S.

The setup using the same iPhone ISP to generate RGB images for training and testing provides the best performance (see the last row of RGB-Vision in Table~\ref{table:supp_AX620A_iphone_results}). Although we have tried our best to fine-tune the AX620A ISP to mimic the iPhone ISP, the setup using the same AX620A ISP (Tuned) to generate RGB images for training and testing is inferior to the case using the iPhone ISP that is deliberately calibrated by Apple imaging experts, e.g., 0.529 vs. 0.546 mAP. The detection performance is sharply degraded if we use different ISPs to generate training and testing RGB samples (see 1st and 2nd rows of Table~\ref{table:supp_AX620A_iphone_results} in RGB-Vision), suggesting that the ISP configuration is vital for task performance.

%which owes to the ISP difference.  
%Even through we apply the expert-tuning to improve AX620A ISP, its performance 
%Having expert-tuned AX620A can improve the

Fig.~\ref{fig:supp_ax620a_iphone_results} visualizes detection results on testing images, further confirming the observations in Table~\ref{table:supp_AX620A_iphone_results} where inappropriate use of ISPs would lead to catastrophic performance degradation (see missing objects in the first column).

%improved performance. 

%indicating a modest enhancement in detection accuracy. 

%Importantly, the RGB-Vision framework's results are contingent upon ISP quality, an aspect that is elaborated upon in Section \ref{sec:impact_of_isp}.

%More notably,

%\subsection{Impact of ISP}
%\label{sec:impact_of_isp}

%The influence of ISP configurations on model accuracy is substantially demonstrated in Table~\ref{table:supp_AX620A_iphone_results} and Fig.~\ref{fig:supp_ax620a_iphone_results}. Disparities in ISPs during training and testing phases introduce domain discrepancies that are directly reflected in model performance. As shown in the first two rows of Table~\ref{table:supp_AX620A_iphone_results} and the corresponding sections of Fig.~\ref{fig:supp_ax620a_iphone_results}, these discrepancies can lead to a pronounced drop in accuracy. Despite efforts to calibrate the AX620A ISP (marked as Tuned) to approximate the iPhone ISP's characteristics, a significant mAP gap of over 14.7 points remains when tested against the actual iPhone ISP.

%Furthermore, ISP quality critically affects model performance, particularly in challenging lighting conditions that demand high dynamic range capabilities. As observed in the last row of Fig.~\ref{fig:supp_ax620a_iphone_results}, the AX620A's ISP inadequately renders darker regions, which leads to the non-detection of objects, such as persons that are crucial for real-world applications.

By contrast, under the $\rho$-Vision setup, YOLOv8-S is trained and tested on iPhone RAW images directly. The average detection performance is the same as using the iPhone ISP for both training and testing in RGB-vision. More importantly, expert tuning or dedicated calibration of ISP is no longer required. All of these suggest the encouraging prospects of using $\rho$-Vision in vision tasks.

%framework processes RAW images directly, ensuring consistent performance irrespective of ISP quality and lighting conditions. While Table~\ref{table:supp_AX620A_iphone_results} shows that the RAW domain performance is slightly lower than that of the iPhone ISP, it is important to note that the greater bit-depth of RAW images contributes to larger quantization errors upon deployment. Specifically, the RAW domain exhibited an mAP of 54.6 pre-quantization, aligning with the iPhone ISP's accuracy. However, post-quantization, the iPhone ISP's mAP performance decreased by only 1.1 points, whereas the RAW's decreased by 1.6 points. This aspect of quantization error due to increased bit-depth in the RAW domain is a subject for our future research endeavors.

{\bf Challenging Imaging Conditions} are additionally examined to compare the efficiency of $\rho$-Vision and RGB-Vision pipelines. Two representative contexts are considered: the low-light illumination with high-noise levels and the scenario with high dynamic range (HDR) conditions.
} % end of edited

%\subsection{Comparative Analysis of Challenge Scenarios}
\newcommand{\RGBImageNet}{$\text{RGB}_{\text{IN}}$}
\newcommand{\RGBpixel}{$\text{RGB}_{\text{AX}}$}
\newcommand{\RGBpixelpixel}{$\text{RGB}_{\text{GP}}$}
\newcommand{\RGBpixelAI}{$\text{RGB}_{\text{AX-AI}}$}
\newcommand{\RAWpixel}{$\text{RAW}_{\text{GP}}$}
\newcommand{\pseudoRAWpixel}{$\text{RAW}_{\text{IN}}$}

\begin{table*}
    \normalsize
    \setlength{\tabcolsep}{2.5pt}
    \centering
    \caption{\edited{\textbf{Classification Accuracy of RGB-Vision and $\rho$-Vision Frameworks Under Low-Light Conditions.} Latency measures the total processing duration by both the ISP and model, as well as the power consumption (Power.) and memory requirements (Mem.) for each method, besides the Top-1 classification accuracy (Acc.). *The results of Google Pixel ISP are copied from the paper~\cite{diamond2021dirty}. The ``invISP" is used in $\rho$-Vision to generate simulated RAW samples to train the classifier, while RGB-vision methods do not require this step.  RGB-Vision methods train the RGB-domain classifier using RGB images from the ImageNet dataset (\RGBImageNet) while $\rho$-Vision trains the RAW-domain classifier using simulated RAW images generated using the invISP.  RAW images acquired by Google Pixel (\RAWpixel)~\cite{diamond2021dirty} under extreme low-light conditions are used for evaluation. In the RGB-Vision pipeline, these RAW images are converted using different ISPs to RGB samples for using the RGB-domain classifier, while in the $\rho$-vision paradigm, these RAW images are directly fed to the RAW-domain classifier.}%Simulated \RAWpixel (sim\pseudoRAWpixel) images were synthesized using our proposed Unpaired CycleR2R method. % For testing, images underwent processing via the AX620A's standard ISP (\RGBpixel) as well as its advanced AI-ISP variant (\RGBpixelAI), which transforms \RAWpixel images. 
    }
    \label{table:ax620a_dirtypix_results}
    \edited{
    \begin{tabular}{lcccccccccccr}
    \toprule
    \multirow{2}{*}{\textbf{Method}} 
    &
    & \textbf{invISP} 
    &
    & \multicolumn{2}{c}{\textbf{Classifier}} 
    &
    
    & \multicolumn{2}{c}{\textbf{Latency}} 
    
    &
    & \multirow{2}{*}{\textbf{Power.}} 
    & \multirow{2}{*}{\textbf{Mem.}}
    & \multirow{2}{*}{\textbf{Acc.}} 
    \\ \cmidrule{3-3} \cmidrule{5-6} \cmidrule{8-9}

    &
    & \textbf{Train}
    &
    & \textbf{Train} 
    & \textbf{Test} 
    &
    & \textbf{ISP} 
    & \textbf{Model} 
    &
    &
    &
    &
    \\ \midrule
    RGB-Vision w/  AX620A ISP 
    &
    & - 
    &
    & \RGBImageNet 
    & \RGBpixel 
    &
    & 48.65 ms 
    & 2.73 ms 
    &
    & 0.128 J 
    & 65 MB
    & 0.0
    \\
    RGB-Vision w/ AX620A AI-ISP 
    &
    & - 
    &
    & \RGBImageNet 
    & \RGBpixelAI
    &
    & 64.75 ms 
    & 4.36 ms 
    &
    & 0.162 J
    & 81 MB
    & 0.0
    \\
    RGB-Vision w/ *Google Pixel ISP 
    &
    & - 
    &
    & \RGBImageNet 
    & \RGBpixelpixel
    &
    & -
    & - 
    &
    & -
    & -
    & 1.4
    \\
    $\rho$-Vision
    &
    & \RGBImageNet, \pseudoRAWpixel 
    &
    & sim\pseudoRAWpixel 
    & \RAWpixel 
    &
    & \textbf{0 ms}
    & \textbf{2.71 ms} 
    &
    & \textbf{0.006 J}
    & \textbf{25 MB}
    & \textbf{19.8}
    \\ \bottomrule
    \end{tabular}
    }
\end{table*}

\newcommand{\RAWlucid}{$\text{RAW}_{\text{LT}}$}
% \newcommand{\pseudoRAWpixel}{$\text{RAW}_{\text{P}}$}
% % \newcommand{\Baselines}{\textcolor{blue}{$\spadesuit$~}}
% % \newcommand{\InvISP}{\textcolor{orange}{$\blacksquare$~}}
% % \newcommand{\DomainAdap}{\textcolor{green}{$\blacklozenge$~}}
% % \newcommand{\Ours}{\textcolor{red}{$\bigstar$~}}

\begin{table*}
    \normalsize
    \setlength{\tabcolsep}{5pt}
    \centering
    \caption{\edited{\textbf{Comparative Analysis of RGB-Vision and $\rho$-Vision Frameworks in High Dynamic Range (HDR) Scenarios.} The RAW-domain detector is calibrated with 24-bit LUCID TRI054S RAW images (\RAWlucid). The RGB-domain detector is trained and evaluated on RGB images generated using AX620A ISP (\RGBpixel). Latency encompasses the total processing time of both the ISP and the detection model. We present the power consumption (Power.) and memory footprint (Mem.) alongside the mean Average Precision (mAP). Abbreviations ``Tr. L.'' and ``Tr. S.'', denote traffic light and traffic sign, respectively.}}
    \label{table:ax620a_lucid_results}
    
    \edited{
    \begin{tabular}{lccccccccccccc}
    \toprule
    \multirow{2}{*}{\textbf{Framework}} 
    &
    & \multicolumn{2}{c}{\textbf{Detector}} 
    &
    
    & \multicolumn{2}{c}{\textbf{Latency}} 
    
    &
    & \multirow{2}{*}{\textbf{Power.}} 
    & \multirow{2}{*}{\textbf{Mem.}}
    & \multirow{2}{*}{\textbf{$\text{AP}_{\text{Car}}$}} 
    & \multirow{2}{*}{\textbf{$\text{AP}_{\text{Tr. L}}$}} 
    & \multirow{2}{*}{\textbf{$\text{AP}_{\text{Tr. S}}$}} 
    & \multirow{2}{*}{\textbf{mAP}} 
    \\ 
    \cmidrule{3-4} \cmidrule{6-7}

    &
    & \textbf{Train} 
    & \textbf{Test} 
    &
    & \textbf{ISP} 
    & \textbf{Model} 
    &
    &
    &
    &
    \\ \midrule
    RGB-Vision
    &
    & \RGBpixel 
    & \RGBpixel 
    &
    & 48.55 ms 
    & \textbf{17.07 ms }
    &
    & 0.152 J 
    & 55 MB
    & 81.3
    & 27.9
    & 61.2
    & 56.8
    \\
    $\rho$-Vision
    &
    & \RAWlucid 
    & \RAWlucid 
    &
    & \textbf{0 ms}
    & 18.18 ms 
    &
    & \textbf{0.058 J}
    & \textbf{35 MB}
    & \textbf{84.8}
    & \textbf{35.5}
    & \textbf{69.7}
    & \textbf{63.3}
    \\ \bottomrule
    \end{tabular}
    }
\end{table*}

%Beyond the typical scenarios outlined in Sec.~\ref{sec:hardware_platform}, in order to further ascertain the efficacy of RGB-Vision and $\rho$-Vision frameworks under extreme operational conditions, we conducted evaluations in scenarios characterized by exceedingly low illumination coupled with high noise, as well as environments demanding extensive dynamic range.
\edited{
{\it Low-light illumination with high noise} scenario is evaluated with object classification. We closely follow~\cite{diamond2021dirty} to perform the task, which involves training a MobileNet-V1 %(architecture with an input size of 224x224, 
using noise-augmented ImageNet samples, then testing real-world noisy images acquired using a Google Pixel camera under low-light/high-noise conditions.

As for RGB-Vision, we directly train an RGB-domain MobileNet-V1 using the ImageNet dataset (RGB$_{\rm IN}$) (with noise augmentation). In the meantime, we respectively use AX620A ISP and AX620A AI-ISP to transform RAW images acquired using Google Pixel camera (RAW$_{\rm GP}$) to the corresponding RGB datasets, e.g., RGB$_{\rm AX}$ and RGB$_{\rm AX-AI}$ to test aforementioned RGB-domain MobileNet-V1.

As for $\rho$-Vision, we first train our Unpaired CycleR2R model using clean RAW and RGB images from the Google Pixel and ImageNet datasets, i.e., RAW$_{\rm GP}$ and RGB$_{\rm IN}$, respectively. Then, we use the invISP module in this Unpaired CycleR2R to convert RGB images in ImageNet to simulated RAW samples, i.e., simRAW$_{\rm IN}$, to train the RAW-domain MobileNet-V1. 
The same noise augmentation is performed upon simRAW$_{\rm IN}$.
Such a RAW-domain MobileNet-V1 tests RAW samples directly from RAW$_{\rm GP}$.

%In the case of RGB, In contrast, for RAW processing, The ImageNet RGB images were then transformed into simulated RAW, with noise added, to train a RAW-domain MobileNet-V1, which was tested on the Google Pixel RAW dataset.

Evaluations presented in Table~\ref{table:ax620a_dirtypix_results} clearly evidence the superiority of $\rho$-Vision paradigm. Notable reductions are reported for power consumption, memory footprint, and computational latency, owing to removing the ISP subsystem in the proposed $\rho$-Vision framework.

%illustrate several gains from utilizing the RAW-domain approach. Firstly, there is a notable advantage in power consumption and latency with RAW processing. In particular, the 
$\rho$-Vision only requires 0.006 J for task inference, compared to 0.128 and 0.162 J consumed by RGB-Vision methods using AX620A ISP and AX620A AI ISP. Furthermore, it exhibits the lowest latency at 2.71 ms, a substantial decrease from the 48.65 ms and 64.75 ms observed with the methods using AX620A ISP and AX620A AI ISP. This is because small-size images, e.g., 224$\times$224, are used in the classifier, but ISPs must process images with the original resolution (2560$\times$1440). Such a sharp increase in data volume increases power consumption, memory footprint, and latency. 

%benefit arises despite the network's input size of 224$\times$224, as the highly hardware-dependent ISP still needs to process images of the, leading to redundant power usage and delays. It is important to note that while the ISP can achieve 30 FPS, this rate is achieved through a parallel pipeline architecture, which does not imply that latency can be controlled within 1/30 of a second. 
$\rho$-Vision also presents better classification accuracy. We attribute it to noise separation and suppression in the RAW domain being more tractable than in the RGB domain (after a serial nonlinear transformation)~\cite{zhu2017unpaired}. 

%The second factor that leads to superior performance in the RAW domain is that our Unpaired CycleR2R model generates sufficiently diverse and realistic RAW images. However, it is important to note that, compared to the results shown in Table~\ref{table:supp_wo_labeled_classification}, the performance degradation observed in both the RGB and RAW domains when using int-8 quantization is primarily due to the substantial bias introduced by int-8 PTQ, which arises from the influence of high input noise on feature distribution. This is a critical consideration in such scenarios. 
Notably, the AX620A AI ISP does not enhance classification performance under such extreme low-light conditions, as AX620A AI ISP models are typically trained for some specific cameras and may not generalize well to a new camera from the above discussions. %under severely low-light conditions.

{\it HDR conditions} are studied with the detection task. 24-bit LUCID TRI054S RAW images (\RAWlucid) covering the tunnel exit scenes are used. These HDR scenes are often encountered when driving through the tunnel and simultaneously experiencing extraordinarily bright and dark regions.

%within our MultiRAW dataset.

As for $\rho$-Vision, we train the RAW-domain detector (YOLOv8-S) using \RAWlucid. In contrast, RAW samples in \RAWlucid are first converted to RGB counterparts using the AX620A ISP to train the RGB-domain detector used in the RGB-Vision framework. 

% The compared RGB-domain detector was trained and tested on images processed through the AX620A's standard ISP (\RGBpixel), which converts RAW images from the same tunnel scenarios. This process ensures that the comparison focuses on the detectors' performance rather than differences in ISP.

Besides the reductions in power consumption, memory footprint, and latency, the $\rho$-Vision framework achieves superior mAP across all categories, particularly in detecting traffic lights and signs (e.g., labeled as ``Tr. L.'' and ``Tr. S.'') in Table~\ref{table:ax620a_lucid_results}. The improvement in mAP indicates the enhanced capability of the $\rho$-Vision to discern features in HDR conditions. This is essential for applications such as autonomous driving, where accurate and prompt traffic detection is crucial.

The combination of reduced latency, lower power consumption, and memory usage, along with higher mAP scores, affirms the effectiveness of the $\rho$-Vision framework in challenging HDR scenarios, highlighting its potential for real-world applications where both performance and efficiency are of paramount importance.
} % end of edited

\section{Details of the Unpaired CycleR2R}
\label{sec:details_of_nn}

\subsection{Architecture of Basic Neural Network}
\label{sec:encoder_details}
Table~\ref{tab:supp_enc_arch} details the architecture of the basic neural network $E\left( \cdot \right)$ used in Unpaired CycleR2R. This basic network $E\left( \cdot \right)$ consists of five layers in total and is used for IEM (Illumination Estimation Module), AWB (Auto White Balance), BA (Brightness Adjustment), and CC (Color Correction). The first layer applies the 5$\times$5 convolution with 32 channels, and the subsequent two layers use 3$\times$3 convolutions and 64 channels. The final two layers use simple linear layers instead. 

The example of ``Conv: k5c32s2'' stands for a convolutional layer having convolutions with spatial kernel size at 5$\times$5 (k5), 32 channels (c32), and a stride of two based spatial downsampling (s2) at both dimensions. The same convention is applied to the linear layer (Linear) and average pooling layer (Avg Pool). ``Leaky RELU''~\cite{barron2017fast} is used as the activation, and ``Mean'' stands for the average operator in the spatial domain for each channel. Considering the output channel of $E\left( \cdot \right)$ is specific for different purposes across aforementioned modular components, we mark it using a predefined variable C$_\textrm{out}$.

\subsection{Architectures of Discriminators}

As in the main paper, $D_\textrm{color}$ and $D_\textrm{bright}$ are applied to measure the
similarity between generated and real images. $D_\textrm{color}$ stacks five
convolutional layers with Leaky ReLU~\cite{barron2017fast} and $D_\textrm{bright}$ uses
five linear layers instead to process 1D grayscale histogram. Details of kernel size, channels, and strides are listed in Tabel.~\ref{tab:supp_enc_arch}.
\begin{table}[hbp]
    % \small
    \setlength{\tabcolsep}{6.8pt}
    \centering
    \caption{Network settings of Unpaired CycleR2R.}
    \label{tab:supp_enc_arch}
    \begin{tabular}{ccc}
    \toprule
        Basic Encoder $E\left(\cdot \right)$ & Discriminator $D_\textrm{color}$ & Discriminator $D_\textrm{bright}$ \\ \midrule
        Conv: k5c32s2                       & Conv: k4c64s2                     & Linear: c1024\\ 
        Leaky RELU                          & Leaky RELU                        & Leaky RELU\\
        Avg Pool: s2                        & Conv: k4c128s2                    & Linear: c1024\\
        Conv: k3c64s2                       & Leaky RELU                        & Leaky RELU\\
        Leaky RELU                          & Conv: k4c256s2                    & Linear: c256\\ 
        Avg Pool: s2                        & Leaky RELU                        & Leaky RELU \\
        Conv: k3c64s1                       & Conv: k4c512s2                    & Linear: c256\\
        Leaky RELU                          & Leaky RELU                        & Leaky RELU\\
        Mean                                & Conv: k4c1s2                      & Linear: c1\\
        Linear: c256                        & Mean                              & -\\
        Linear: cC$_\textrm{out}$            & -                                 & -\\
    \bottomrule
    \end{tabular}
\end{table}

\subsection{Gamma Correction Standard}
Gamma correction matches the non-linear characteristics of a display device or human perception~\cite{farid2001blind}.
We adopt the correction function recommended in ITU-R BT.~709 standard~\cite{stokes1996standard}, noted as $f_{g}$, which is widely used in commodity ISPs today~\cite{drago2003adaptive}.
\begin{align}
\boldsymbol{y} &= f_{g} \circ \boldsymbol{x}_{cc} \nonumber\\
\begin{split}
     &=\left\{
        \begin{aligned}
            &12.92\cdot \boldsymbol{x}_{cc}, &\boldsymbol{x}_{cc} \leq 0.00304,\\
            &1.055\cdot \boldsymbol{x}_{cc}^{1/2.4}-0.055, &\boldsymbol{x}_{cc}>0.00304.
        \end{aligned}
    \right.
\end{split}
\end{align}
Correspondingly, the inverse function $g_{g}$ is:
\begin{align}
\boldsymbol{x}_{cc} &= g_{g} \circ \boldsymbol{x}_{g} \nonumber\\
\begin{split}
      &=\left\{
        \begin{aligned}
            &\frac{\boldsymbol{y}}{12.92}, &\boldsymbol{y} \leq 0.04045,\\
            &\left(\frac{\boldsymbol{y}+0.055}{1.055}\right)^{2.4}, &\boldsymbol{y} > 0.04045.
        \end{aligned}
    \right.
\end{split}
\end{align}

\section{Details of {\it Distribution Analysis of RAW images}}
\label{sec:proofs}

\subsection{The proof of the equation \eqref{eq:var_gradient_weights}}
% \subsection{Expansion of $\frac{\partial \mathcal{L}}{\partial w}$}
\label{appendix:expansion_l_w}
We start from the loss function $\mathcal{L}$:
\begin{align}
    \mathcal{L} = \frac{1}{H\times W} \left( \boldsymbol{w} \ast \left( \boldsymbol{P}-0.5 \right) + \mathbf{b} - \hat{\vy} \right)^2,
\end{align}
where $\boldsymbol{w} \in \mathbb{R}^{S\times S}$ is the convolution kernel with kernel size $S$.

Then the partial derivative of $w \in \boldsymbol{w}$ could be formulated as:  
\begin{align}
    \begin{split}
     \frac{\partial \mathcal{L}}{\partial w} &= \frac{1}{H\times W} \sum^H_{j=0} \sum^W_{i=0} 
     2 \left( \vy_{ij} - \hat{\vy} \right) \left( \vx_{ij+mn} -0.5 \right)
    \end{split}
\end{align}
where $\vx_{ij+mn} \in \boldsymbol{P}$ and $mn$ is the shift position of $w$ according to the kernel center of $\boldsymbol{w}$. $\vy_{ij}$ is the convolution output at position $ij$. To calculate $\vy_{ij}$, we define $\boldsymbol{x}^{w}_{ij}$ as a window of $\boldsymbol{P}$ with the same size of $\boldsymbol{w}$ located at $ij$. Considering the similarity among adjacent pixels, for a neighborhood pixel of $\vx_{ij}$, \ie, $\vx_{neibor} \in \boldsymbol{x}_{ij}^{w}$, we have $\vx_{neibor} = \vx_{ij} + \delta$, where $\delta$ follows a Gaussian distribution with zero mean. Thus, $\vy_{ij}$ could be expanded as:
\begin{align}
    \vy_{ij}  
    = \left( \vx_{ij}-0.5 \right) \sum w + \vb + \sum w\delta.
\end{align}
For simplify, we use $\tilde{\boldsymbol{x}}$ and $\tilde{\mu}$ to replace $\boldsymbol{x}-0.5$ and $\mu-0.5$, respectively. Besides, we set $A=\sum w, ~B=\vb + \sum w\delta, ~C=2A,~D=2\left( B - \hat{\vy} \right)$.
Having $\vy_{ij}=A\tilde{\vx} + B$, the $\frac{\partial \mathcal{L}}{\partial w}$ will be:
\begin{align}
\begin{split}
\frac{\partial \mathcal{L}}{\partial w}
    &= 2 \mathbb{E} \left[ \left( \vy - \hat{\vy} \right) \left(\vx - 0.5 \right) \right] \\
    &= 2 \mathbb{E}  \left[ \left( A \tilde{\vx} + B - \hat{\vy} \right) \tilde{\vx} \right]
    \\
    &= 2\mathbb{E} \left[ A\tilde{\vx}^2 + (B-\hat{\vy}) \tilde{\vx} \right] \\
    &= 2\mathbb{E} \left[A\right] \mathbb{E}  \left[ \tilde{\vx}^2 \right] + 2(\mathbb{E}  \left[ B \right] -\hat{\vy})  \mathbb{E} \left[ \tilde{\vx} \right] \\
    &= 2A (\tilde{\mu}^2 - \sigma^2) + 2(b-\hat{\vy}) \tilde{\mu} \\
     &= C (\tilde{\mu}^2 - \sigma^2) + D \tilde{\mu}.
\end{split}
\end{align}

% \subsection{Expansion of $\Var \left[ \frac{\partial \mathcal{L}}{\partial w} \right]$}
% \label{supply:expansion_var_l_w}
Since $\mu$ and $\sigma$ are independent and we only concern with the impact of $p\left(\mu\right)$, we set $\Var\left[ \sigma^2 \right]$ to a constant. Then the variance could be expanded as:
\begin{align}
\begin{split}
    \Var\left[ \frac{\partial \mathcal{L}}{\partial w} \right]
    &= \Var \left[ C \tilde{\mu}^2 + D \tilde{\mu} \right] + \Var\left[ C \sigma^2 \right]
    \\
    \begin{split}
        &= \mathbb{E} \left[ \left( C \tilde{\mu}^2 + D \tilde{\mu} \right)^2 \right] 
        - \mathbb{E} \left[C \tilde{\mu}^2 + D \tilde{\mu} \right]^2 
        \\
        &\quad + \textrm{const}
    \end{split}
    \\
    \begin{split}
        &= \mathbb{E}\left[ C^2 \tilde{\mu}^4 \right] + 
        \cancel{\mathbb{E}\left[ CD \tilde{\mu}^3 \right]} + 
        \mathbb{E}\left[ D^2 \tilde{\mu}^2 \right] \\
        &\quad - \left(  \mathbb{E}\left[ C \tilde{\mu}^2 \right] +  
        \cancel{\mathbb{E}\left[ D \tilde{\mu} \right]} \right)^2 
        + \textrm{const}
    \end{split}
    \\
    \begin{split}
        &= \mathbb{E}\left[ \left( C \tilde{\mu}^2 \right)^2 \right] 
        - \left( \mathbb{E}\left[ C \tilde{\mu}^2 \right] \right)^2 \\
        &\quad + \mathbb{E}\left[ \left( D \tilde{\mu} \right)^2 \right] 
        - \left( \mathbb{E}\left[ D \tilde{\mu} \right] \right)^2 
        + \textrm{const}
    \end{split}
    \\
    &= \Var\left[ C \tilde{\mu}^2 \right] +
    \Var\left[ D \tilde{\mu} \right] + 
    \textrm{const} \\
    &= C^2 \Var\left[ \tilde{\mu}^2 \right] + 
    D^2 \Var\left[ \tilde{\mu} \right] + 
    \textrm{const}.
    % + \mathbb{E}\left[ C^2 \mu^4 \right]
\end{split}
\end{align}

\subsection{The proof of the equation \eqref{eq:simp_var_gradient_weights}}

Given the $\mu$ following the distribution in \eqref{eq:k_quard}, the $\Var\left[ \tilde{\mu} \right]$ could be written as:
\begin{align}
    \begin{split}
    \Var\left[ \tilde{\mu} \right] &= \Var \left[ \mu -0.5 \right] = \Var \left[ \mu \right] \\
    &= \int^1_0 \left[ \mu - \mathbb{E}\left(\mu\right) \right]^2 p\left( \mu \right)  \dd \mu \\
    &= \int^1_0 \left( \mu - 0.5 \right)^2 (k\mu^2-k\mu+\frac{k}{6}+1) \dd \mu \\
    & = F(\mu=1) - F(\mu=0) \\
    & = (\frac{1}{21}-\frac{k}{720}) - (-\frac{k}{144}-\frac{1}{24}) \\
    & = \frac{k}{180} + \frac{1}{12},
    \end{split}
\end{align}
where $F\left(\mu\right)=k\left( \frac{\mu^5}{5} - \frac{\mu^4}{2} + \frac{\mu^3}{12} - \frac{\mu^2}{8} \right)+\frac{k}{18}\left(\mu - \frac{1}{2}\right)^3.$

Thus, $\Var\left[ \frac{\partial \mathcal{L}}{\partial w} \right]$ will be:
\begin{align}
    \begin{split}
        \Var\left[ \frac{\partial \mathcal{L}}{\partial w} \right] &\approx  D^2 \Var\left[ \tilde{\mu} \right] + 
    \textrm{const} \\
    &= D^2 \left( \frac{k}{180} + \frac{1}{12} \right) + \textrm{const} \\
    &= D^2 \frac{k}{180} + \textrm{const}.
    \end{split}
\end{align}

\section{\edited{RAW-domain Classification}}
\label{sec:raw_cls}
\newcommand{\RGBanscombe}{$\text{RGB}_{\text{Ans-ISP}}$}

\begin{table*}
    \small
    \setlength{\tabcolsep}{6pt}
    \centering
    \caption{
    {\edited{\textbf{Classification Accuracy On Google Pixel RAW images.}}
    %Classifier training was performed on RGB images from the ImageNet dataset (\RGBImageNet) and simulated RAW Pixel images (sim\pseudoRAWpixel), using the simple Mosaic or Unpaired CycleR2R method for synthesis. The evaluation involved testing the classifiers on \RAWpixel~ images after being processed by the respective inverse ISP techniques. The table reports Top-1 and Top-5 accuracy metrics, the number of parameters (\# Parameters), and the computational complexity in terms of FLOPs (Floating Point Operations).
    }
    }
    \label{table:supp_wo_labeled_classification}
    
    \edited{
    \begin{threeparttable}
    \begin{tabular}{lcrcccccccccc}
    \toprule
            \multirow{2}*{\textbf{Method}}
            & \textbf{invISP}
            &
            & \multicolumn{2}{c}{\textbf{Classifier}}
            &
            & \multirow{2}*{\textbf{Top-1 Acc.}}
            & \multirow{2}*{\textbf{Top-5 Acc.}}
            & \multirow{2}*{\textbf{\# Parameters}}
            & \multirow{2}*{\textbf{FLOPs}}
            \\\cmidrule{2-2}\cmidrule{4-5}
            
            & {\textbf{Train}}
            &
            & {\textbf{Train}}
            & {\textbf{Test}}
            &
            &
            &
            & 
            &
            \\ \midrule
         Anscombe ISP\tnote{*}~\cite{diamond2021dirty} 
        & -
        &
        &  \RGBanscombe
        & \RGBanscombe
        &  
        & 33.1
        & 58.4
        & 4.28
        & 282
        \\
        Mosaic RAW\tnote{*}~\cite{diamond2021dirty} 
        & -
        &
        & sim\pseudoRAWpixel
        & \RAWpixel        
        &  
        & 27.0
        & 52.5
        & \textbf{4.23}
        & \textbf{181}
        \\ \midrule
        \textbf{Unpaired CycleR2R} 
        & \RGBImageNet,~\RAWpixel
        &
        & sim\pseudoRAWpixel
        & \RAWpixel  
        &  
        & \textbf{35.5}
        & \textbf{72.1}
        & \textbf{4.23}
        & \textbf{181}
        \\
    \bottomrule
    \end{tabular}
    
    \begin{tablenotes}
        \item[*] Both the Anscombe ISP and Mosaic RAW apply simple mosaic operations to generate RAW samples from the corresponding RGB images. They don't need to train the invISP. 
    \end{tablenotes}
    % \vspace{2mm}
    % Such a mosaic operation is  referred to~\cite{}
\end{threeparttable}
}
\end{table*}

% \newcolumntype{M}[1]{>{\centering\arraybackslash}p{#1}}

\newcommand{\AddClsImg}[2]{\includegraphics[width=#1\textwidth]{#2}}
\newcommand{\AddClsImgs}[4]{
\multirow{-7}{*}{\rotatebox[origin=c]{90}{\textbf{#4}}} \quad&
\AddClsImg{0.145}{Figs/classification_zero_shot_visual/#1/#2_w_label.png} &
\AddClsImg{0.145}{Figs/classification_zero_shot_visual/#1_clean_input/#2_w_label.png} &
\AddClsImg{0.145}{Figs/classification_zero_shot_visual/#1/#2_Conv2d_0_channel_#3.png} &
\AddClsImg{0.145}{Figs/classification_zero_shot_visual/#1_clean_input/#2_Conv2d_0_channel_#3.png} &
\AddClsImg{0.145}{Figs/classification_zero_shot_visual/#1/#2_gradcam.png} &
\AddClsImg{0.145}{Figs/classification_zero_shot_visual/#1_clean_input/#2_gradcam.png}
}

\newcommand{\AddDirtypix}[1]{
\AddClsImgs{joint_real_2to200lux}{#1}{22}{RGB}
}

\newcommand{\AddinvISP}[1]{
\AddClsImgs{invISP_2to200lux}{#1}{29}{RAW}
}

\begin{figure*}
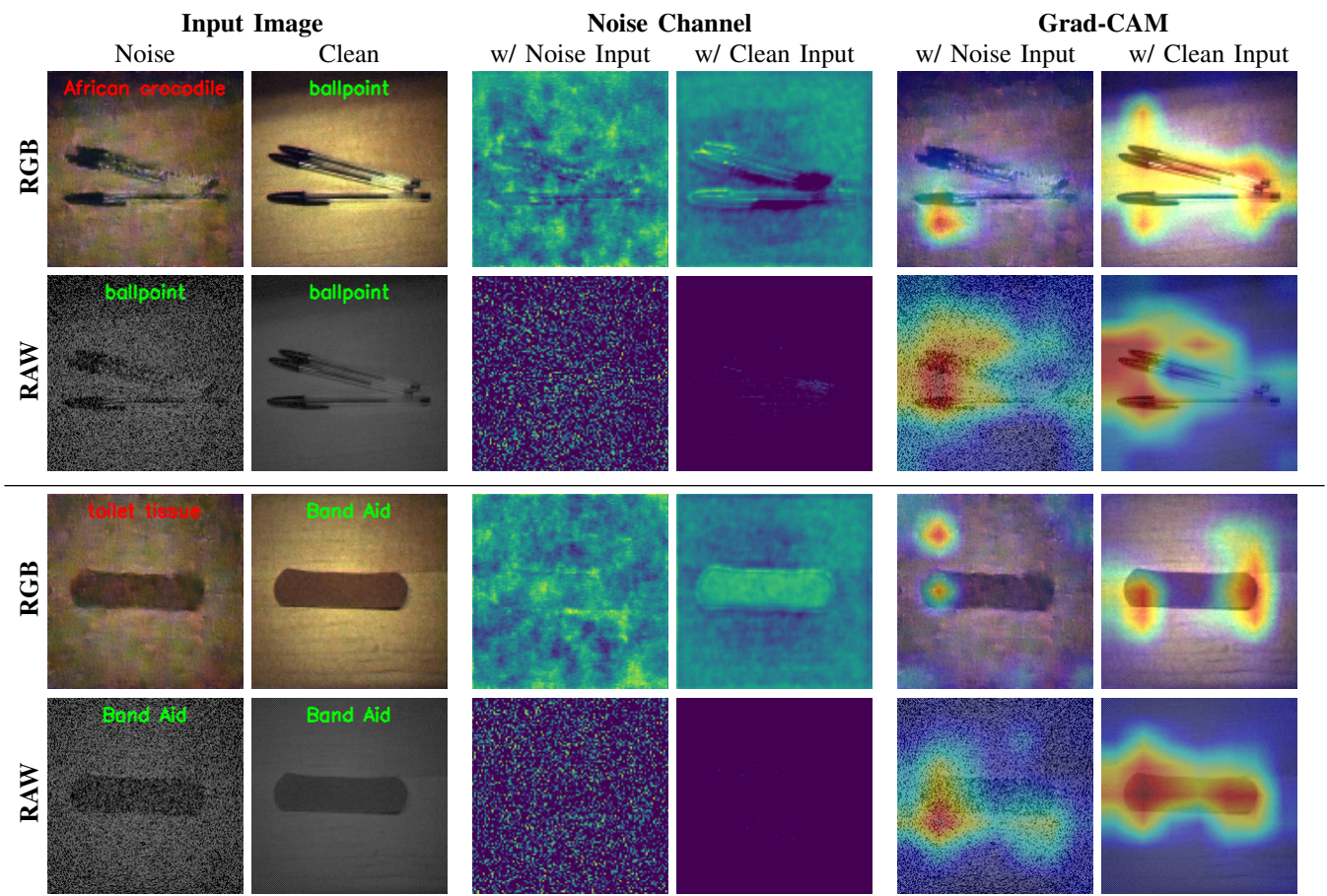

\begin{center}
% \begin{tabular}{p{0.145\textwidth}@{ } p{0.145\textwidth}@{ } p{0.145\textwidth}@{ } p{0.145\textwidth}@{ } p{0.145\textwidth}@{ } p{0.145\textwidth}@{ }}
\begin{tabular}{c@{} c@{ } c@{ } | c@{ } c@{ } | c@{ } c@{ }}
&
\multicolumn{2}{c}{\textbf{Input Image}}
& 
    \multicolumn{2}{c}{\textbf{Noise Channel}}

& 
    \multicolumn{2}{c}{\textbf{Grad-CAM}}
\\
& Noise
& Clean
& 
    w/ Noise Input
& 
    w/ Clean Input
&
    w/ Noise Input
& 
    w/ Clean Input
\\

\AddDirtypix{IMG_20180314_190359}
\\
\AddinvISP{IMG_20180314_190359}
\\ 
\midrule
\AddDirtypix{IMG_20180314_200348}
\\
\AddinvISP{IMG_20180314_200348}
\\ 

% \AddDirtypix{IMG_20180314_200620}
% \\
% \AddinvISP{IMG_20180314_200620}
% \\ 

% \makecell{noise input} & \makecell{clean input} & \makecell{noise channel \\ (w/ noise input)} & \makecell{noise channel \\ (w/ clean input)} & \makecell{grad-cam \\ (w/ noise input)} & \makecell{grad-cam \\ (w/ clean input)} \\

\end{tabular}
\end{center}
\caption{\edited{\textbf{Visualization of Classifier Response to Noisy and Clean Inputs} The ``RGB'' rows represent the processing using the Anscombe ISP~\cite{diamond2021dirty}   where it inputs the RGB image for classification; In contrast, the ``RAW'' rows stand for the processing using Unpaired CycleR2R where the RAW images are directly processed. Noise is augmented upon the clean inputs to form Noisy samples. 
 The ``Noise Channel" is the feature channel in the shallow layer ``Conv2d\_0" that presents the maximum difference when processing the noise and clean inputs respectively. The Grad-CAM~\cite{selvaraju2017grad} visualizations are based on the last convolutional layer ``Conv2d\_13\_pointwise". A comparison between the ``Noise Channel" under different inputs reveals that the RAW-domain classifier is adept at extracting noise patterns, effectively separating noise from the signal, which results in Grad-CAM visualizations that more closely resemble the clean input. In contrast, the RGB-domain classifier struggles to disentangle noise from the signal due to the complex non-linear processing by the Anscombe ISP, leading to significant deviations in Grad-CAM under noisy conditions and consequently to misclassification.}}
\label{figure:supp_classification_zero_shot_visual}
\end{figure*}

\renewcommand{\arraystretch}{1} % 调整这个值

\edited{
In this section, we present the application of our Unpaired CycleR2R model for the classification task in the RAW domain.

\subsection{Datasets and Baselines}
We utilize the identical dataset for training and testing as in~\cite{diamond2021dirty}.  For generating the training set, we use ImageNet~\cite{deng2009imagenet} to generate simulated RAW images with noises. As for testing, a real-world RAW dataset captured by a Google Pixel camera, e.g., RAW$_{\rm GP}$, is used. This dataset collects images acquired with low-light conditions spanning a range of illumination from 1 lux to 200 lux and containing 1103 images in 40 categories.
%\subsection{Training Details}

We employed the MobileNet-V1~\cite{howard2017mobilenets} for classification as suggested by~\cite{diamond2021dirty}.

As for the proposed $\rho$-Vision, Unpaired CycleR2R is first trained using RGB images in ImageNet (\RGBImageNet) and Google Pixel RAWs (\RAWpixel) to generate a simulated RAW dataset (sim\pseudoRAWpixel). This sim\pseudoRAWpixel~ is augmented with noises and applied to train the RAW-domain classifier MobileNet-V1. Consequently, the trained RAW-domain MobileNet-V1 examines the testing RAWs from \RAWpixel~for task inference.

% Following the methodology in~\cite{diamond2021dirty}, we trained our model using a two-step process. Initially, the dataset was processed using the Unpaired CycleR2R method to generate simulated RAW images that better represent the characteristics of actual RAW data, as opposed to the simple mosaic RAW images used as baselines in~\cite{diamond2021dirty}.

As for the Anscombe ISP method proposed in ~\cite{diamond2021dirty}, ImageNet RGB images (\RGBImageNet) undergo mosaic operations to generate simulated RAWs, which are then injected with Gaussian-Poisson noise to produce noisy simRAWs. The training has two steps: First, the Anscombe ISP is trained with paired noisy RAW and clean RGB images. Second, Anscombe ISP and Imagenet pre-trained MoblieNet-V1 are jointly trained using noisy simRAWs and classification label annotations. 
%Note that the samples fed into the RGB-domain MobileNet-V1 are RGB images outputted from the Anscombe ISP.
During the testing, the Anscombe ISP converts Google Pixel RAW images (\RAWpixel) to the corresponding RGB format (\RGBanscombe)  for classification. %As a result, %Anscombe ISP belongs to is a RGB-Vision
 %A joint training approach is adopted, combining an Anscombe ISP with a MobileNet-V1 classifier. During testing, Google Pixel RAW images (\RAWpixel) are first processed through the Anscombe ISP to RGB format (\RGBanscombe) before classification.

As for the Mosaic RAW method~\cite{diamond2021dirty}, ImageNet images are simply mosaiced to drive RAW samples to form the sim\pseudoRAWpixel.  Noise is then augmented onto the sim\pseudoRAWpixel~ to train the RAW-domain classifier. Subsequently, samples in (\RAWpixel) are tested directly.

Note that noise augmentation closely follows the studies in~\cite{diamond2021dirty} for all approaches. 

%\zhihao{Our Unpaired CycleR2R method differs in its initial training phase, which involves a learnable inverse ISP to generate simRAW images from ImageNet (sim\pseudoRAWpixel), also noise-augmented. The classifier is then trained on these images for RAW domain classification, with direct testing on Google Pixel RAW images (\RAWpixel).}

\subsection{Comparative Studies of RAW-domain Classification}

Table~\ref{table:supp_wo_labeled_classification} reports the image classification under low-light illumination with high noises. The proposed $\rho-Vision$ using Unpaired CycleR2R demonstrates the compellingly superior performance to the approaches, e.g.,  Anscombe ISP and Mosaic RAW, provided by~\cite{diamond2021dirty}. 
 
The gain of the proposed Unpaired CycleR2R to the Mosaic RAW owes the better characterization of real-life RAW images in training/devising the invISP to generate realistic simRAWs. The Mosaic RAW approach~\cite{diamond2021dirty}, instead,  only applies the basic mosaicking by simply neglecting the impacts of gamma correction and white balance that are vital in the transformation between RGB and RAW space..%, leading to a substantial divergence from real RAW images.

The improvement of the Anscombe ISP to the Mosai RAW is due to the mapping between a noisy RAW image and the corresponding clean RGB sample offered by the Anscombe ISP, which significantly helps the subsequent task. 

The gain of the proposed Unpaired CycleR2R to the Anscombe ISP is attributed to the better noise separation and suppression in the RAW domain. This improvement is visually corroborated in Fig.~\ref{figure:supp_classification_zero_shot_visual}, where the ``Noise Channel'' columns under the Unpaired CycleR2R method (RAW row) exhibit a more apparent distinction between noisy and clean features. The efficacy of our model in noise modeling and separation in the RAW domain, as proofed in ~\cite{zhu2017unpaired}, is further evidenced by the Grad-CAM visualizations. These visualizations of noisy inputs are similar to those generated from clean inputs, illustrating the model's ability to preserve essential image characteristics despite noise. In contrast, the Anscombe ISP (RGB row) reveals a significant disparity in the Grad-CAM outputs when comparing noisy and clean inputs, which may lead to classification errors. 

Our Unpaired CycleR2R achieves this superior noise discrimination without increasing computational complexity, thereby maintaining the same level of FLOPs as the Mosaic RAW (lower than the Anscombe ISP). 

%The Unpaired CycleR2R  demonstrates that direct processing of RAW images can lead to accurate classification in low-light conditions while maintaining computational efficiency, a critical factor for practical deployment.

%Mosaic RAW  For image classification under low-light illumination with high noises, characterizing 

%the similarity between training and testing images is critical for inference accuracy.  The Mosaic RAW approach~\cite{diamond2021dirty}, by only applying basic mosaicking, omits essential inverse ISP operations like gamma correction and white balance, leading to a substantial divergence from real RAW images. In contrast, the Anscombe ISP method~\cite{diamond2021dirty} aligns the clean RGB images produced by the ISP, which serve as inputs for the classifier more closely with the noise-free domain, thus reducing the discrepancy and enhancing accuracy. As provided in Table~\ref{table:supp_wo_labeled_classification}, the Anscombe method achieves a +\SI{6.1}{\percent} higher accuracy over the basic Mosaic approach, while at the expense of increased computational demands, indicated by a \SI{56}{\percent} rise in FLOPs.

%Compared with Mosaic RAW, our Unpaired CycleR2R model produces simulated RAW images that better reflect the characteristics of real RAW images, resulting in improved classification accuracy significantly (+\SI{33}{\percent}). Compared with the Anscombe ISP, the Unpaired CycleR2R model demonstrates enhanced performance in RAW-domain classification. 
}

\section{RAW-domain Segmentation}
\label{sec:raw_seg}
In the main text of this paper, the detection task is successfully executed in the RAW domain with superior performance to that using the same RGB-domain model. Here we explore the feasibility of RAW-domain segmentation. Similar to the discussions in Sec.~\ref{sec:exp_obj_det} and~\ref{sec:ablation_gamma}, we first demonstrate that the segmentation model trained with simRAW images can directly infer the segmentation cues upon the real RAW images. Second, a few-shot finetuning simRAW-pretrained segmentation model using limited labeled real RAW images further improves its performance and shows consistent gains to the model trained from scratch. Finally, ablation studies show that gamma correction is also vital for segmentation tasks in the RAW domain. 

\subsection{Datasets}
\textit{Cityscapes}~\cite{Cordts2016Cityscapes} is a large-scale dataset recorded in different urban streets in Europe containing 5,000 frames with high-quality pixel-level segmentation annotations. Considering the different traffic signs in China where the MultiRAW is captured, we use a communal subset including road, building, fence, traffic light, sky, person, car, truck, and bus for evaluation. Following the setup in Sec.~\ref{sec:exp_obj_det} of the main paper, we convert the RGB samples, a.k.a RGB$_{\textrm{c}}$, into simRAW image set simRAW$_{\textrm{c}}$ to train/refine RAW-domain segmentation model.

\subsection{Training Details}
We use the famous HRNetv2~\cite{wang2020deep} as our segmentation network. All segmentation models are optimized by a SGD optimizer with $0.9$ momentum, $5\times10^{-4}$ weight decay and initial learning rate of $10^{-2}$ dropped into $10^{-4}$ linearly. The batch size is set as $8$, and inputs are randomly cropped into $512\times1024$ with random flip augmentation. The experiments are conducted using an Nvidia 3090Ti GPU.

\subsection{Comparative Studies of RAW-domain Segmentation}
\newcommand{\RGBcity}{$\text{RGB}_{\text{c}}$}
\newcommand{\RGBipone}{$\text{RGB}_{\text{i}}$}
\newcommand{\RAWipone}{$\text{RAW}_{\text{i}}$}
\newcommand{\pseudoRAWcity}{$\text{RAW}_{\text{c}}$}
\newcommand{\Baselines}{\textcolor{blue}{$\spadesuit$~}}
\newcommand{\InvISP}{\textcolor{orange}{$\blacksquare$~}}
\newcommand{\DomainAdap}{\textcolor{green}{$\blacklozenge$~}}
\newcommand{\Ours}{\textcolor{red}{$\bigstar$~}}

\begin{table*}
    \small
    \setlength{\tabcolsep}{1.7pt}
    \centering
    \caption{
    {\textbf{mIoU (Mean Intersection over Union) of Segmentation on the testing set of iPhone RAW images.} RGB-domain segmentation model is trained using original RGB images in \textit{Cityscapes}~\cite{Cordts2016Cityscapes} (e.g., \RGBcity); Various simRAW datasets associated with \RGBcity~are generated using different methods which are marked as sim\pseudoRAWcity~to train RAW-domain segmentation model. 
    The testing RAW images in iPhone RAW \RAWipone~and their paired RGB images in \RGBipone~converted using built-in iPhone ISP are tested accordingly. HRNetv2~\cite{wang2020deep} is used as the base segmentation model.
    \Baselines Baselines, \DomainAdap Domain Adaptation Solutions, \InvISP invISP Methods, \Ours Ours.}
    }
    \label{table:supp_wo_labeled_segmentation}
    
    \begin{threeparttable}
    
    \begin{tabular}{lcrccccccccccccccc}
    \toprule
            \multirow{2}*{\textbf{Method}}
            & \textbf{invISP}
            &
            & \multicolumn{2}{c}{\textbf{Segmentor}}
            &
            & \multirow{2}*{\textbf{Road}}
            & \multirow{2}*{\textbf{Build.}}
            & \multirow{2}*{\textbf{Fence}}
            & \multirow{2}*{\textbf{Tr. L.}}
            & \multirow{2}*{\textbf{Sky}}
            & \multirow{2}*{\textbf{Person}}
            & \multirow{2}*{\textbf{Car}}
            & \multirow{2}*{\textbf{Truck}}
            & \multirow{2}*{\textbf{Bus}}
            & \multirow{2}*{\textbf{mIoU}}
            \\\cmidrule{2-2}\cmidrule{4-5}
            
            & {\textbf{Train}}
            &
            & {\textbf{Train}}
            & {\textbf{Test}}
            &
            &
            &
            & 
            &
            &
            & 
            & 
            & 
            &
            \\ \midrule
         \Baselines Naive Baseline 
         & -  
         &
         & \RGBcity 
         & \RAWipone 
         &  
         & 0.3
         & 21.6
         & 14.8
         & 5.7
         & 20.7
         & 0.4
         & 30.0
         & 0.4
         & 6.2
         & 11.1
         \\ 
         \Baselines RGB Baseline  
         & -  
         &
         & \RGBcity 
         & \RGBipone 
         &  
         & \textbf{89.6}
         & 65.1
         & 35.6
         & 20.7
         & \textbf{96.1}
         & 11.1
         & 62.9
         & \textbf{21.5}
         & 25.3
         & 47.5
         \\ \midrule
        \DomainAdap DAFormer (CVPR'22)~\cite{hoyer2022daformer} 
        &  -
        &
        &  \RGBcity,~\RAWipone 
        & \RAWipone         
        &  
        & 75.8
        & 49.5
        & 15.2
        & 1.5
        & 90.0
        & 5.3
        & 58.3
        & 0.2
        & 6.3
        & 32.9
        \\
        \DomainAdap HRDA (ECCV'22)~\cite{hoyer2022hrda} 
        &  -
        &
        & \RGBcity,~\RAWipone 
        & \RAWipone         
        &  
        & 73.8
        & 69.1
        & 38.5
        & 12.3
        & 80.6
        & 15.0
        & 51.2
        & 16.2
        & 20.9
        & 42.0
        \\ \midrule
        \InvISP InvGamma (ICIP'19)~\cite{koskinen2019reverse} 
        &  \RGBipone,~\RAWipone  
        &
        & sim\pseudoRAWcity 
        & \RAWipone  
        &  
        & 47.5
        & 55.7
        & 31.2
        & 8.3
        & 90.0
        & 7.3
        & 23.9
        & 11.2
        & 17.6
        & 32.5
        \\
        \InvISP CycleISP (CVPR'20)~\cite{zamir2020cycleisp}
        & \RGBipone,~\RAWipone
        &
        & sim\pseudoRAWcity
        & \RAWipone 
        &  
        & 84.8
        & 63.9
        & 35.0
        & 18.0
        & 86.3
        & 9.7
        & 55.7
        & 18.0
        & 20.6
        & 43.6
        \\
        \InvISP CIE-XYZ Net (TPAMI'21)~\cite{afifi2020cie} 
        & \RGBipone,~\RAWipone
        &
        & sim\pseudoRAWcity 
        & \RAWipone 
        &  
        & 78.7
        & 64.4
        & 36.7
        & 3.0
        & 84.2
        & 5.4
        & 48.6
        & 2.3
        & 15.4
        & 37.6
        \\
        \InvISP MBISPLD (AAAI'22)~\cite{conde2022model} 
        & \RGBipone,~\RAWipone 
        &
        & sim\pseudoRAWcity 
        & \RAWipone        
        &  
        & 72.5
        & 60.8
        & 39.4
        & 7.3
        & 78.6
        & 13.3
        & 41.0
        & 17.7
        & 20.8
        & 39.0
        \\ \midrule
        \Ours \textbf{Unpaired CycleR2R} 
        & \RGBcity,~\RAWipone 
        &
        & sim\pseudoRAWcity 
        & \RAWipone 
        &  
        & 88.9
        & \textbf{70.5}
        & \textbf{40.9}
        & \textbf{24.7}
        & 95.5
        & \textbf{21.4}
        & \textbf{64.3}
        & 19.1
        & \textbf{30.0}
        & \textbf{50.6}
        \\
    \bottomrule
    \end{tabular}

    \begin{tablenotes}
        \footnotesize 
        \item \textbf{Build.} $\leftarrow$ Building; \textbf{Tr. L.} $\leftarrow$ Traffic Light.
    \end{tablenotes}
    
    \end{threeparttable}

\end{table*}

\newcommand{\reducedstrut}{\vrule width 0pt height .9\ht\strutbox depth .9\dp\strutbox\relax}

\begin{figure*}
\begin{center}
\begin{tabular}[b]{c@{ } c@{ }  c@{ } c@{ } c@{ }}

RAW image &
\InvISP InvGamma~\cite{koskinen2019reverse} &
\InvISP CycleISP~\cite{zamir2020cycleisp} &
\InvISP CIE-XYZ~\cite{afifi2020cie} &
\InvISP MBISPLD~\cite{conde2022model}
\\ \midrule

Ground-truth &
\DomainAdap DAFormer~\cite{hoyer2022daformer} &
\DomainAdap HRDA~\cite{hoyer2022daformer} &
\Baselines RGB Baseline &
\Ours Ours
\vspace{+0.2cm}
\\

\includegraphics[width=.18\textwidth]{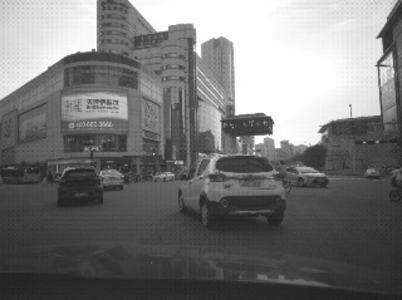}&
\includegraphics[width=.18\textwidth]{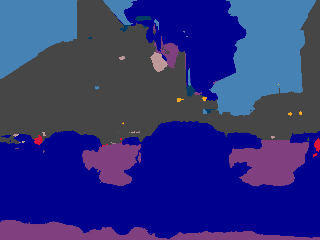}&
\includegraphics[width=.18\textwidth]{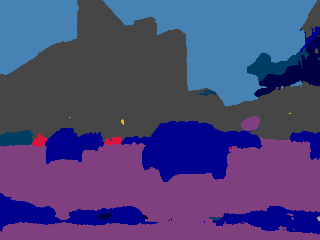}&
\includegraphics[width=.18\textwidth]{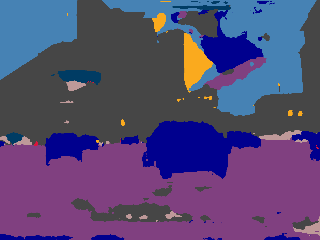}&
\includegraphics[width=.18\textwidth]{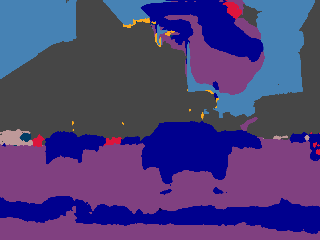}
\\

\includegraphics[width=.18\textwidth]{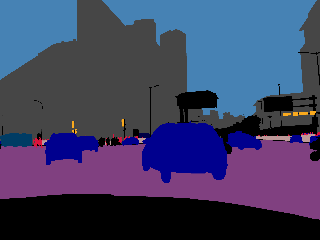}&
\includegraphics[width=.18\textwidth]{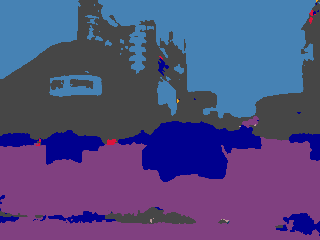}&
\includegraphics[width=.18\textwidth]{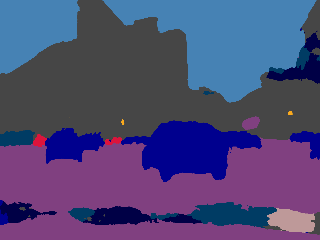}&
\includegraphics[width=.18\textwidth]{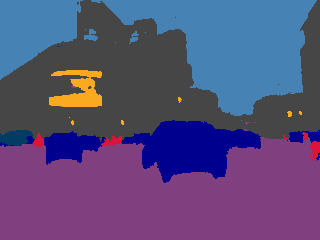}&
\includegraphics[width=.18\textwidth]{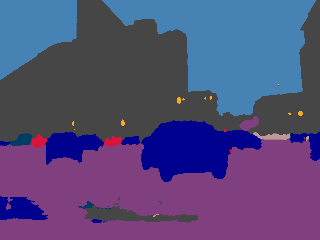}
\\ \vspace{-0.4cm} \\\hdashline \vspace{-0.25cm} \\

\includegraphics[width=.18\textwidth]{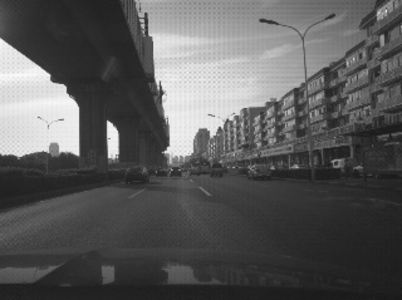}&
\includegraphics[width=.18\textwidth]{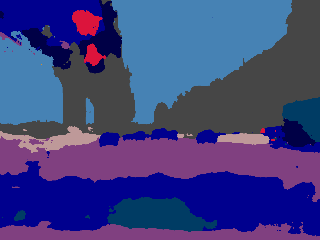}&
\includegraphics[width=.18\textwidth]{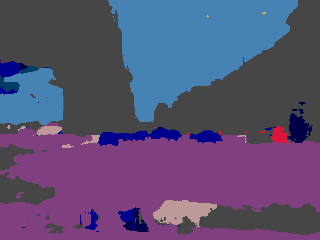}&
\includegraphics[width=.18\textwidth]{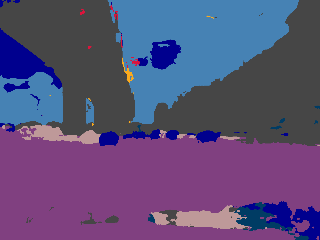}&
\includegraphics[width=.18\textwidth]{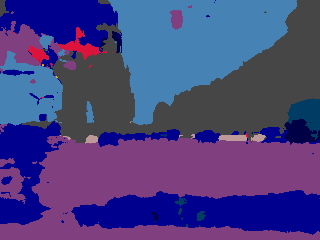}
\\

\includegraphics[width=.18\textwidth]{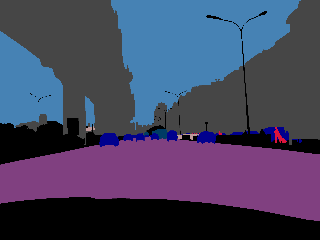}&
\includegraphics[width=.18\textwidth]{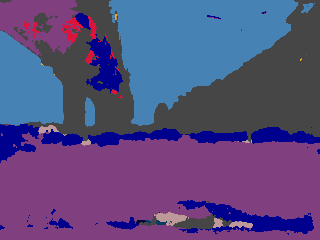}&
\includegraphics[width=.18\textwidth]{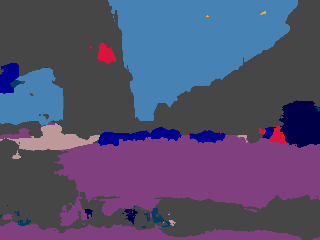}&
\includegraphics[width=.18\textwidth]{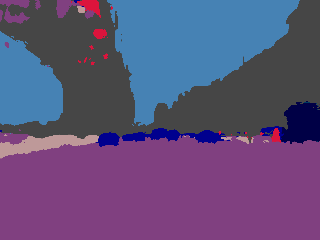}&
\includegraphics[width=.18\textwidth]{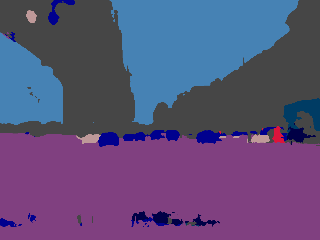}
\\ \vspace{-0.4cm} \\\hdashline \vspace{-0.25cm} \\

\includegraphics[width=.18\textwidth]{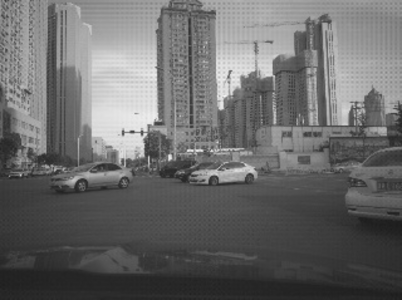}&
\includegraphics[width=.18\textwidth]{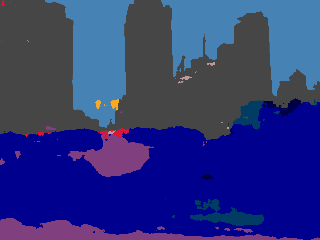}&
\includegraphics[width=.18\textwidth]{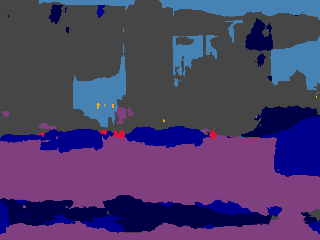}&
\includegraphics[width=.18\textwidth]{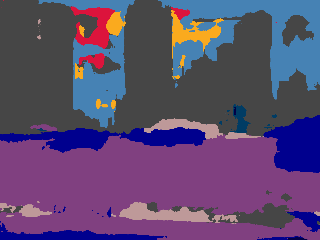}&
\includegraphics[width=.18\textwidth]{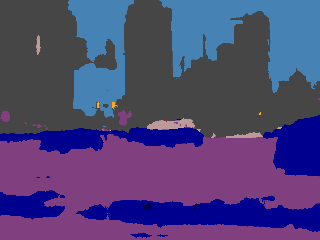}
\\

\includegraphics[width=.18\textwidth]{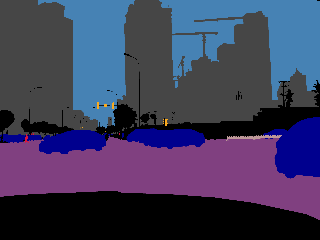}&
\includegraphics[width=.18\textwidth]{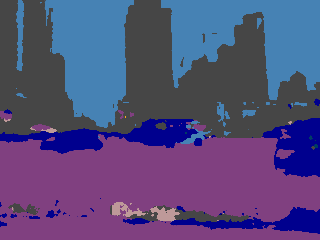}&
\includegraphics[width=.18\textwidth]{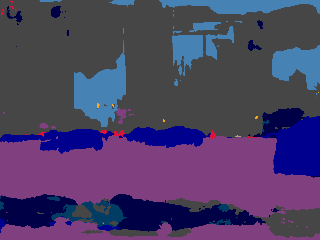}&
\includegraphics[width=.18\textwidth]{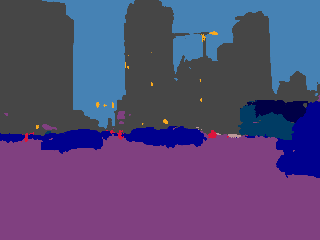}&
\includegraphics[width=.18\textwidth]{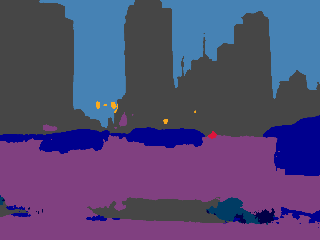}
\\ \vspace{-0.4cm} \\\hdashline \vspace{-0.25cm} \\

\includegraphics[width=.18\textwidth]{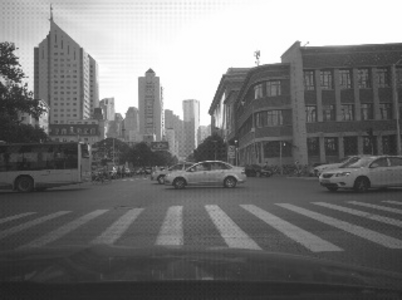}&
\includegraphics[width=.18\textwidth]{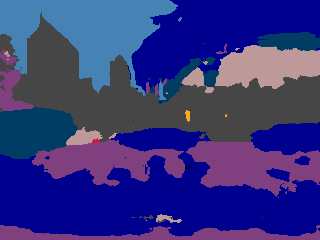}&
\includegraphics[width=.18\textwidth]{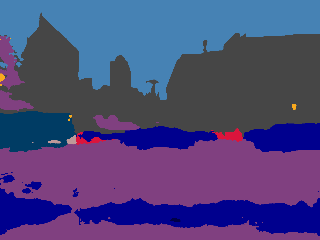}&
\includegraphics[width=.18\textwidth]{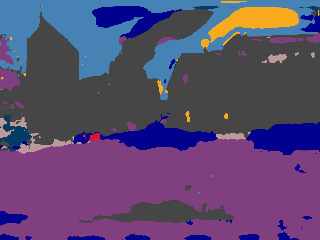}&
\includegraphics[width=.18\textwidth]{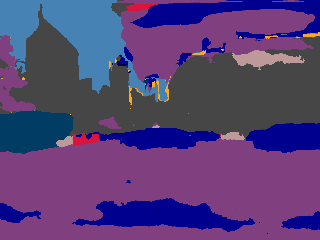}
\\

\includegraphics[width=.18\textwidth]{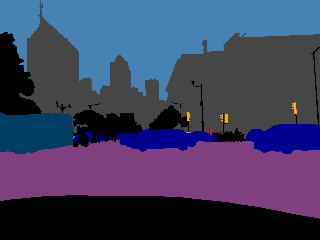}&
\includegraphics[width=.18\textwidth]{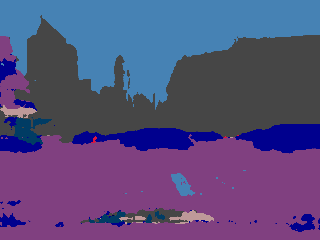}&
\includegraphics[width=.18\textwidth]{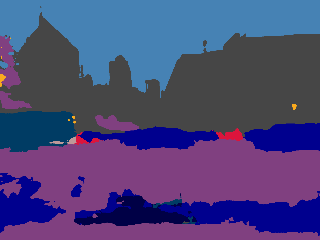}&
\includegraphics[width=.18\textwidth]{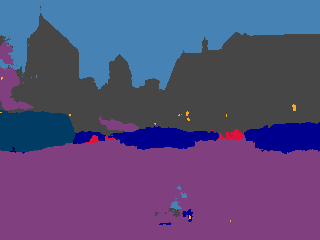}&
\includegraphics[width=.18\textwidth]{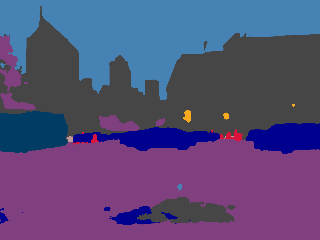}
\\

\multicolumn{5}{c}{
\colorbox[RGB]{128, 64, 128}{\parbox[c][0.8em]{4em}{\makebox[4em]{\footnotesize \color{white} Road}}}\reducedstrut
\colorbox[RGB]{70, 70, 70}{\parbox[c][0.8em]{4em}{\makebox[4em]{\footnotesize \color{white} Build.}}}\reducedstrut
\colorbox[RGB]{190, 153, 153}{\parbox[c][0.8em]{4em}{\makebox[4em]{\footnotesize Fence}}}\reducedstrut
\colorbox[RGB]{250, 170, 30}{\parbox[c][0.8em]{4em}{\makebox[4em]{\footnotesize Tr. L.}}}\reducedstrut
\colorbox[RGB]{70, 130, 180}{\parbox[c][0.8em]{4em}{\makebox[4em]{\footnotesize Sky}}}\reducedstrut
\colorbox[RGB]{220, 20, 60}{\parbox[c][0.8em]{4em}{\makebox[4em]{\footnotesize \color{white} Person}}}\reducedstrut
\colorbox[RGB]{0, 0, 142}{\parbox[c][0.8em]{4em}{\makebox[4em]{\footnotesize \color{white} Car}}}\reducedstrut
\colorbox[RGB]{0, 0, 70}{\parbox[c][0.8em]{4em}{\makebox[4em]{\footnotesize \color{white} Truck}}}\reducedstrut
\colorbox[RGB]{0, 60, 100}{\parbox[c][0.8em]{4em}{\makebox[4em]{\footnotesize \color{white} Bus}}}\reducedstrut
\colorbox[RGB]{0, 0, 0}{\parbox[c][0.8em]{4em}{\makebox[4em]{\footnotesize \color{white} N/A.}}}\reducedstrut
}
\\

\end{tabular}
\end{center}
\caption{{\bf Qualitative Visualization of Pretrained RAW Segmentation Model.} Example predictions show better recognition of buildings, sky, and traffic lights by our Unpaired CycleR2R on Cityscapes RGB $\rightarrow$ iPhone RAW. Gamma correction and brightness adjustment have been applied to RAW images for a better view.}
\label{figure:zero_shot_seg_visual}
\end{figure*}
Table~\ref{table:supp_wo_labeled_segmentation} and Fig.~\ref{figure:zero_shot_seg_visual} compares our Unpaired CycleR2R and other methods using invISP approach~\cite{conde2022model, zamir2020cycleisp, afifi2020cie, koskinen2019reverse} and domain-adaptation (DA) solution~\cite{hoyer2022hrda, hoyer2022daformer}.
It can be seen that Unpaired CycleR2R outperforms the state-of-the-art CycleISP 
by a significant margin of \SI{7}{mIoU} and improves the IoU across all classes of objects. More gains are presented against other approaches.

Our model also surpasses the RGB Baseline on mIoU. Note that this RGB Baseline is prevalent in real-world applications. Such a convincing performance suggests the potential for RAW-domain segmentation. We also observe the lower IoU for some specific classes of objects between our method and the RGB Baseline. This is probably due to the optimization strategy for maximizing the overall performance but not balancing each class. This is an interesting topic for future study.

Apparently, inputting real RAW images to the RGB-domain segmentation model directly for task execution is a failure, as exemplified in the Naive Baseline, e.g., mIoU of 11.1 versus the mIoU of 47.5 in the RGB Baseline, which is due to the large discrepancy between the RGB-domain and RAW-domain models.

\textbf{Implementation Friendliness.} As aforementioned in Sec.~\ref{sec:exp_obj_det}, our method could generate simRAW images to train task-dependent models. However, DA-based approaches~\cite{hnewa2021multiscale,li2022cross} designed for object detection tasks could not be applied to segmentation tasks. And DA-based segmentation methods~\cite{hoyer2022daformer,hoyer2022hrda} could not support the detection task either.

\subsection{Comparative Studies of Few-Shot Finetuning}
\begin{figure}
\centering
{\includegraphics[width=1.0\linewidth]{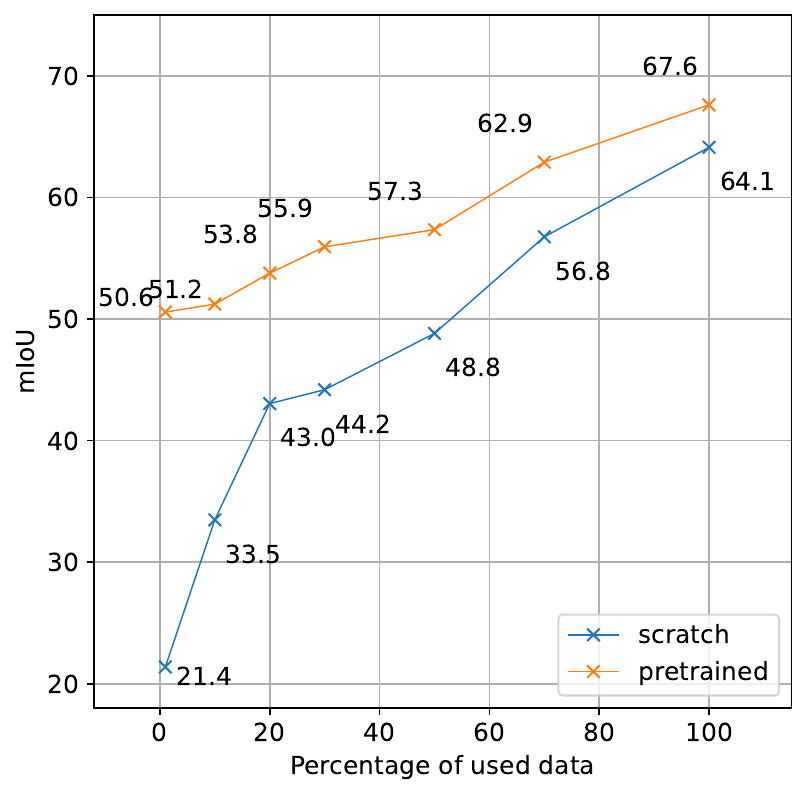} \label{fig:segmentation-few-shot}}
  \caption{{\bf Few-shot finetuning using limited camera RAWs.} The simRAW-pretrained HRNetv2~\cite{wang2020deep} is obtained by using samples in simRAW$_\text{c}$ generated by our Unpaired CycleR2R, which is then finetuned using limited camera RAW images; and the ``scratch'' model is randomly initialized and then trained using the same number of labeled real RAW images.}
\label{fig:Few_Shot_Segmentation}
\end{figure}
The performance of the simRAW-pretrained segmentation model could be further boosted by feeding more real labeled RAW images. We further finetune our segmentation model using our MultiRAW dataset (iPhone XSmax) with all classes. As depicted in Fig.~\ref{fig:Few_Shot_Segmentation}, the segmentation accuracy is improved and consistently outperforms the scratch model which is initialized randomly and then trained using the same labeled real RAW images. 

\section{Extra Quantitative Visualization}
\label{sec:more_figures}
\begin{figure*}
\centering

    \subfloat[simRAW is a superset of real RAW data]{\includegraphics[width=0.36\linewidth]{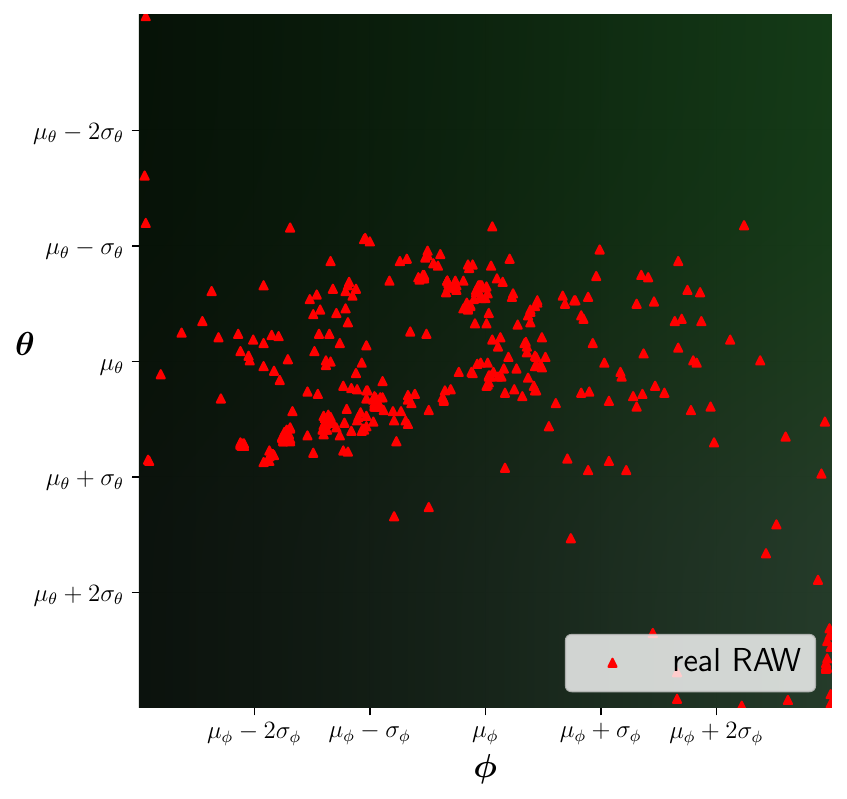} \label{fig:iem_plot}}
    \quad
    \subfloat[simRAW examples]{\includegraphics[width=0.593\linewidth]{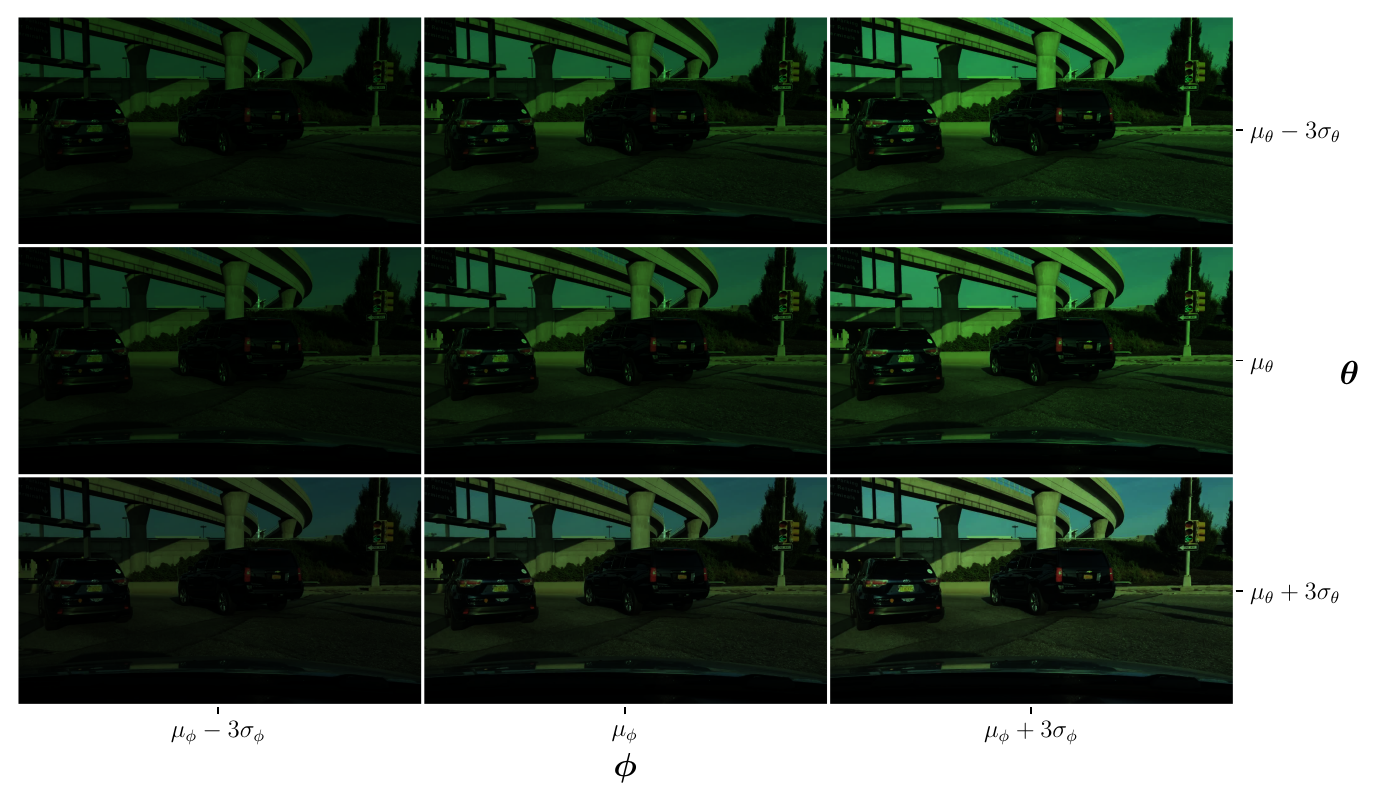} \label{fig:iem_visual}}

    \subfloat[ Testing RGB]{\includegraphics[width=0.19\linewidth]{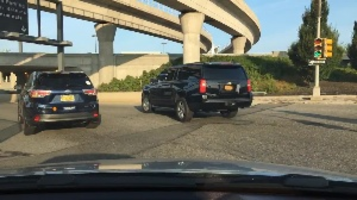} \label{fig:bdd_sample}}
    \quad
    \subfloat[Random real RAW images from multiRAW dataset]{\includegraphics[width=0.77\linewidth]{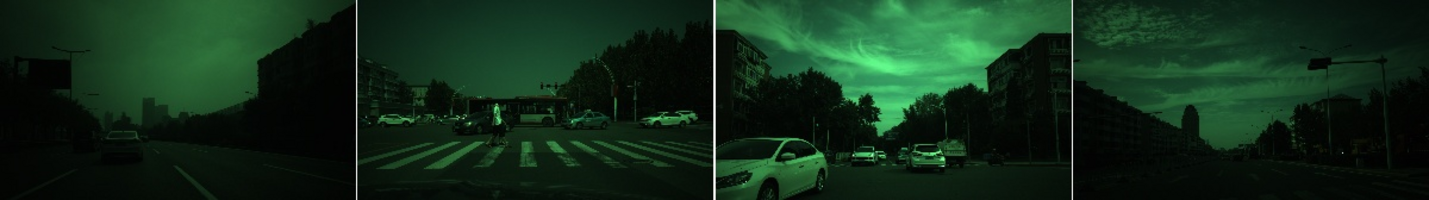} \label{fig:iphone_batch_samples}}
    
    \subfloat[simRAW w/o IEM]{\includegraphics[width=0.19\linewidth]{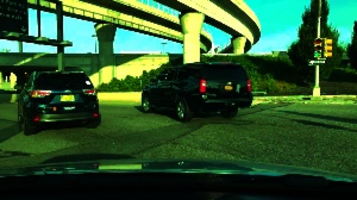} \label{fig:wo_iem_certain}}
    \quad
    \subfloat[simulating RAW images by injecting noise to $E^{-1}(\cdot)$ outputs w/o using IEM]{\includegraphics[width=0.77\linewidth]{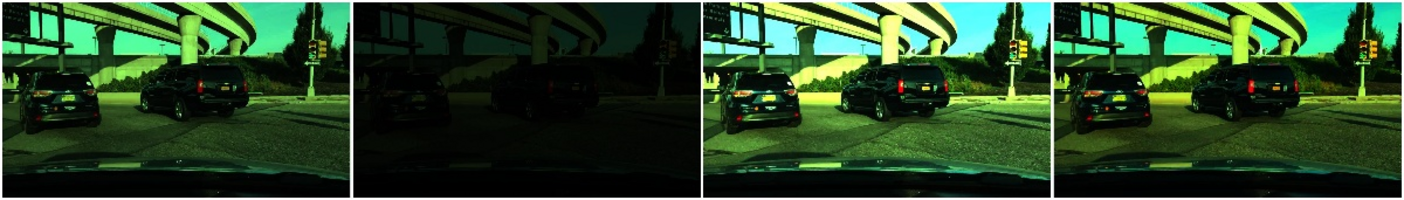} \label{fig:wo_iem_random}}
    
    \caption{
    {\bf Evaluation of the Illumination Estimation Module (IEM). {\it Demosaicing has been applied to all images to enhance visibility.}} 
    (a) Adapting IEM to generate the simRAW's coverage using the mean color, where the color of each point $\phi_i$, $\theta_j$ represents the average color of a simRAW generated by sampled illumination parameters $\phi_i$, $\theta_j$. In contrast,  red markers indicate the average color of real RAW images.
    It clearly reveals that adapting IEM can cover all real-world illumination conditions in real RAW data. 
    (b)  simRAW examples generated by our Unpaired CycleR2R with various $\phi$, $\theta$, illustrating the IEM's ability to produce a wide range of illumination variations.
    (c) The corresponding RGB image fed into the invISP of our Unpaired CycleR2R, which is from the BDD100K dataset.
    (d) Random real RAW images from the multiRAW dataset, displaying the natural variability in illumination and color temperature.
    (e) Simulating a RAW image without using IEM, which can only produce a single simRAW per RGB input due to the absence of probabilistic illumination estimation. 
    (f) Simulating RAW images by injecting noise to $E^{-1}(\cdot)$ outputs, which can produce multiple simRAW samples without requiring IEM but demonstrate unrealistic diversity induction in RAW Images.  $E^{-1}(\cdot)$ is defined in invISP (see Fig. 2 in the main paper).
    }
\label{fig:supp_iem_ablation}
\end{figure*}

\newcommand{\AddCompareImg}[2]{\includegraphics[width=#1\textwidth]{#2}}

\newcommand{\AddCompareImgs}[2]{
\begin{minipage}{0.3\textwidth}
\begin{center}
\AddCompareImg{1.0}{Figs/lossy_compression_compare/#1/#2.png} \vspace{-0.7em}\\
\AddCompareImg{0.48}{Figs/lossy_compression_compare/#1/crop_0/#2.png}
\AddCompareImg{0.48}{Figs/lossy_compression_compare/#1/crop_1/#2.png}
\end{center}
\end{minipage}
}

\newcommand{\AddCompareLineA}[1]{%
    \AddCompareImgs{#1_adjust}{hm} &
    \AddCompareImgs{#1_adjust}{hm_10bits} &
    \AddCompareImgs{#1_adjust}{ours}
}

\newcommand{\AddCompareLineB}[6]{%
    HEVC &
    HECV-10bits &
    Ours \\
    \SI{#1}{bpp} / \SI{#2}{\dB} &
    \SI{#3}{bpp} / \SI{#4}{\dB} &
    \SI{#5}{bpp} / \SI{#6}{\dB}
}

\newcommand{\AddCompareLineC}[1]{%
    \AddCompareImgs{#1_adjust}{vvc} &
    \AddCompareImgs{#1_adjust}{vvc_12bits} &
    \AddCompareImgs{#1_adjust}{gt}
}

\newcommand{\AddCompareLineD}[4]{%
    VVC &
    VVC-12bits &
    GT \\
    \SI{#1}{bpp} / \SI{#2}{\dB} &
    \SI{#3}{bpp} / \SI{#4}{\dB} &
    \SI{12}{bpp} / -
}
    
\begin{figure*}
    \begin{center}
    \begin{tabular}[b]{c@{ } c@{ } c@{}}

    \AddCompareLineA{1087} \\
    \AddCompareLineB
        {0.010}{41.73}
        {0.010}{42.28}
        {0.010}{44.89} \vspace{+0.5em} \\
    \AddCompareLineC{1087} \\
    \AddCompareLineD
        {0.010}{42.94}
        {0.010}{43.87} \\
    
    \end{tabular}
    \end{center}
    \caption{{\bf Qualitative Visualization of Lossy RIC at Low Bits-rate.} Reconstructions and close-ups of the HEVC, VVC, and our method. Corresponding bpp and PSNR are marked. Gamma correction and brightness adjustment have been applied for a better view. {\it Zoom for better details.}}
    \label{figure:supp_lossy_compression_visual_low}
\end{figure*}

\begin{figure*}
    \begin{center}
    \begin{tabular}[b]{c@{ } c@{ } c@{}}

    \AddCompareLineA{945} \\
    \AddCompareLineB
        {0.031}{44.73}
        {0.031}{46.05}
        {0.032}{48.46} \vspace{+0.5em} \\
    \AddCompareLineC{945} \\
    \AddCompareLineD
        {0.032}{45.25}
        {0.032}{46.95} \\
    
    \end{tabular}
    \end{center}
    \caption{{\bf Qualitative Visualization of Lossy RIC at High Bits-rate.} Reconstructions and close-ups of the HEVC, VVC, and our method. Corresponding bpp and PSNR are marked. Gamma correction and brightness adjustment have been applied for a better view. {\it Zoom for better details.}}
    \label{figure:supp_lossy_compression_visual_high}
\end{figure*}
\newcommand{\AddImg}[2]{\includegraphics[width=#1\textwidth]{#2}}
\newcommand{\AddImgs}[3]{
\begin{minipage}{0.18\textwidth}
\begin{center}
\AddImg{1.0}{Figs/progressive_decoding/#1/#2/#3} \vspace{-1em}\\
\AddImg{0.48}{Figs/progressive_decoding/#1/#2/crop_0/#3}
\AddImg{0.48}{Figs/progressive_decoding/#1/#2/crop_1/#3}
\end{center}
\end{minipage}
}

\newcommand{\AddIphone}[6]{
\AddImgs{iphone_adjust}{#1}{#2} &
\AddImgs{iphone_adjust}{#1}{#3} &
\AddImgs{iphone_adjust}{#1}{#4} &
\AddImgs{iphone_adjust}{#1}{#5} &
\AddImgs{iphone_adjust}{#1}{#6}
}
\newcommand{\AddHuawei}[6]{
\AddImgs{huawei_adjust}{#1}{#2} &
\AddImgs{huawei_adjust}{#1}{#3} &
\AddImgs{huawei_adjust}{#1}{#4} &
\AddImgs{huawei_adjust}{#1}{#5} &
\AddImgs{huawei_adjust}{#1}{#6}
}
\newcommand{\AddASI}[6]{
\AddImgs{asi_adjust}{#1}{#2} &
\AddImgs{asi_adjust}{#1}{#3} &
\AddImgs{asi_adjust}{#1}{#4} &
\AddImgs{asi_adjust}{#1}{#5} &
\AddImgs{asi_adjust}{#1}{#6}
}

\newcommand{\AddRAWCaption}[9]{
\SI{#1}{bpp} / \SI{#6}{\dB} &
\SI{#2}{bpp} / \SI{#7}{\dB} &
\SI{#3}{bpp} / \SI{#8}{\dB} &
\SI{#4}{bpp} / \SI{#9}{\dB} &
\SI{#5}{bpp} / GT 
}
\newcommand{\AddRGBCaption}[9]{
\SI{#1}{\second} / \SI{#6}{\dB} &
\SI{#2}{\second} / \SI{#7}{\dB} &
\SI{#3}{\second} / \SI{#8}{\dB} &
\SI{#4}{\second} / \SI{#9}{\dB} &
\SI{#5}{\second} / GT
}

\begin{figure*}
\begin{center}
\begin{tabular}[b]{c@{ } c@{ } c@{ } c@{ } c@{ } }
\AddIphone
{602}
{stage_1_mu_raw_bpp0.14160490036010742_psnr25.547616481781006.png}
{stage_2_mu_raw_bpp0.2102944254875183_psnr29.08344268798828.png}
{stage_3_mu_raw_bpp0.48261576890945435_psnr34.78823661804199.png}
{stage_4_mu_raw_bpp1.5291390419006348_psnr44.12456035614014.png}
{gt_raw.png}
\\
\AddRAWCaption
{0.14}{0.21}{0.48}{1.53}{5.62}
{25.55}{29.08}{34.78}{44.12}
\vspace{+0.3cm}
\\

% Iphone

\AddIphone
{602}
{stage_1_mu_rgb_bpp0.14160490036010742_psnr22.4583941661336.png}
{stage_2_mu_rgb_bpp0.2102944254875183_psnr25.75342095254015.png}
{stage_3_mu_rgb_bpp0.48261576890945435_psnr30.929055058926565.png}
{stage_4_mu_rgb_bpp1.5291390419006348_psnr37.45431981618415.png}
{gt_rgb.png}
\\
\AddRGBCaption
{0.13}{0.24}{0.35}{0.50}{0.67}
{22.46}{25.75}{30.93}{37.45}
\\ \vspace{-0.2cm} \\\hdashline \vspace{-0.05cm} \\

\AddIphone
{1000}
{stage_1_mu_raw_bpp0.05445677787065506_psnr31.984107494354248.png}
{stage_2_mu_raw_bpp0.10802081972360611_psnr35.64331531524658.png}
{stage_3_mu_raw_bpp0.3288505971431732_psnr40.812034606933594.png}
{stage_4_mu_raw_bpp1.1682013273239136_psnr48.76646041870117.png}
{gt_raw.png}
\\
\AddRAWCaption
{0.05}{0.11}{0.33}{1.17}{4.46}
{31.98}{35.64}{40.81}{48.76}
\vspace{+0.3cm}
\\

\AddIphone
{1000}
{stage_1_mu_rgb_bpp0.05445677787065506_psnr20.581228739017998.png}
{stage_2_mu_rgb_bpp0.10802081972360611_psnr22.319588827976908.png}
{stage_3_mu_rgb_bpp0.3288505971431732_psnr24.130143501165445.png}
{stage_4_mu_rgb_bpp1.1682013273239136_psnr26.762751963093393.png}
{gt_rgb.png}
\\
\AddRGBCaption
{0.11}{0.22}{0.36}{0.58}{0.73}
{20.58}{22.32}{24.13}{26.76}
    
\end{tabular}
\end{center}
\caption{{\bf Qualitative Visualization of Lossless RIC Progressive Decoding (iPhone XSmax).} The gradual reconstruction of RAW images and their corresponding RGB images converted by an in-camera ISP. Bits per pixel (bpp) / PSNR (dB) is shown under RAW images. Decoding latency (s) / PSNR (dB) is also listed below RGB images. PSNR is derived against the GT (ground truth). Gamma correction and brightness adjustment have been applied for a better view. {\it Zoom for more details.}}
\label{figure:supp_progressive_decoding_iphone}
\end{figure*}

% Huawei

\begin{figure*}
    \begin{center}
    \begin{tabular}[b]{c@{ } c@{ } c@{ } c@{ } c@{ } }
    \AddHuawei
    {IMG_20210911_165107}
    {stage_1_mu_raw_bpp0.07082276046276093_psnr34.41326379776001.png}
    {stage_2_mu_raw_bpp0.1337396502494812_psnr38.606181144714355.png}
    {stage_3_mu_raw_bpp0.3613351583480835_psnr44.27355766296387.png}
    {stage_4_mu_raw_bpp1.2054872512817383_psnr51.069722175598145.png}
    {gt_raw.png}
    \\
    \AddRAWCaption
    {0.07}{0.13}{0.36}{1.20}{4.32}
    {34.41}{38.60}{44.27}{51.06}
    \vspace{+0.3cm}
    \\
    
    \AddHuawei
    {IMG_20210911_165107}
    {stage_1_mu_rgb_bpp0.07082276046276093_psnr24.85024428605815.png}
    {stage_2_mu_rgb_bpp0.1337396502494812_psnr28.47787646776431.png}
    {stage_3_mu_rgb_bpp0.3613351583480835_psnr33.240419936837654.png}
    {stage_4_mu_rgb_bpp1.2054872512817383_psnr38.75480322314341.png}
    {gt_rgb.png}
    \\
    \AddRGBCaption
    {0.13}{0.23}{0.34}{0.59}{0.74}
    {24.85}{28.48}{33.24}{38.75}
    \\ \vspace{-0.2cm} \\\hdashline \vspace{-0.05cm} \\
    
    \AddHuawei
    {IMG_20210911_194243}
    {stage_1_mu_raw_bpp0.04557344689965248_psnr31.701772212982178.png}
    {stage_2_mu_raw_bpp0.11266462504863739_psnr34.772000312805176.png}
    {stage_3_mu_raw_bpp0.3728232979774475_psnr39.783713817596436.png}
    {stage_4_mu_raw_bpp1.4020719528198242_psnr45.96439838409424.png}
    {gt_raw.png}
    \\
    \AddRAWCaption
    {0.05}{0.11}{0.37}{1.40}{5.45}
    {31.70}{34.77}{39.78}{45.96}
    \vspace{+0.3cm}
    \\
    
    \AddHuawei
    {IMG_20210911_194243}
    {stage_1_mu_rgb_bpp0.04557344689965248_psnr24.33571219758751.png}
    {stage_2_mu_rgb_bpp0.11266462504863739_psnr26.31830644884154.png}
    {stage_3_mu_rgb_bpp0.3728232979774475_psnr27.878994824018335.png}
    {stage_4_mu_rgb_bpp1.4020719528198242_psnr29.951772818535666.png}
    {gt_rgb.png}
    \\
    \AddRGBCaption
    {0.18}{0.23}{0.36}{0.58}{0.71}
    {24.34}{26.32}{27.88}{29.95}
        
    \end{tabular}
    \end{center}
    \caption{{\bf Qualitative Visualization of Lossless RIC Progressive Decoding (Huawei P30pro).} The gradual reconstruction of RAW images and their corresponding RGB images converted by an in-camera ISP. Bits per pixel (bpp) / PSNR (dB) is shown under RAW images. Decoding latency (s) / PSNR (dB) is also listed below RGB images. PSNR is derived against the GT (ground truth). Gamma correction and brightness adjustment have been applied for a better view. {\it Zoom for more details.}}
    \label{figure:supp_progressive_decoding_huawei}
\end{figure*}

% ASI 

\begin{figure*}
    \begin{center}
    \begin{tabular}[b]{c@{ } c@{ } c@{ } c@{ } c@{ } }
    \AddASI
    {2021-09-11-0802_1-CapObj_0000}
    {stage_1_mu_raw_bpp0.029787881299853325_psnr29.673008918762207.png}
    {stage_2_mu_raw_bpp0.07659978419542313_psnr33.231613636016846.png}
    {stage_3_mu_raw_bpp0.24867552518844604_psnr36.852171421051025.png}
    {stage_4_mu_raw_bpp1.079272985458374_psnr43.690290451049805.png}
    {gt_raw.png}
    \\
    \AddRAWCaption
    {0.03}{0.08}{0.25}{1.08}{3.58}
    {29.67}{33.23}{36.85}{43.69}
    \vspace{+0.3cm}
    \\
    
    \AddASI
    {2021-09-11-0802_1-CapObj_0000}
    {stage_1_mu_rgb_bpp0.029787881299853325_psnr19.670416584845146.png}
    {stage_2_mu_rgb_bpp0.07659978419542313_psnr22.18316301378723.png}
    {stage_3_mu_rgb_bpp0.24867552518844604_psnr25.26291162893699.png}
    {stage_4_mu_rgb_bpp1.079272985458374_psnr29.89796580507353.png}
    {gt_rgb.png}
    \\
    \AddRGBCaption
    {0.12}{0.23}{0.34}{0.59}{0.82}
    {19.67}{22.18}{25.26}{29.89}
    \\ \vspace{-0.2cm} \\\hdashline \vspace{-0.05cm} \\
    
    \AddASI
    {20210910204035}
    {stage_1_mu_raw_bpp0.02449415996670723_psnr31.679437160491943.png}
    {stage_2_mu_raw_bpp0.0374918058514595_psnr35.413851737976074.png}
    {stage_3_mu_raw_bpp0.08529061079025269_psnr40.651092529296875.png}
    {stage_4_mu_raw_bpp0.37319064140319824_psnr48.040008544921875.png}
    {gt_raw.png}
    \\
    \AddRAWCaption
    {0.02}{0.04}{0.09}{0.37}{1.05}
    {31.68}{35.41}{40.65}{48.04}
    \vspace{+0.3cm}
    \\
    
    \AddASI
    {20210910204035}
    {stage_1_mu_rgb_bpp0.02449415996670723_psnr20.49284781241702.png}
    {stage_2_mu_rgb_bpp0.0374918058514595_psnr22.56786474576234.png}
    {stage_3_mu_rgb_bpp0.08529061079025269_psnr25.417073162649594.png}
    {stage_4_mu_rgb_bpp0.37319064140319824_psnr29.716557587771355.png}
    {gt_rgb.png}
    \\
    \AddRGBCaption
    {0.12}{0.23}{0.39}{0.60}{0.80}
    {20.49}{22.57}{25.42}{29.72}
        
    \end{tabular}
    \end{center}
    \caption{{\bf Qualitative Visualization of Lossless RIC Progressive Decoding (asi 294mcpro).} The gradual reconstruction of RAW images and their corresponding RGB images converted by an in-camera ISP. Bits per pixel (bpp) / PSNR (dB) is shown under RAW images. Decoding latency (s) / PSNR (dB) is also listed below RGB images. PSNR is derived against the GT (ground truth). Gamma correction and brightness adjustment have been applied for a better view. {\it Zoom for more details.}}
    \label{figure:supp_progressive_decoding_asi}
\end{figure*}
%\zhihao{The change of this part is moved to rho\_vision$.$tex}

\edited{In Fig.~\ref{fig:supp_iem_ablation}, we present a visual comparison between our simulated RAW images and real RAW images.} We also offer more qualitative visualizations of our lossy RIC at low Bits-rate and high Bits-rate in Fig.~\ref{figure:supp_lossy_compression_visual_low} and Fig.~\ref{figure:supp_lossy_compression_visual_high} respectively. Similar to the results in the main content of this work, we can clearly observe the subjective improvements of the proposed lossy RIC compared to the HEVC and VVC. Especially for the traffic light and car information, our lossy RIC provides sharper and less noisy
reconstructions closer to the ground truth samples. Also, we give visualizations of progressive decoding using our lossless RIC within various cameras in Fig.~\ref{figure:supp_progressive_decoding_iphone}-\ref{figure:supp_progressive_decoding_asi}. Our lossless RIC could provide low-resolution previews for different cameras (iPhone XSmax, Huawei P30pro, and asi 294mcpro) and different scenes (both daylight and nighttime), which is helpful for professional photography and network transmission.

\bibliographystyle{IEEEtran}
\bibliography{supply_material/supp_ref.bib}
%\appendix
%\input{sections/Appendix}